\documentclass[10pt,letterpaper]{article}

\newcommand{\ReportTitle}{\GreenWAM: A World Foundation Model for Robotics}
\newcommand{\ReportTitlePlain}{GreenWAM: A World Foundation Model for Robotics}

\usepackage[letterpaper,margin=1in]{geometry}

\usepackage{fontspec}
\usepackage{unicode-math}
\newfontfamily\cyrillicfontsf{NotoSans-Regular.ttf}[
  BoldFont=NotoSans-Bold.ttf,
  ItalicFont=NotoSans-Italic.ttf,
  BoldItalicFont=NotoSans-BoldItalic.ttf
]
\usepackage{polyglossia}
\setdefaultlanguage{english}
\setotherlanguage{russian}
\usepackage[protrusion=true,expansion=false]{microtype}

\usepackage[dvipsnames,table]{xcolor}
\usepackage{graphicx}
\usepackage{adjustbox}
\usepackage{booktabs}
\usepackage{multirow}
\usepackage{makecell}
\usepackage{float}
\usepackage{placeins} %
\usepackage{pifont}
\usepackage{xspace}

\usepackage[font={small},labelfont={bf,sf},labelsep=space,
  justification=justified,singlelinecheck=false]{caption}
\AtBeginEnvironment{table}{\renewcommand{\arraystretch}{1.12}}
\newcommand{\tablenote}[1]{\par\vspace{5pt}{\footnotesize\raggedright #1\par}}

\usepackage{amsmath}

\definecolor{brandgreen}{HTML}{1EB980}
\definecolor{brandpurple}{HTML}{6C63FF}
\definecolor{seedblue}{HTML}{2E5AA8}
\definecolor{warmorange}{HTML}{B4531A}
\definecolor{wmteal}{HTML}{2A817B}  %
\definecolor{reportnavy}{HTML}{1F3A68}
\definecolor{accentgreen}{HTML}{12875F}
\definecolor{panelfill}{HTML}{F2F6F4}
\definecolor{vizreal}{HTML}{2A78D6}
\definecolor{vizsim}{HTML}{EB6834}
\usepackage{pgfplots}
\pgfplotsset{compat=1.18}
\definecolor{top1color}{HTML}{63D3B0}
\definecolor{top2color}{HTML}{D6D4FF}
\definecolor{top3color}{HTML}{EBF4FF}

\usepackage{titlesec}
\titleformat{\section}
  {\Large\sffamily\bfseries\color{reportnavy}}
  {\textcolor{accentgreen}{\thesection}}{0.8em}{}
\titlespacing*{\section}{0pt}{3.5ex plus 1ex minus .2ex}{2.3ex plus .2ex}
\titleformat{\subsection}
  {\large\sffamily\bfseries\color{reportnavy}}
  {\textcolor{accentgreen}{\thesubsection}}{0.8em}{}
\titlespacing*{\subsection}{0pt}{3.25ex plus 1ex minus .2ex}{1.5ex plus .2ex}
\titleformat{\subsubsection}
  {\normalsize\sffamily\bfseries\color{reportnavy}}
  {\textcolor{accentgreen}{\thesubsubsection}}{0.8em}{}
\titlespacing*{\subsubsection}{0pt}{3.25ex plus 1ex minus .2ex}{1.5ex plus .2ex}

\DeclareTextFontCommand{\textbf}{\bfseries\sffamily}

\newcommand{\para}[1]{\noindent\textbf{#1}\quad}

\usepackage[numbers]{natbib}
\usepackage[colorlinks=true]{hyperref}
\usepackage[capitalize,noabbrev]{cleveref}
\hypersetup{
  pdftitle={\ReportTitlePlain},
  pdfauthor={Gennady Sidorov, Valerii Kornilov, Denis Grankin, Polina Fedotova, Aleksey Postnikov},
  linkcolor=accentgreen,
  citecolor=accentgreen,
  urlcolor=accentgreen
}

\usepackage[most]{tcolorbox}
\newtcolorbox{titlepanel}{enhanced,colback=panelfill,colframe=brandgreen!45,boxrule=0.6pt,arc=3mm,
  left=7mm,right=7mm,top=5mm,bottom=4.5mm}

\definecolor{qfill}{HTML}{F1EFFE}
\newtcolorbox{questionbox}[1]{enhanced,colback=qfill,colframe=brandpurple!55,boxrule=0.5pt,arc=2mm,
  left=3mm,right=3mm,top=1.6mm,bottom=1.6mm,before skip=6pt,after skip=8pt,
  fontupper=\sffamily\small,before upper={\textbf{\color{brandpurple}Question #1.}\ }}

\usepackage{wrapfig}
\usepackage{needspace}
\usepackage{wrapstuff}
\newcommand{\figpair}[6]{%
  \begin{figure}[!htb]
    \begin{minipage}[t]{.48\linewidth}\centering #1\caption{#2}\label{#3}\end{minipage}\hfill
    \begin{minipage}[t]{.48\linewidth}\centering #4\caption{#5}\label{#6}\end{minipage}
  \end{figure}}

\newlength{\sideneed}
\newlength{\sideleft}
\newcommand{\sidefitcheck}[4]{%
  \setbox0=\hbox{\includegraphics[width=#1\linewidth]{#2}}%
  \setbox2=\vbox{\hsize=#1\linewidth\small #3}%
  \setlength{\sideneed}{\dimexpr\ht0+\dp0+\ht2+\dp2+3\baselineskip\relax}%
  \setlength{\sideleft}{\dimexpr\pagegoal-\pagetotal\relax}%
  \ifdim\sideleft>\sideneed\typeout{SIDEFIG fits #4 left=\the\sideleft\space need=\the\sideneed\space goal=\the\pagegoal\space total=\the\pagetotal}%
  \else\typeout{SIDEFIG NOFIT #4 left=\the\sideleft\space need=\the\sideneed\space goal=\the\pagegoal\space total=\the\pagetotal}\fi}
\newcommand{\sidefigure}[4][0.40]{%
  \par\sidefitcheck{#1}{#2}{#3}{#4}%
  \begin{wrapfigure}{r}{#1\linewidth}\vspace{-6pt}\centering
    \includegraphics[width=\linewidth]{#2}\vspace{-4pt}
    \caption{#3}\label{#4}\vspace{-8pt}
  \end{wrapfigure}}

\usepackage{tikz}
\usetikzlibrary{arrows.meta,positioning,fit,calc,shapes.geometric}
\definecolor{deepgreen}{HTML}{0F5C4E}
\definecolor{warningred}{HTML}{A33A3A}
\tikzset{
  flow/main/.style={
    draw=brandgreen!75!black,fill=brandgreen!11,very thick,rounded corners=2pt,
    align=center,font=\small\sffamily,minimum height=9mm,text width=39mm
  },
  flow/node/.style={
    draw=brandgreen!55!black,fill=black!1,rounded corners=2pt,
    align=center,font=\small\sffamily,minimum height=9mm,text width=37mm
  },
  flow/guard/.style={
    diamond,aspect=2.1,draw=warmorange,fill=orange!9,very thick,
    align=center,font=\small\sffamily,inner sep=1.5pt
  },
  flow/blocked/.style={
    draw=warningred,fill=red!6,rounded corners=2pt,
    align=center,font=\small\sffamily,minimum height=9mm,text width=37mm
  },
  flow/arrow/.style={-{Latex[length=2.4mm]},thick,draw=brandgreen!70!black},
  flow/deny/.style={-{Latex[length=2.4mm]},thick,dashed,draw=warningred}
}

\usepackage{enumitem}
\setlist{nosep,leftmargin=*,itemsep=1pt,topsep=2pt}

\newcommand{\GreenWAM}{%
  \textcolor{brandgreen}{Green}\textcolor{brandpurple}{WAM}\xspace%
}

\newcommand{\res}[1][]{\textcolor{warmorange}{---}}
\newcommand{\figstub}[3][]{%
  \begin{figure}[H]\centering
    \fcolorbox{black!30}{black!3}{\parbox[c][30mm][c]{.9\linewidth}{\centering\sffamily\small
      \textbf{Figure placeholder: #2}\\[1mm]\footnotesize #3}}
    \caption{#2}\if\relax\detokenize{#1}\relax\else\label{#1}\fi
  \end{figure}}

\begin{document}

\thispagestyle{empty}
\begin{titlepanel}
  \begin{minipage}[t]{.76\linewidth}\raggedright
    {\fontsize{22}{26}\selectfont\sffamily\bfseries\color{reportnavy}\ReportTitle\par}
  \end{minipage}\hfill
  \begin{minipage}[t]{.2\linewidth}\raggedleft\vspace{1mm}
    \includegraphics[width=\linewidth]{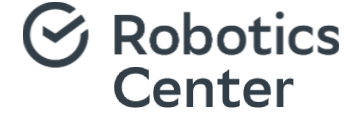}
  \end{minipage}\par\vspace{5mm}
  {\sffamily Gennady Sidorov, Valerii Kornilov, Denis Grankin, Polina Fedotova, Aleksey Postnikov\par}\vspace{0.5mm}
  {\small\sffamily\color{black!60}Brain Team, Sber Robotics Center\par}\vspace{2.5mm}
  {\small A robot needs a world model that predicts how its actions will change the scene and a policy that chooses those actions. Models are usually built and evaluated for only one of these roles, and every robot records its actions in its own format.
We ask whether one robot world model can serve both roles across different bodies, and what makes this work.
We present \GreenWAM, a world foundation model for robotics, together with the recipe that produces it: (1)~a shared anatomical action space that describes 20 robot families, from bimanual arms to dexterous humanoids, in common joint coordinates; (2)~joint training on these families in visual, action-conditioned and world--action modes; and (3)~policies that split an action expert off the generator.
As a world model, \GreenWAM attains top-tier trajectory, depth and 3D accuracy on WorldArena~1.0, ahead of every world--action model and video generator in our comparison.
As a policy, it scores among the top entries of the RoboTwin~2.0 leaderboard, reaches the highest zero-shot LIBERO-plus success among the compared methods and is competitive with dedicated vision--language--action and world--action policies on CALVIN and LIBERO.
We further adapt \GreenWAM to a real humanoid, where it succeeds more often than policies adapted from $\pi_{0.5}$ and from Cosmos~3 Edge, and distill it into a single-step policy for real-time control.
Controlled experiments yield three findings: robot specialization improves world prediction and control together; an action expert helps when split off at the start of adaptation; and keeping world-model modes during policy adaptation preserves robustness to scene changes.
\par}\vspace{2.5mm}
  {\small\sffamily \href{https://greenvla.github.io/GreenWAM}{\textbf{Project page}}\quad$\cdot$\quad\href{https://github.com/greenvla/GreenWAM}{\textbf{Code}}\quad$\cdot$\quad\href{https://huggingface.co/SberRoboticsCenter/models}{\textbf{Weights}}\quad$\cdot$\quad October 2026\par}
\end{titlepanel}
{\let\thefootnote\relax\footnotetext{\sffamily\color{black!55}\textcopyright{} 2026 Sber Robotics Center.}}
\begin{figure}[t]
  \centering
  \includegraphics[width=\linewidth]{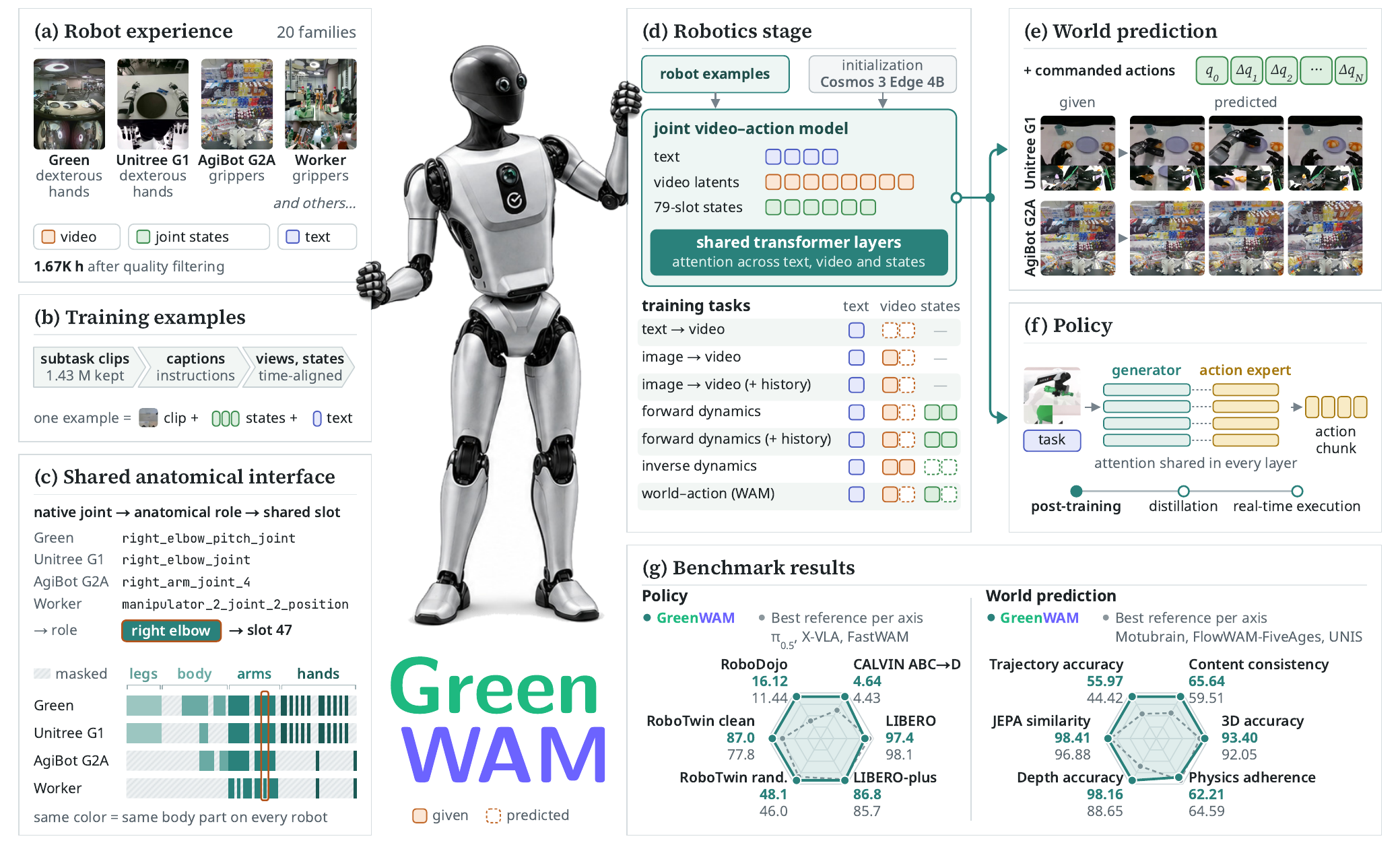}
  \caption{\textbf{From robot experience to prediction and control.} (a--b)~Video, measured states and text from 20 robot families form aligned subtask examples. (c--d)~A shared 79-slot anatomical action space connects robot motion to a joint video--action model built on Cosmos~3 Edge. (e)~Forward prediction of \GreenWAM{} for Unitree G1 and AgiBot G2A on held-out recordings (\cref{sec:exp-heldout}). (f)~As a policy, the foundation adds a coupled action expert and is distilled for humanoid control. (g)~\GreenWAM{} against the best of three references on every axis (grey dashed outline). Left: policy, against $\pi_{0.5}$~\cite{pi05}, X-VLA~\cite{xvla} and FastWAM~\cite{fastwam}; the $\pi_{0.5}$ CALVIN value is the supervised fine-tuned checkpoint reported by $\pi_{\mathrm{RL}}$~\cite{pirl}, RoboDojo scores come from the official leaderboard~\cite{robodojoboard}. Right: WorldArena~1.0 against Motubrain~\cite{motubrain}, FlowWAM-FiveAges~\cite{flowwam} and UNIS~\cite{xiaomiu0}: three categories of the motion-capped score (\cref{sec:exp-worldarena}; for the other models an upper bound) and three fidelity metrics. Every axis starts where its bar chart in \cref{fig:benchmarks-manipulation,fig:benchmarks-worldarena1} starts, but not below half of its best value, and ends at the best value on that axis.}\label{fig:intro-illustrations}
\end{figure}

\section{Introduction}\label{sec:introduction}

World foundation models learn how scenes evolve from large-scale video, and world--action models reuse this knowledge to predict observations and actions jointly~\cite{cosmos3,dreamzero}. For a robot, the two uses are linked: the model must predict how the scene changes when the robot moves, and it must produce that motion itself. Both uses therefore depend on how the model represents the robot body. Yet models are usually built and evaluated for one of these roles: world models are judged by their predictions, world--action models by the success of the resulting policy.

Robot data are fragmented across bodies and recording conventions. Joint layouts, grippers and dexterous hands, camera sets, frame rates and task boundaries differ between datasets, and the same tensor dimension can mean different motions on different robots. Current models describe a robot through an action projection for each embodiment domain, through end-effector poses with gripper or hand states, or through joint vectors padded to a common length~\cite{cosmos3,geact2}. In each case the correspondence between the arms, torso, legs and fingers of different bodies remains implicit or absent. We ask whether one robot world model can serve both as a world model and as a policy across different bodies, and what makes this work.

We present \GreenWAM, a world foundation model for robotics built on Cosmos~3 Edge~\cite{cosmos3}, together with the recipe that produces it (\cref{fig:intro-illustrations}). Cosmos~3 Edge learned robot actions only coarsely: from six embodiments dominated by a single dataset, as end-effector pose deltas and a gripper opening, with a separate projection for each embodiment domain (\cref{fig:robot-hours}). The recipe replaces this in three steps. First, a shared anatomical action space describes the bodies of 20 robot families, from bimanual arms to humanoids with dexterous hands, in common joint coordinates: a codec for each robot layout maps its joints onto shared slots for the legs, torso, arms and finger joints, so that one slot means the same joint motion on every robot (\cref{sec:interface}). Second, one model is trained jointly on these families in visual, action-conditioned and world--action modes, on recordings prepared as whole subtasks with instructions (\cref{sec:modes,sec:data}). Third, policies split an action expert off the generator (\cref{sec:action-expert}).

Both roles are obtained from this one foundation, through one action interface. A single model predicts video on held-out recordings of six robot families, with no robot-specific generator or action projection (\cref{sec:exp-gw-forward}), and a single-arm Franka, absent from the robotics corpus, is controlled through the same shared interface on CALVIN and LIBERO. As a world model, \GreenWAM attains top-tier trajectory, depth and 3D accuracy on WorldArena~1.0, ahead of every world--action model and video generator in our comparison, including Motubrain~\cite{motubrain} and FlowWAM-FiveAges~\cite{flowwam} (\cref{sec:exp-worldarena}). As a policy, it scores among the top entries of the RoboTwin~2.0 leaderboard, reaches the highest zero-shot LIBERO-plus success among the compared methods and is competitive on CALVIN and LIBERO (\cref{sec:exp-policy}). On a real humanoid, it completes 95.6\% of in-distribution instructions, against 88.9\% for a policy adapted from its parent and 84.4\% for $\pi_{0.5}$, and a distilled single-step student answers 12.3 times faster than the unoptimized policy (\cref{sec:train-distillation,sec:exp-green}).

Controlled comparisons (\cref{sec:findings}) identify what makes this work. First, robot specialization improves world prediction and control together (\cref{sec:findings-specialization}). Adapted from \GreenWAM rather than from its action-trained parent, the world model raises trajectory accuracy on WorldArena by 3.0 points and lowers its video prediction error by 23\%. The policy predicts held-out RoboTwin frames more accurately and succeeds 5.4\% more often, or 8.4\% with the action expert. Second, splitting an action expert off the generator at the start of adaptation raises policy success by 3.3\% (\cref{sec:findings-expert}). Third, keeping world-model modes during policy adaptation raises success in randomized scenes by 8.0\% (\cref{sec:findings-modes}).

\needspace{4\baselineskip}
\para{Contributions.} We make four contributions.
\begin{itemize}
  \item \textbf{Recipe.} A shared anatomical action space and a data pipeline that adapt an action-trained world foundation model to 20 robot families (\cref{sec:greenwm,sec:data}).
  \item \textbf{One model, two roles.} Joint training in seven visual, action-conditioned and world--action modes yields a world model and, through an action expert split off the generator, a policy (\cref{sec:modes,sec:train-gw,sec:action-expert,sec:train-policy}).
  \item \textbf{What it brings and why.} Results in both roles on WorldArena~1.0, RoboTwin~2.0, CALVIN, RoboDojo, LIBERO and LIBERO-plus, and controlled comparisons showing that robot specialization benefits both roles (\cref{sec:exp-gw,sec:exp-policy,sec:findings}).
  \item \textbf{Real robot.} Adaptation to a real humanoid and distillation into a single-step policy for real-time control (\cref{sec:train-distillation,sec:exp-green}).
\end{itemize}

\section{Related Work}\label{sec:related-work}

\GreenWAM builds on four lines of work: robotics specialization of world foundation models, action representations shared across robot bodies, policies derived from video generators, and temporal context for long rollouts and simulation. \Cref{tab:related} compares the closest technical reports by the capabilities of their models.

\begin{table}[!htbp]
  \centering\small\renewcommand{\arraystretch}{0.88}
  \caption{\textbf{Closest technical reports by the capabilities of the model they present.} A mark means the report presents the model in that role; it does not rank quality. \emph{Actions$\to$video}: generation conditioned on robot actions (forward dynamics, action-conditioned simulation); \emph{Video$\to$actions}: inverse dynamics; \emph{Policy}: joint prediction of video and actions used for control; \emph{Long rollout}: autoregressive continuation with memory of past blocks; \emph{Real robot}: experiments on a physical robot, for \GreenWAM on the Green humanoid (\cref{sec:exp-green}).}\label{tab:related}
  \begin{adjustbox}{max width=\linewidth}
  \begin{tabular}{@{}llcccccc@{}}
    \toprule
    Report & Group & Video gen. & Actions$\to$video & Video$\to$actions & Policy & Long rollout & Real robot \\
    \midrule
    Cosmos 3~\cite{cosmos3} & NVIDIA & \ding{51} & \ding{51} & \ding{51} & \ding{51} &  & \ding{51} \\
    DreamZero~\cite{dreamzero} & NVIDIA & \ding{51} &  &  & \ding{51} & \ding{51} & \ding{51} \\
    DreamDojo~\cite{dreamdojo} & NVIDIA & \ding{51} & \ding{51} &  &  & \ding{51} & \ding{51} \\
    DreamGen~\cite{dreamgen} & NVIDIA & \ding{51} &  & \ding{51} &  &  & \ding{51} \\
    GigaWorld-0, GigaWorld-1, GigaWorld-Policy-0.5~\cite{gigaworld0,gigaworld1,gigaworldpolicy} & GigaAI & \ding{51} & \ding{51} & \ding{51} & \ding{51} & \ding{51} & \ding{51} \\
    Motubrain, Motus2~\cite{motubrain,motus2} & ShengShu & \ding{51} & \ding{51} & \ding{51} & \ding{51} & \ding{51} & \ding{51} \\
    GE-Act 2.0, GE-Sim 2.0~\cite{geact2,gesim2} & AgiBot & \ding{51} & \ding{51} & \ding{51} & \ding{51} & \ding{51} & \ding{51} \\
    LingBot-VA, 2.0~\cite{lingbotva,lingbotva2} & Robbyant & \ding{51} & \ding{51} & \ding{51} & \ding{51} & \ding{51} & \ding{51} \\
    ABot-PhysWorld~\cite{abotphysworld} & AMAP & \ding{51} & \ding{51} &  &  &  & \\
    WALL-WM, WALL-SS~\cite{wallwm,wallss} & X Square & \ding{51} & \ding{51} &  & \ding{51} & \ding{51} & \ding{51} \\
    Xiaomi-Robotics-U0~\cite{xiaomiu0} & Xiaomi & \ding{51} & \ding{51} &  &  &  & \ding{51} \\
    1X World Model~\cite{onexwm} & 1X & \ding{51} & \ding{51} &  &  &  & \ding{51} \\
    \midrule
    \GreenWAM & Sber Robotics & \ding{51} & \ding{51} & \ding{51} & \ding{51} & \ding{51} & \ding{51} \\
    \bottomrule
  \end{tabular}
  \end{adjustbox}
\end{table}

\para{Robotics specialization of world foundations.} Cosmos 3 already combines visual generation, forward and inverse dynamics, and world--action modeling in a reasoner--generator architecture~\cite{cosmos3}. Its robot data, however, cover six embodiments, with 81\% of the hours from one dataset, and represent actions as end-effector pose deltas with a gripper opening. \GreenWAM retrains this foundation on measured joint and Cartesian trajectories of 20 robot families, aligned with subtask video and text. Genie Envisioner likewise connects a shared world foundation to prediction, control and simulation applications~\cite{genie}. Alternative data routes include human video with latent proxy actions in DreamDojo and generated robot experience in GigaWorld-0~\cite{dreamdojo,gigaworld0}. Earlier work established action-conditioned video prediction as an interactive simulator learned from diverse datasets~\cite{unisim}, and V-JEPA~2 post-trains a latent action-conditioned world model for robot planning~\cite{vjepa2}.

\para{Correspondences across robot bodies.} Cross-embodiment policies face the same correspondence problem. Open X-Embodiment pools demonstrations of 22 robots~\cite{oxe}; Octo is fine-tuned to new sensors and action spaces~\cite{octo}, and CrossFormer controls robots for manipulation, navigation, locomotion and aviation without manually aligning their observation or action spaces~\cite{crossformer}. HPT maps the proprioception and vision of each embodiment to tokens processed by a shared trunk~\cite{hpt}, X-VLA adds embodiment-specific soft prompts~\cite{xvla}, UniAct learns a universal space of atomic actions~\cite{uniact}, and RDT-1B defines a unified action space whose coordinates keep their physical meaning~\cite{rdt}. Green-VLA uses a unified, embodiment-aware action interface with which a single policy controls humanoids, mobile manipulators and fixed-base arms, and then adapts this policy to the Green humanoid~\cite{greenvla}. Among world models, Cosmos 3 represents camera and end-effector motion together with gripper or fingertip state, using an input and output projection for each embodiment domain~\cite{cosmos3}. GE-Sim~2.0 represents actions as action maps: the end-effector pose of each arm is projected into every camera image with calibrated intrinsics and extrinsics, drawn as a marker whose shading gives the gripper opening, and concatenated with the video latents~\cite{gesim2}. Such maps are natural for a video generator and simple to build, and they suit robots whose cameras are calibrated in every frame. For humanoids with moving head and wrist cameras they require exact calibration of every view, and they cannot express finger joints. \GreenWAM instead passes the measured joint and Cartesian states through one unified projector (\cref{sec:interface}) and still follows the commanded actions closely (\cref{sec:exp-action-following,sec:exp-worldarena}). \GreenWAM uses explicit slots and masks: its 79 anatomical slots cover legs, base, torso, head, arm joints and finger joints.

\sidefigure[0.38]{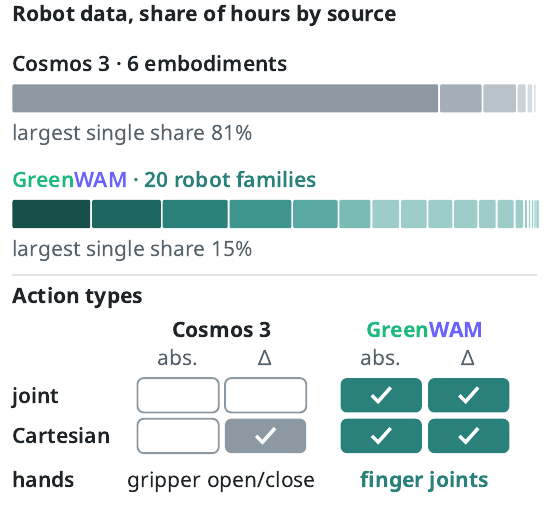}{\textbf{Robot data and action types} of Cosmos~3~\cite{cosmos3} and \GreenWAM: shares of robot hours (top; AgiBot World supplies 81\% of the robot hours of Cosmos~3, the largest \GreenWAM family 15\% of ours) and action representations (bottom).}{fig:robot-hours}
\para{From world prediction to policy.} Early approaches generate a plan as video and extract the actions from it~\cite{unipi}, or pretrain on video and fine-tune for both video and action prediction~\cite{gr2}. UVA learns a joint video--action latent with decoupled decoding heads~\cite{uva}, and UWM couples video and action diffusion so that one model acts as a policy, forward dynamics, inverse dynamics or video generator depending on the noise levels of the two streams~\cite{uwm}; the modes of \GreenWAM follow the same idea across many robot bodies. Video-model priors support control through several prediction interfaces: joint video--action generation in DreamZero, causal prediction and asynchronous execution in LingBot-VA, and visual planning followed by inverse dynamics in GE-Act~2.0~\cite{dreamzero,lingbotva,geact2}. As a policy, \GreenWAM retains joint video--action prediction and specializes the action stream through a dedicated expert.

Policies built on video models also differ in whether actions share the video backbone or receive an expert of their own, and in how that expert is initialized. Motus pretrains an action expert with latent actions, and ME-Dex inherits its video and action experts~\cite{motus,medex}. \GreenWAM copies its generator into an action expert at the split, preserving the parent function before further optimization. \Cref{sec:findings-expert} compares this split with a shared tower from the same start.

\para{Temporal context and simulation.} WALL-WM aligns captions, video and actions at the level of semantic events~\cite{wallwm}, related to the complete-subtask atoms used in our data pipeline. Genie Envisioner and GE-Sim~2.0 use sparse history for long rollouts, with GE-Sim~2.0 perturbing history during training to account for generated-context errors~\cite{genie,gesim2}. DreamZero updates its cached predictions with real observations during closed-loop execution~\cite{dreamzero}. \GreenWAM takes four anchors at fixed positions of the history, each encoded separately, and trains continuation with recorded history; \cref{fig:worldarena-rollouts} shows its rollouts of more than 20\,s on WorldArena.

World simulators also support uses beyond video prediction. GigaWorld-1 studies agreement between generated rollouts and real policy executions~\cite{gigaworld1}; GE-Sim~2.0 predicts proprioception and evaluates task completion in simulated rollouts~\cite{gesim2}. WALL-SS addresses long action-conditioned simulation with bounded memory and alignment against drift~\cite{wallss}, and the 1X World Model supports policy and checkpoint selection~\cite{onexwm}.

\section{\GreenWAM}\label{sec:greenwm}

\GreenWAM maps the states of different robots into shared anatomical coordinates and trains one generator to predict video, robot motion or both; a policy is obtained by splitting an action expert off this generator (\cref{sec:action-expert}). Layout-specific codecs absorb the differences between native robot interfaces; \cref{sec:data} describes the data and the training.

\subsection{Model Overview}\label{sec:base}

\begin{figure}[!htbp]
  \centering
  \includegraphics[width=\linewidth]{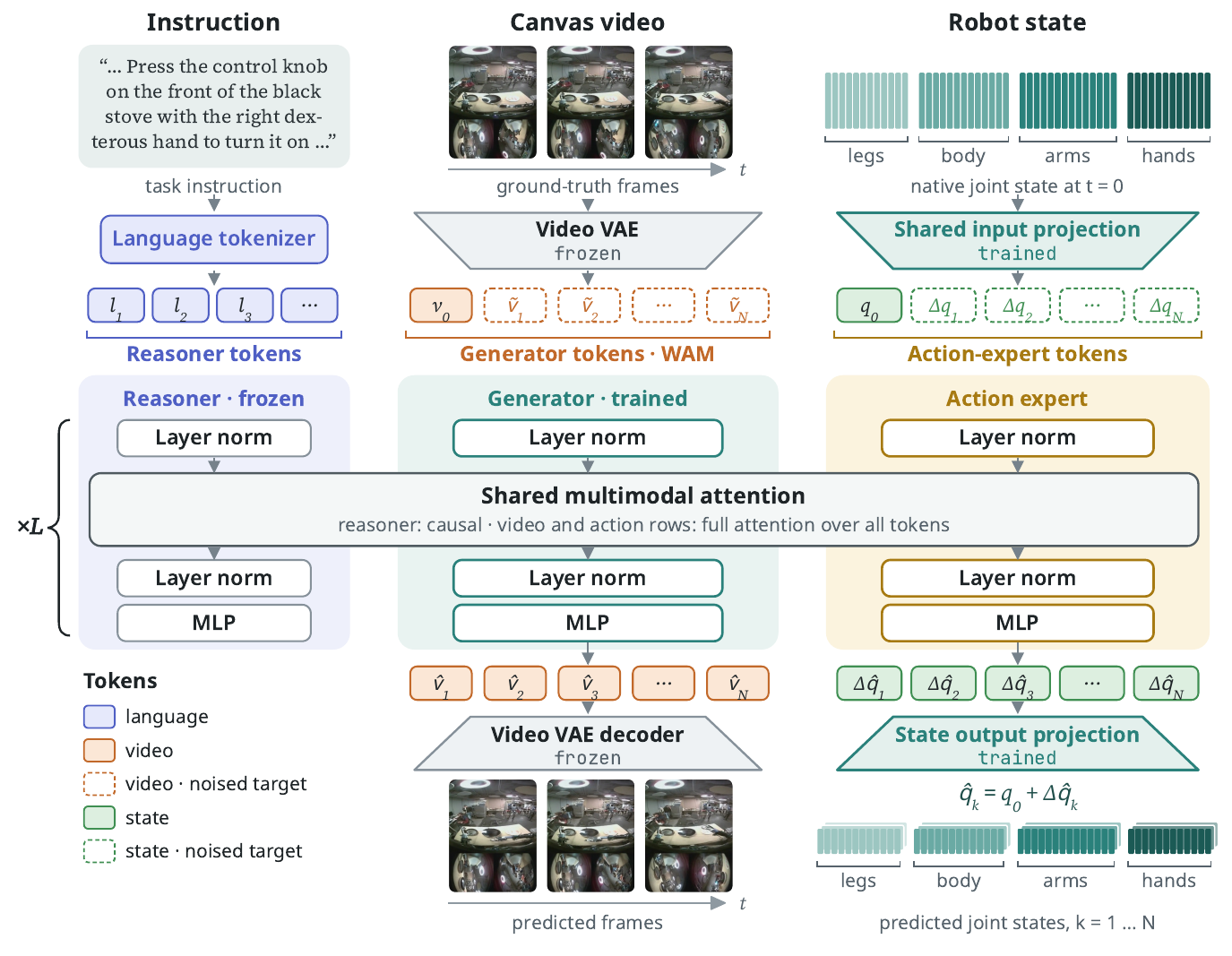}
  \caption{\textbf{\GreenWAM architecture, shown for the world--action (WAM) mode on one held-out Green humanoid episode.} The task instruction, the multi-camera canvas and the robot's native joint state at $t=0$ enter a frozen reasoner, a trained generator for the video tokens and, in the adapted policy, an action expert for the state and action tokens, all sharing multimodal attention (the Cosmos 3 Edge mixture of transformers); the layout's codec maps the native joints to the 79 anatomical slots before the shared input projection (\cref{fig:interface}). In \GreenWAM the state and action tokens pass through the generator; the action expert is split off it for policy adaptation (\cref{sec:action-expert}). Filled tokens are clean conditions and dashed tokens are noised targets; which streams are conditions changes with the mode (\cref{fig:modes}). Predicted state targets are decoded relative to $q_0$ (\cref{eq:offsets}) and mapped back to native joint states. Video: ground truth (top) and the \GreenWAM prediction (bottom). Frozen and trained refer to the robotics stage.}
  \label{fig:architecture}
\end{figure}

We initialize \GreenWAM from the released Cosmos 3 Edge 4B checkpoint~\cite{cosmos3}. Its mixture-of-transformers backbone (\cref{fig:architecture}) couples a reasoner and a generator, with a causal video VAE and rectified-flow training~\cite{rectifiedflow}. Its robot actions, however, come from six embodiments and are represented as end-effector pose deltas with a gripper opening (\cref{fig:robot-hours}); the robotics stage retrains the action modes on measured joint and Cartesian states of 20 robot families, with the joints in a shared anatomical action space.

During the robotics stage the reasoner and the VAE stay frozen. The generator, the shared action projections and the action-modality embeddings are trained. All modes retain the Cosmos 3 JSON prompt format (\cref{sec:captions}).

\subsection{Shared Anatomical Action Space}\label{sec:interface}

\begin{figure}[!htbp]
  \centering
  \includegraphics[width=\linewidth]{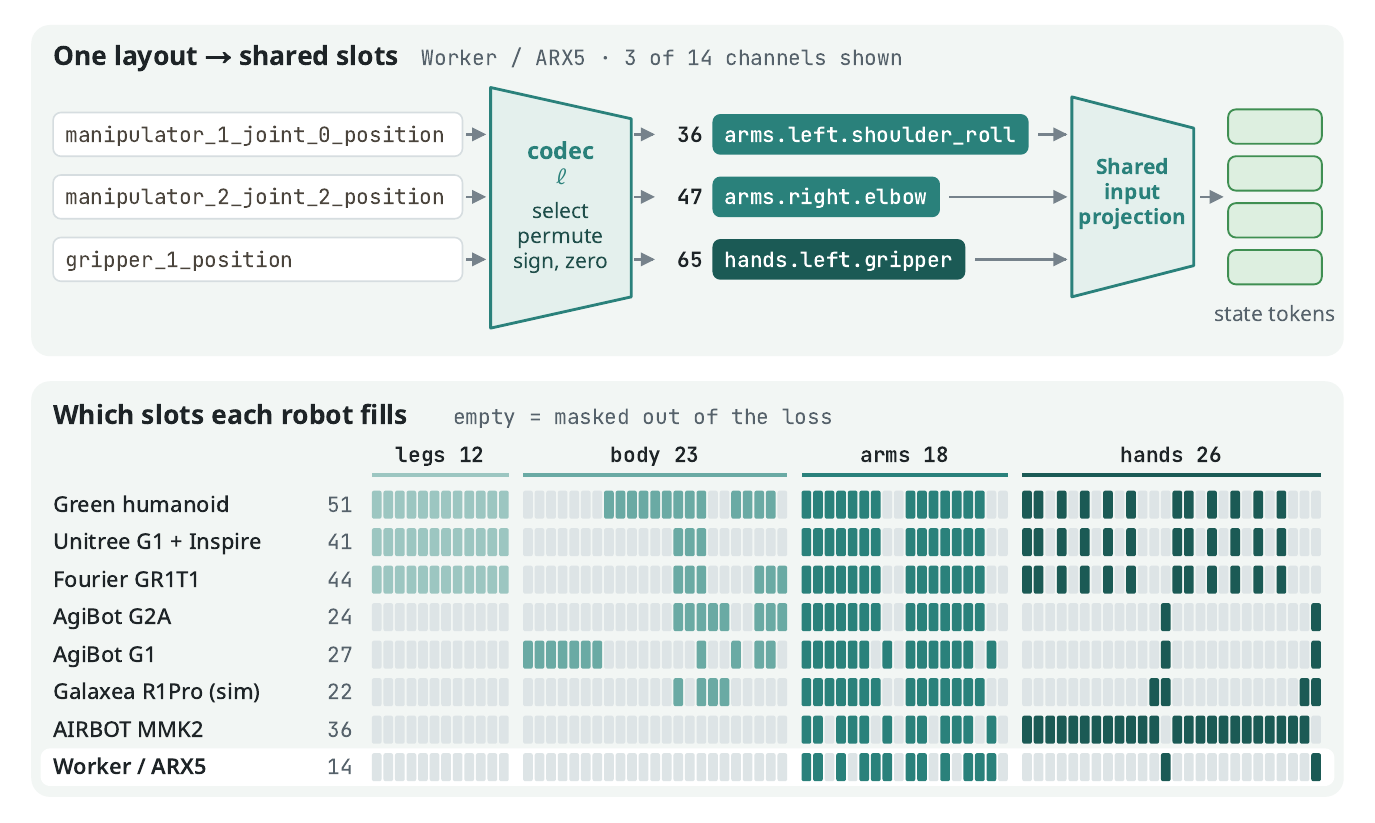}
  \caption{\textbf{Shared 79-slot anatomical action space.} Top: the codec of one layout selects its native channels, puts them in slot order and flips signs and shifts zeros where needed, so that a slot means the same joint motion on every robot, for example the same positive direction of the left shoulder roll; three of the fourteen Worker channels are shown. A shared input projection maps the slot vector to state tokens. Bottom: slots filled by eight source layouts; empty slots are masked in the loss.}
  \label{fig:interface}
\end{figure}

Robot datasets disagree on state layout, joint order, units, sign and zero conventions. Embodiment-specific projectors accommodate these formats but leave the correspondence between their input coordinates implicit. \GreenWAM establishes that correspondence before projection: layout-specific codecs map native channels to 79 anatomical coordinates, which then pass through one shared input and output projector (\cref{fig:interface}). Thus the left elbow occupies the same slot on every platform, while platform-specific conventions are handled by the codec. All 20 families therefore share one action projector, and no robot has action parameters of its own. This is a stricter design than one projection per embodiment domain, as in Cosmos~3 Edge, or end-effector poses rendered into the camera images, as in GE-Sim~2.0~\cite{gesim2}. The vocabulary has four blocks:

\begin{itemize}
  \item \textbf{Legs} (12).
  \item \textbf{Body} (23): base pose (7), base twist (6), torso (7) and head (3). Pose and twist are separate quantities.
  \item \textbf{Arms} (18).
  \item \textbf{Hands} (26): one open--closed coordinate per parallel gripper and per-finger coordinates for dexterous hands, with auxiliary slots for a second gripper finger, encoders and finger abduction.
\end{itemize}

Layout-specific codecs select, reorder and normalize native channels into anatomical slots. Validity masks exclude missing or unresolved coordinates from the loss. Correspondences are established from documented channels, robot kinematics and measured trajectories; \cref{app:anatomical-mappings} gives the details. New post-training layouts use these same slots and shared projectors. For a layout $\ell$ with native state $x$, the codec assigns slot $j$ either a native channel $\pi_\ell(j)$ with a sign $\rho_{\ell j}\in\{-1,1\}$ and a zero offset $o_{\ell j}$, or nothing:

\begin{equation}\label{eq:codec}
q_j=\begin{cases}\rho_{\ell j}\big(x_{\pi_\ell(j)}-o_{\ell j}\big), & m_{\ell j}=1,\\ 0, & m_{\ell j}=0,\end{cases}\qquad \bar q_j=\frac{q_j-\mu_{\ell j}}{s_{\ell j}},
\end{equation}
where $m_{\ell j}$ is the validity mask and $\mu_{\ell j}$, $s_{\ell j}$ are statistics of the layout's training data. Masked slots are excluded from the loss, so a coordinate that a robot does not have never receives a target.

\para{Encodings by mode.} Joint states and Cartesian end-effector (EEF) poses are two views of the robot state; absolute and relative values are two encodings. Inverse dynamics predicts the absolute trajectory $q_{0:T}$, including the initial state, from video alone. Forward dynamics, forward dynamics with history and WAM express a subset of coordinates relative to the state at the block boundary:

\begin{equation}\label{eq:offsets}
\delta_k = q_k - d_\ell\odot q_0, \qquad \hat q_k = \hat\delta_k + d_\ell\odot q_0 ,
\end{equation}
where $q_0$ is the state at the block boundary, $q_k$ the state at the $k$-th target time and $d_\ell$ a binary mask of the layout. By default leg, torso, head and arm joints are relative, while grippers, hands and fingers stay absolute; a layout can declare its whole state relative, as the Green humanoid does. The relative values are neither velocities nor accumulated increments, and $d_\ell$ is separate from the validity mask $m_\ell$, which only removes unavailable coordinates from the loss. Where a source also records a reliable end-effector pose, the joint or EEF view is chosen with equal probability after the window is sampled, so a window never gains sampling mass from having two views. EEF poses occupy separate slots with their own frames; their codec composes poses in SE(3), so the additive rule above applies to joint coordinates only. The projector performs no forward or inverse kinematics.

Families overlap mostly in the arm block, while legs, torso, head and fingers are present only for part of the corpus (\cref{fig:interface}). Every robot that has a joint contributes to training its slot, and an absent finger or leg joint never becomes a zero target that the model would have to learn as motion.

\subsection{Joint Video--Action Modeling}\label{sec:modes}

A robot clip with its states can answer more than one question: what the scene will look like, what the robot will see if it follows a given trajectory, which trajectory produced the observed video, and what to do next. Instead of training a separate model for each question, we train one set of generator weights on seven modes (\cref{tab:modes}). A mode is defined by which streams are clean conditions and which are noised targets. For a mode $m$, let $c_m$ be the clean conditions and $\mathcal{S}_m$ the generated streams, such as video latents $z$ or anatomical states $q$. All generated streams of an example share one noise level $\sigma\in[0,1]$. A clean target $x_0$ with Gaussian noise $\epsilon$ is noised along the rectified-flow path, and the generator predicts its velocity:

\begin{equation}\label{eq:flow}
x_\sigma=(1-\sigma)\,x_0+\sigma\,\epsilon,\qquad v=\epsilon-x_0,\qquad \mathcal{L}_m=\mathbb{E}_{\sigma,\epsilon}\sum_{s\in\mathcal{S}_m}\lambda_s\,\frac{\big\|M_s\odot\big(\hat v_\theta(x_\sigma,c_m,\sigma)_s-v_s\big)\big\|^2}{\|M_s\|_1},
\end{equation}
where $M_s$ masks the valid elements of stream $s$ and $\lambda_s$ weights the stream; for states the average runs over body-part groups (\cref{eq:state-loss}). The WAM mode generates future video and the state offsets of \cref{eq:offsets} from the first frame $z_0$, the measured state $q_0$ and the instruction $c$. History modes instead condition the next block on an ordered tuple $A_b$ of four independently encoded RGB anchors:

\begin{equation}\label{eq:wam-history}
\begin{gathered}
p_\theta\big(z_{1:H},\delta_{1:H}\mid z_0,q_0,c\big),\\
p_\theta\big(z_{b+1}\mid A_b,c_{b+1}\big),\qquad
p_\theta\big(z_{b+1}\mid A_b,U_{b+1},c_{b+1}\big),
\end{gathered}
\end{equation}
where the last two distributions correspond to image-to-video and forward dynamics with history, respectively, and $U_{b+1}$ contains the supplied state-trajectory tokens. The anchors may repeat when history is short; their exact positions are specified in \cref{app:temporal-sampling}.

An eligible example can be used in several modes; the admissible subsets depend on available modalities, text and outcome labels (\cref{sec:curation,app:mixture}).

\begin{table}[!htbp]
  \centering\small
  \caption{\textbf{Training modes.} Shares are sampling probabilities over examples, not over tokens or compute.}\label{tab:modes}
  \begin{tabular}{@{}>{\raggedright\arraybackslash}p{.20\linewidth}>{\raggedright\arraybackslash}p{.34\linewidth}>{\raggedright\arraybackslash}p{.28\linewidth}r@{}}
    \toprule
    Mode & Clean conditions & Targets & Share \\
    \midrule
    Text-to-video & Visual caption & Subtask video & 20\% \\
    Image-to-video & Visual caption, initial RGB & Subtask video & 10\% \\
    Image-to-video with history & Block caption, four RGB anchors from history & Next block video & 10\% \\
    Forward dynamics & Task instruction, initial RGB, future states & Subtask video & 15\% \\
    Forward dynamics with history & Block instruction, four RGB anchors, future states & Next block video & 20\% \\
    Inverse dynamics & Observed video, inverse instruction & Absolute state trajectory incl.\ $q_0$ & 10\% \\
    World--action (WAM) & Task instruction, initial RGB, measured $q_0$ & Future video and state offsets & 15\% \\
    \bottomrule
  \end{tabular}
\end{table}

The fixed mixture gives three quarters of the examples to predicting video, with or without the robot's trajectory, and a quarter to producing robot states through inverse dynamics and WAM. Among the autoregressive examples, forward dynamics with history receives twice the share of image-to-video with history.

The modes share a physical-time grid for video and state targets. Visual and dynamics modes use whole prepared subtasks; autoregressive modes continue across complete subtask blocks, while WAM predicts a short future inside a successful demonstration. The exact horizons, resampling rules and condition--target arrangement are given in \cref{app:temporal-sampling,fig:modes}.

\subsection{Multi-View Observations and Temporal Context}\label{sec:canvas}

\begin{figure}[!htbp]
  \centering
  \includegraphics[width=\linewidth]{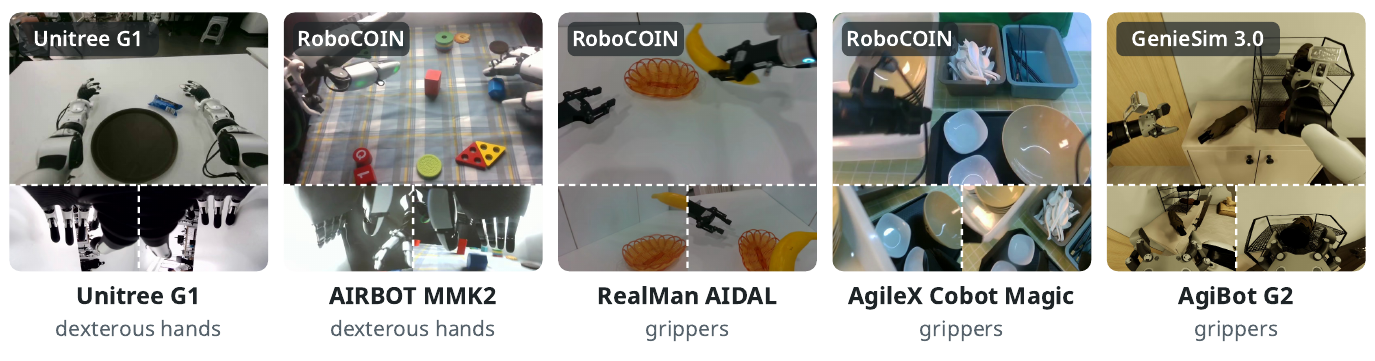}
  \caption{\textbf{Training canvases of five robot families.} From left: Unitree G1 and AIRBOT MMK2 with dexterous hands, RealMan AIDAL, AgileX Cobot Magic and AgiBot G2 with grippers; all recorded on real robots except AgiBot G2, which comes from simulation. The head camera occupies the top two thirds of the canvas, and the left and right wrist cameras share the bottom row (dashed lines); each view is resized to its panel without cropping. All five use the 640$\times$640 canvas, the most frequent of the five canvas sizes.}
  \label{fig:canvases}
\end{figure}

Our corpus contains episodes with either one camera or three synchronized cameras. For three-view sources, we compose the head and two wrist views offline into one video per contiguous kept segment (\cref{fig:canvases}). The head view occupies approximately the top two thirds, with the wrist views below in a fixed Left\,|\,Right order. Choosing among five 480p canvas buckets preserves the head-camera aspect ratio and keeps the aspect-ratio change of every panel within 0.94--1.15; single-camera sources use the full frame. Camera roles are verified per source by correlating view motion with left- and right-arm motion. Canvas sizes and temporal sampling are specified in \cref{app:temporal-sampling}.

The autoregressive modes condition each new block on sparse visual history, as in Genie Envisioner and GE-Sim~2.0~\cite{genie,gesim2}. As in Genie Envisioner, the generator receives four RGB anchors; unlike its random draws during training, ours sit at the same fixed positions in training and at inference: the start, one third, two thirds and end of the available history on the effective video grid, so the last anchor is always the most recent frame. Each anchor is encoded separately by the VAE; at the start of an episode the initial frame is repeated. Training uses ground-truth history (teacher forcing); at inference the history is the generated video. History may cross neighboring kept subtasks of the same episode and camera, and it is reset at rejected gaps and episode boundaries. Each block carries its own caption or instruction.

\section{Robot Experience and Training}\label{sec:data}\label{sec:training}

Robot datasets differ in frame rate, camera layout, state semantics and task boundaries, yet every training example must align a robot's motion with the video and language of the same interval. We cut recordings into subtask clips, filter unusable ones and derive separate text for visual prediction and for action learning, and then train the robotics stage. Application-specific adaptations are described with their evaluations in \cref{sec:exp-worldarena,sec:exp-policy}.

\subsection{Cross-Embodiment Corpus}\label{sec:corpus}

\para{Families.} The training data span 20 main robot families (\cref{fig:corpus-families}), all two-armed: bimanual platforms with parallel grippers and humanoids with grippers or dexterous hands, in real and simulated scenes. \Cref{app:data} lists the layouts and the subsets admitted to training.

They combine public and internal robot datasets in LeRobot format~\cite{lerobot} with our humanoid data. The training unit is an \emph{atom}: a subtask clip from one episode, with synchronized views, measured states, text and provenance. \Cref{app:data} gives the admission rules and the counts of each stage.

\subsection{Data Curation and Training Mixture}\label{sec:curation}

\begin{figure}[!b]
  \centering
  \includegraphics[width=\linewidth]{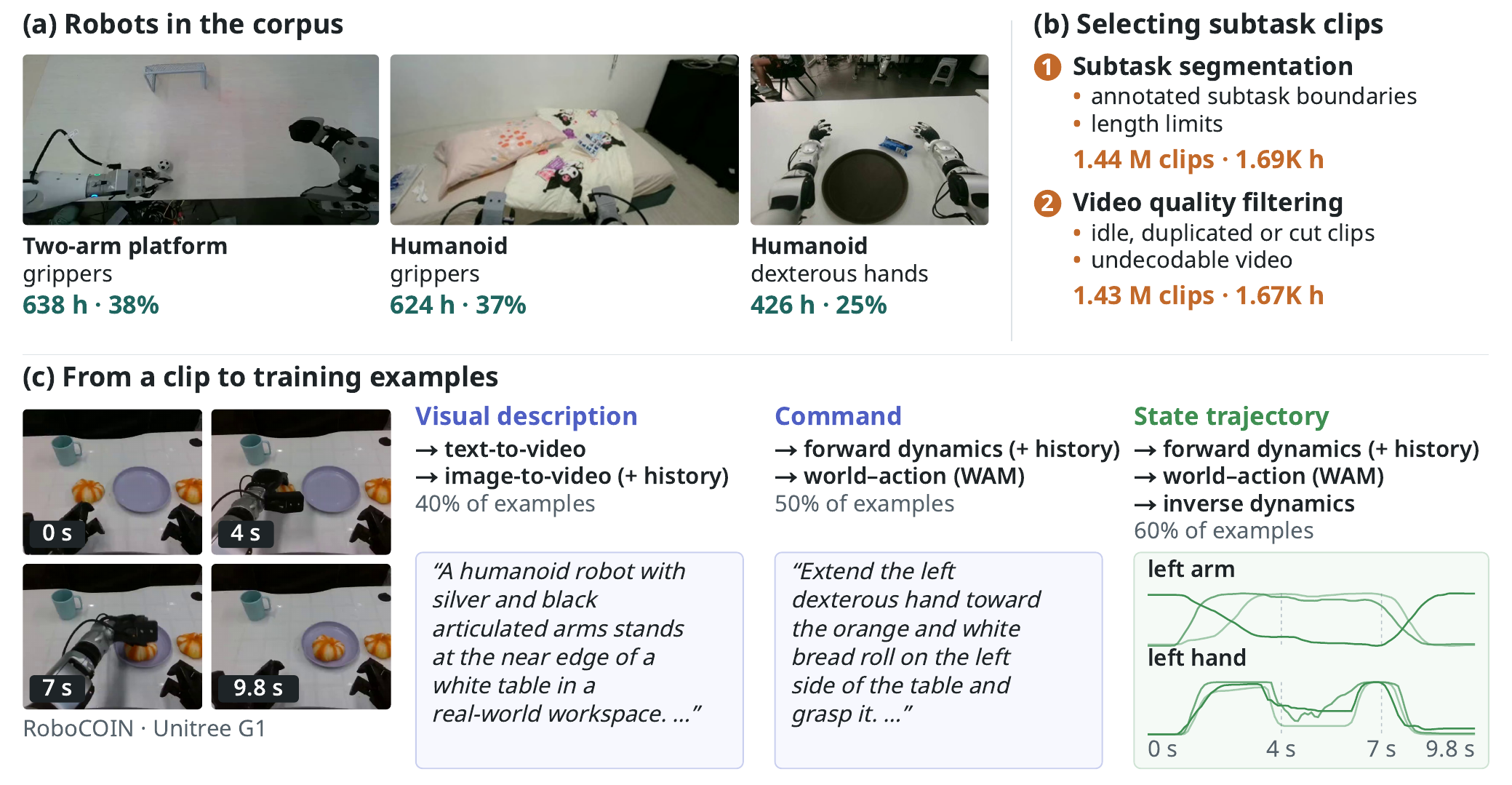}
  \caption{\textbf{Robot data.} (a)~The corpus: 20 robot families, 1\,688.5\,h after admission, 74\% real, all two-armed, in three embodiment groups; each group is shown by a head-camera frame whose width is proportional to its hours. (b)~Episodes are cut into subtask clips (atoms) at annotated subtask boundaries within length limits (\cref{eq:length}) (1.44 million clips, 1\,688.5\,h); video-quality filtering removes idle, duplicated, cut and undecodable clips (1.43 million clips, 1\,674.7\,h). (c)~One clip (RoboCOIN, Unitree G1 with dexterous hands, real), shown at the boundaries of its action phases. Its visual description and derived command are quoted in part, verbatim; its recorded state trajectory shows the three joints that move most of the left arm and of the left hand, each scaled to its own range. Arrows name the training modes that use each output, with their shares of examples (\cref{tab:modes}); the state trajectory serves all three action modes, as a condition of forward dynamics and as the target of world--action prediction and inverse dynamics, so the shares overlap.}
  \label{fig:corpus-data}
\end{figure}

Public robot datasets come as whole episodes, each with its own labels, camera set and quality. Sampling fixed-length windows from them would place unrelated subtasks in one example and pair each window with a label written for a different interval. We therefore prepare the data in stages, each adding one decision without rewriting earlier ones (\cref{fig:corpus-data}). Registration fixes the camera roles, state layout and robot properties of each source, and cutting into atoms yields the admitted corpus of 1.44 million whole-subtask atoms (1\,688.5\,h); the quality pass keeps 1.43 million atoms (1\,674.7\,h); captioning and command derivation give 1.19 million atoms an accepted command.

\para{Admission.}\label{sec:funnel} We first resolve semantic subtask boundaries from the source annotations. An episode is used whole only when it has no finer temporal annotation; ambiguous or incomplete boundaries are excluded. A length limit rejects overlong atoms without cutting them into arbitrary windows (\cref{fig:corpus-data}b). A separate quality pass removes idle intervals, duplicated frame runs, temporal cuts and unreadable clips (\cref{fig:corpus-data}b). Passing the video-quality check does not by itself make an atom available to every training mode: the text-dependent modes also need an accepted caption or command (\cref{app:admission,tab:dataqa}).

\begin{figure}[!htb]
  \centering
  \includegraphics[width=\linewidth]{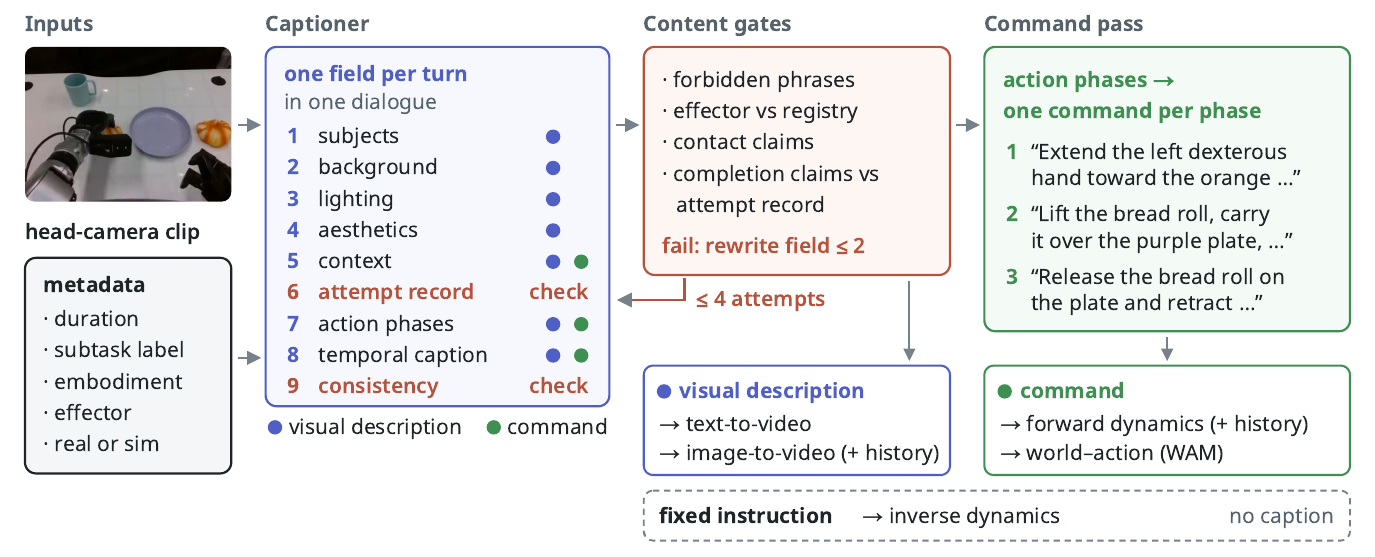}
  \caption{\textbf{Captioner and command pass.} The inputs are the head-camera clip of \cref{fig:corpus-data}c with its duration, subtask label, embodiment, effector and domain. The captioner (Qwen3.8-27B~\cite{qwen38}) writes nine fields in one dialogue, one field per turn; the attempt record and the consistency field (red) are checks and never enter a training prompt. Blue dots mark the fields that form the visual description (fields from subjects through context, action phases and the temporal caption); green dots mark the fields the command pass reads (context, action phases and the temporal caption). Content gates test forbidden phrases, the effector against the registry, and contact and completion claims against the attempt record; a failing field is rewritten at most twice, and the whole dialogue is attempted at most four times. The command pass turns the action phases into one command per phase, keeping all phases in order; shown are the opening words of the three commands derived for the input clip. Bottom: the visual description conditions text-to-video and image-to-video, the command conditions forward dynamics and world--action prediction, and inverse dynamics receives a fixed instruction.}\label{fig:captioner}
\end{figure}

\para{From captions to instructions.}\label{sec:captions} The text pipeline produces a visual description and a prospective command for each atom (\cref{fig:corpus-data}c and \cref{fig:captioner}). A video captioner (Qwen3.8-27B~\cite{qwen38}) writes a structured record one field per turn in a single dialogue, so each field is conditioned on those already written: subjects (the robot and its effector), background, lighting, aesthetics, context, an attempt record, timed action phases, a temporal caption and a consistency check. The source task label supplies the intended goal, and registry facts fix the embodiment, effector type and real or simulated domain. Before the action phases, the captioner records how far the attempt got, reading it from the final frame rather than from the label; later fields must end where the clip ends, and a caption whose motion does not match the object and kind of action named by the label is rejected. A second, text-only stage converts the action phases into a prospective command for the same fragment. It preserves the observed motions, object references and phase order, including partial or failed attempts, rather than completing an intended task that is not demonstrated. Using the source label as the command would instead pair an unfinished attempt with a goal that the clip never reaches. The conversion does not see native dataset instructions. Visual modes use the visual description; forward dynamics and WAM use the derived command; inverse dynamics uses a fixed trajectory-recovery instruction. \Cref{app:caption-details} details the captioner, the conversion, their checks and the final prompt format.

\para{Aligned training inputs.} Camera selection and ordering are explicit for each source. Synchronized RGB views are assembled into one canvas while retaining the original episode timestamps for states and text. Frame-rate reduction selects corresponding video and state times; it does not rescale physical joint positions or displacements. Autoregressive examples join complete adjacent atoms and stop at rejected gaps. WAM additionally requires a positive success label; failed, unknown and unlabeled outcomes are excluded from WAM but can enter other eligible modes. Its short prediction window retains the command of the enclosing atom, so the instruction describes the ongoing subtask rather than a task that must finish within that window.

\para{State-derived targets.}\label{sec:actions}
Recorded \texttt{action} columns mix joint targets, velocities, teleoperation master states and progress signals. We instead construct a common trajectory from measured states with the codec of \cref{eq:codec}:

\begin{equation}\label{eq:state-target}
q_k = \mathrm{codec}_{\ell}\big(s(t_0+\Delta_k)\big),
\end{equation}
where $s$ is the recorded state, $\Delta_k$ is a physical time offset within one admitted interval and $\mathrm{codec}_{\ell}$ maps layout $\ell$ to the shared anatomical slots. Inverse dynamics predicts absolute $q_k$; forward dynamics and WAM use the targets $\delta_k$ of \cref{eq:offsets}. The targets therefore describe realized robot motion. Executing a predicted trajectory additionally requires the robot-specific decoder of \cref{sec:train-policy}.

Normalization statistics are computed on the training split only (\cref{app:anatomical-mappings}).

\para{Training mixture.}\label{sec:mixture} Sampling chooses a mode, a robot cluster (\cref{eq:family-weight}), a dataset and a window. The Worker/ARX5 and Green humanoid clusters receive fixed shares; other clusters and datasets are weighted by the square root of their admissible subtask counts. Each source is sampled only in the modes its data support. Full sampling targets are given in \cref{app:mixture,tab:clusters}.

\subsection{\GreenWAM Training}\label{sec:train-gw}

\para{Objective.} The robotics stage trains all seven modes with the rectified-flow objective (\cref{eq:flow}). The generator predicts the velocity of each noised target stream. Video loss averages over valid target elements. State loss first averages within each represented body-part group and then across groups, so a group with more coordinates does not automatically receive more weight. Let $e_{tj}=\hat v_{tj}-v_{tj}$ be the velocity error of the state targets in \cref{eq:state-target}, $m_{tj}$ select valid noised targets, and $\mathcal{G}$ contain the body-part groups with at least one such target. Then

\begin{equation}\label{eq:state-loss}
\mathcal{L}_{\mathrm{state}} = \frac{1}{|\mathcal{G}|}\sum_{g\in\mathcal{G}}
\frac{\sum_t\sum_{j\in g} m_{tj}\,e_{tj}^{2}}
     {\sum_t\sum_{j\in g} m_{tj}}, \qquad
\mathcal{L} = \lambda\,\big(\mathcal{L}_{\mathrm{video}} + \mathcal{L}_{\mathrm{state}}\big).
\end{equation}

All generated streams share the noise level $\sigma$ of \cref{eq:flow}, drawn from the noise schedule of Cosmos~3 Edge (\cref{app:recipes}). The loss carries no noise-level weight, and both terms are scaled by $\lambda=10$, as in Cosmos~3 Edge, so video and state carry equal weight. Losses are averaged over examples, and a stream without targets contributes zero. The anatomical groups are legs, body, arms and hands; unresolved and absent coordinates are masked. Clean conditions, including the measured $q_0$ in WAM, carry no loss; in inverse dynamics the initial state is a target. The RoboTwin policy recipe instead weights valid action coordinates equally (\cref{app:policy-recipes}). Text conditioning is dropped with probability 0.1 for classifier-free guidance~\cite{cfg}.

With the reasoner and VAE frozen, training runs on 64 H100 GPUs with a global batch of 256 examples. Parameters are sharded within each node of 8 GPUs and replicated across nodes and attention is checkpointed selectively. The full optimizer, schedule and sampling recipe are given in \cref{tab:gw-recipe,app:recipes}.

\para{Training scale.} The robotics stage produces \GreenWAM after 36\,500 optimizer steps, corresponding to 9.34 million sampled examples. This checkpoint initializes downstream adaptations. The learning-rate schedule and checkpoint settings are given in \cref{tab:gw-recipe}. \Cref{sec:exp-gw,sec:exp-policy} evaluate this checkpoint as a world model and as a policy.

\section{\GreenWAM as a World Model}\label{sec:exp-gw}

\Cref{sec:exp-heldout} tests the robotics checkpoint itself, without any benchmark adaptation, on held-out recordings of robots from the training corpus: whether one model predicts video for different bodies and whether the generated motion follows the commanded trajectory. \Cref{sec:exp-worldarena} adapts the checkpoint to WorldArena~1.0, a public benchmark of action-conditioned video prediction, and compares it with the models on its leaderboard. All results in this section use the generator alone, without the action expert of the policies.

\subsection{The Robotics Checkpoint on Held-Out Recordings}\label{sec:exp-heldout}\label{sec:exp-setup}\label{sec:experiments}\label{sec:exp-gw-forward}

We take one held-out recording for each of six robot families: AgiBot G2A, AgileX, Unitree G1, Fourier GR1, Worker/ARX5 and the Green humanoid, the last both in simulation and on the real robot. Given the first frame and the recorded joint trajectory, the model generates the video of the whole recording (\cref{fig:intro-illustrations}e).

Unless stated otherwise, sampling uses 35 steps, classifier-free guidance 6 and shift 5, and comparisons between checkpoints keep the sampler fixed. Protocols and evaluation details are given in \cref{app:protocols,sec:findings-eval}.

\begin{figure}[!htb]
  \centering
  \includegraphics[width=\linewidth]{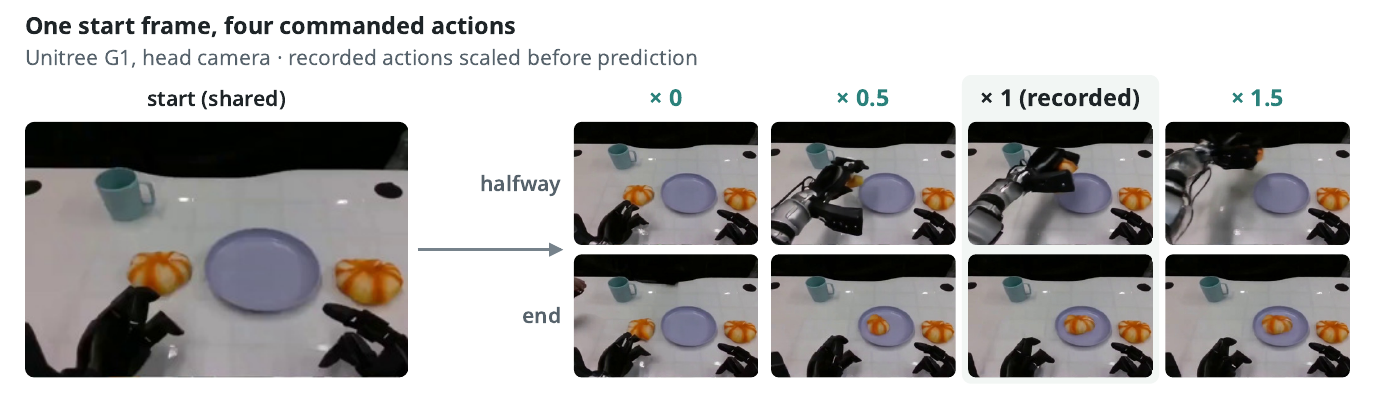}
  \caption{\textbf{Action following.} One start frame of Unitree G1 (head camera) predicted forward with the recorded state offsets scaled by 0, 0.5, 1 and 1.5; each column shows the halfway point (top) and the end (bottom) of the prediction. With zero offsets the bread stays on the table; with the other three scales it ends on the plate, and at the halfway point the larger the scale, the further the arm has moved.}\label{fig:action-following}
\end{figure}

\para{Action following.}\label{sec:exp-action-following} A world model is useful for control only if its video follows the commanded actions instead of continuing the scene on its own. For each of these recordings we keep the first frame and multiply the commanded displacement from the starting pose by 0, 0.5, 1 and 1.5: zero commands the robot to stand still, and 1 replays the recording (\cref{fig:action-following}). On every robot the generated motion grows with the command. With zero displacement the video keeps less than a fifth of the motion it shows under the recorded command, and at one and a half times the displacement it shows about a quarter more; motion is the mean change between consecutive frames. Trajectory accuracy is measured on WorldArena (\cref{sec:exp-worldarena}).

\subsection{WorldArena Adaptation and Results}\label{sec:exp-worldarena}\label{sec:train-worldarena}
WorldArena~1.0 is a public benchmark of embodied video generation; its track~1 scores 15 metrics grouped into visual quality, motion quality, content consistency, physics adherence, 3D accuracy and controllability~\cite{worldarena}. We adapt \GreenWAM on the training data that the benchmark allows and predict its test episodes from their recorded actions. \GreenWAM is strongest where the prediction must match the recorded motion and geometry: among the compared world--action models and video generators it has the highest trajectory, depth and 3D accuracy (56.0, 98.2 and 93.4), and among all 127 entries of the board it is second in JEPA similarity (98.41, after BLM at 98.49).

\begin{figure}[!t]
  \centering
  \includegraphics[width=\linewidth]{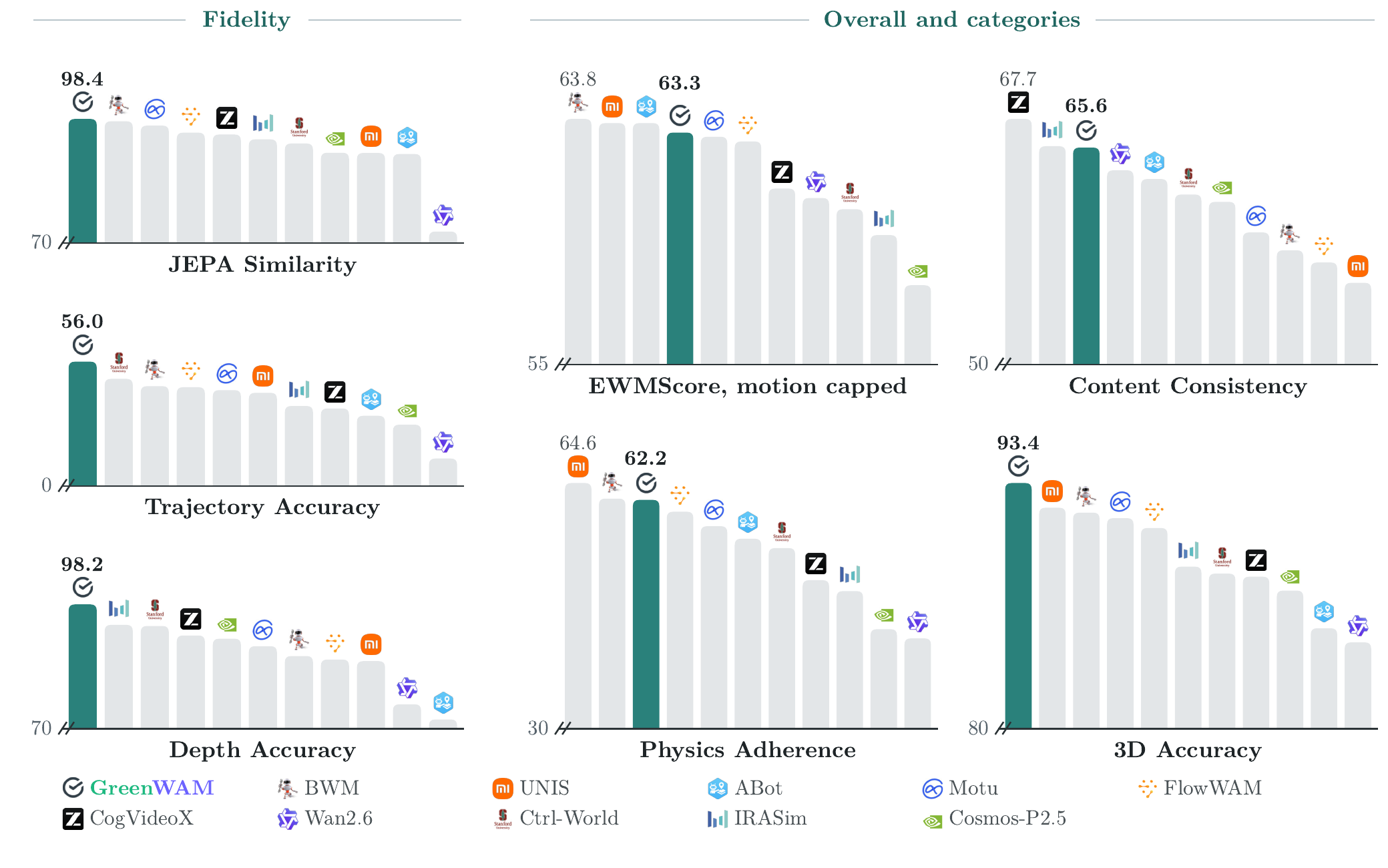}
  \caption{\textbf{WorldArena 1.0, track 1: \GreenWAM among models of the leaderboard.} Fidelity: agreement with the ground-truth video. JEPA similarity compares the distributions of V-JEPA features of real and generated videos at the set level (JEDi~\cite{jedi}; higher is better); trajectory and depth accuracy are computed per video. Overall and categories: EWMScore (\cref{eq:ewmscore}), the mean of the 15 board metrics with dynamic degree, flow score and motion smoothness capped at the ground truth, and three of its categories. \GreenWAM is capped per video on all 1\,000 test episodes; for the other models the caps are applied to their leaderboard means~\cite{worldarenaboard}, which gives them an upper bound. Controllability, visual quality and motion quality also enter EWMScore and are listed with all categories in \cref{tab:worldarena-categories}. Cut marks identify axes that do not start at zero.}\label{fig:benchmarks-worldarena1}
\end{figure}

\begin{figure}[!htb]
  \centering
  \includegraphics[width=\linewidth]{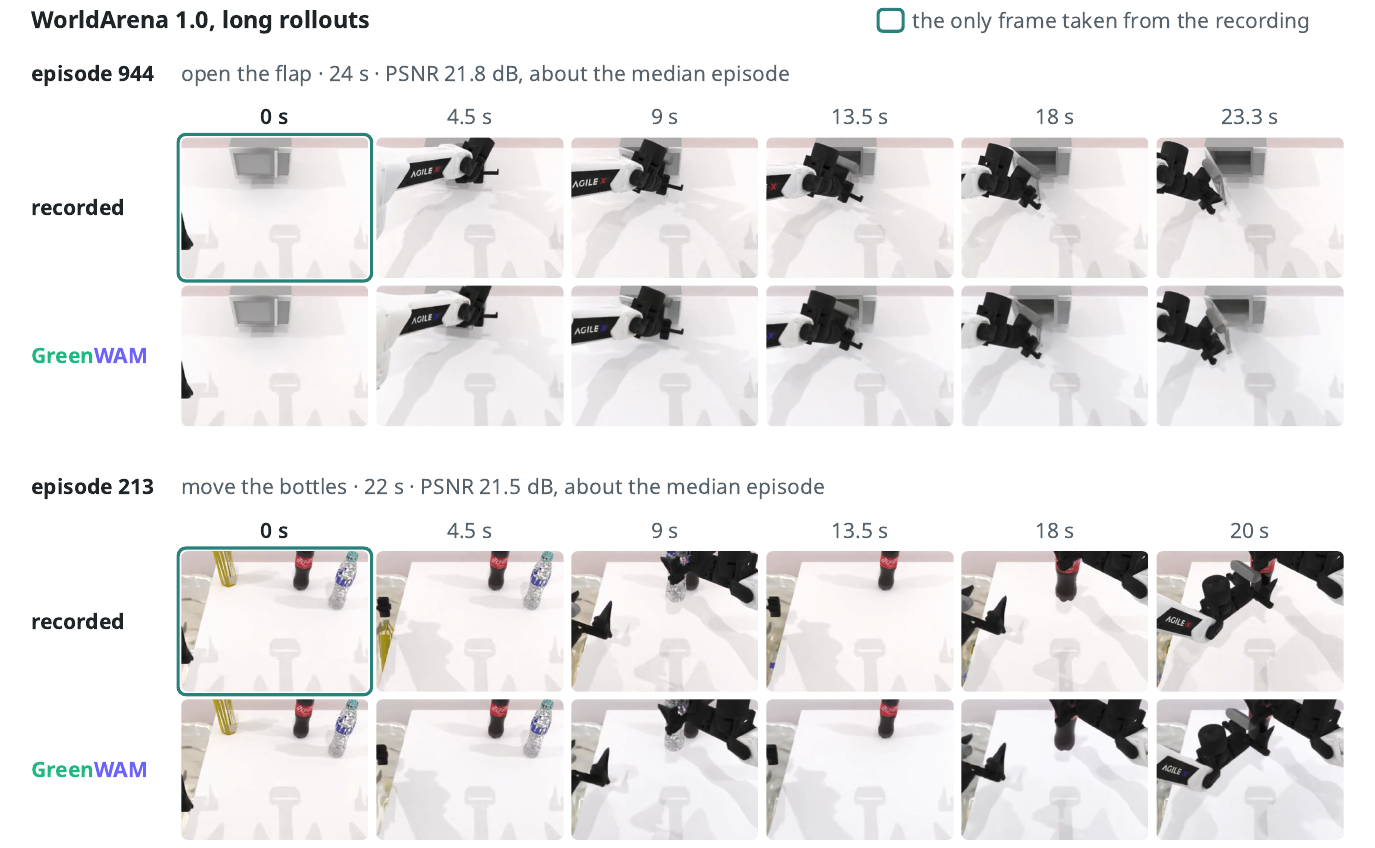}
  \caption{\textbf{Long WorldArena rollouts.} Two test episodes of WorldArena~1.0, ``open the flap'' and ``move the bottles'', each over 20\,s, with the recording above the prediction of \GreenWAM at six moments. Only the first frame, outlined, comes from the recording; the model then follows the recorded actions in blocks of about 7\,s, each conditioned on four of its own generated frames.}\label{fig:worldarena-rollouts}
\end{figure}

The action variant of track~1 gives the first frame and the robot actions of a held-out episode and asks for the video. \Cref{fig:worldarena-rollouts} shows two such episodes. This variant maps onto forward dynamics with history: every block is conditioned on the recorded actions and on four anchors of the history, which for the first block repeat the recorded first frame (\cref{app:policy-recipes}). The text prompt is empty, so sampling uses 35 steps without guidance (\cref{sec:findings-eval}). The benchmark's embodiment has the ARX5 arms of RoboTwin and reuses its layout extension, so the benchmark's actions enter through the shared projection of \GreenWAM rather than an action projection of their own (\cref{sec:interface}).

The overall score averages the board's 15 metrics after capping the three motion metrics at the ground truth, so that a model cannot gain by moving more than the recorded robot:

\begin{equation}\label{eq:ewmscore}
\mathrm{EWMScore}=\frac{1}{15}\sum_{k=1}^{15}\tilde m_k,\qquad \tilde m_k=\begin{cases}\min\big(m_k,\,m_k^{\mathrm{GT}}\big), & k\in\mathcal{K}_{\mathrm{motion}},\\ m_k & \text{otherwise},\end{cases}
\end{equation}
where $\mathcal{K}_{\mathrm{motion}}$ contains dynamic degree, flow score and motion smoothness. The cap is needed because these metrics reward motion beyond the recording; the organizers of WorldArena~2.0 found that they reach abnormally high values without better modeling of the dynamics and adopted the same rule~\cite{worldarena2motion}. Of the 127 entries of the WorldArena~1.0 board, 112 exceed the recorded-video reference in dynamic degree and 113 in flow score, the leading ones by up to four and eight times. Under the cap the scores of the leading entries drop by 6 to 11, UNIS for example from 73.64 to 63.65, while \GreenWAM, whose motion scores stay below the reference, loses 0.05.

Under this aggregate \GreenWAM reaches 63.31 on the 1\,000 test episodes (\cref{fig:benchmarks-worldarena1}), above the published world--action models Motubrain~\cite{motubrain} (63.16) and FlowWAM-FiveAges~\cite{flowwam} (62.99), whose recomputed scores are upper bounds. Scores of other models are recomputed from their leaderboard entries under the same motion cap, so they can differ from the values quoted in their reports.

\para{The recorded videos as a reference.} We also score the real test videos with the same evaluator. They reach 67.23, the score of a model that would reproduce every recording exactly, and on 13 of the 15 metrics \GreenWAM differs from them by less than four. The two categories in which it trails, visual quality and controllability, do not compare a video with its recording: visual quality rates image appearance, and controllability is assigned by a vision--language judge. On these categories many board entries score above the real videos themselves, so they follow image appearance more than agreement with the recorded episode (\cref{app:public-benchmarks}).

\section{\GreenWAM as a Policy}\label{sec:exp-policy}

We adapt \GreenWAM to control by predicting a future state trajectory together with video. \Cref{sec:action-expert,sec:train-policy} describe how a policy is obtained: the action expert split off the generator and the post-training common to all benchmarks, while each benchmark keeps its own demonstrations, command interface and inference settings. \Cref{sec:exp-simulation} reports the four simulation benchmarks. Real-time control on the humanoid benefits from a faster policy, so \cref{sec:train-distillation} distills the policy into a one-step student and measures its latency, and \cref{sec:exp-green} evaluates the teacher and the student on the robot.

\subsection{Policy Architecture}\label{sec:action-expert}

In the robotics stage of \GreenWAM, video and action tokens pass through one generator. For policy adaptation, action tokens move to a dedicated transformer while video tokens remain in the generator. Joint attention connects the two streams at every layer, so action prediction retains access to visual dynamics while the streams acquire separate parameters during adaptation. Related expert architectures differ in how the expert is initialized~\cite{motus,medex}.

Let $h^{z}$ and $h^{q}$ be the video and action tokens entering a layer. The generator $G$ projects the video tokens and the expert $E$ the action tokens, attention runs over both streams together, and each stream keeps its own feed-forward block:

\begin{equation}\label{eq:expert}
(Q,K,V)^{z}=W_G\,h^{z},\quad (Q,K,V)^{q}=W_E\,h^{q},\qquad \big[\tilde h^{z};\tilde h^{q}\big]=\mathrm{Attn}\big([Q^{z};Q^{q}],[K^{z};K^{q}],[V^{z};V^{q}]\big).
\end{equation}

We initialize the action expert by copying the generator tower, $\theta_E\leftarrow\theta_G$. The expert has the same shape as the parent tower, so the split does not change the predictions at initialization. The towers then diverge during post-training. The expert adds as many parameters as the generator tower and about 8\% to the time of a training step. \Cref{sec:findings-expert} measures what the split brings.

\subsection{Policy Post-Training}\label{sec:train-policy}

Policy adaptation connects joint video--state prediction to a target robot's control loop. Each benchmark specifies its demonstrations, camera layout and command interface; the common learning objective is the WAM mode of \GreenWAM (\cref{eq:wam-history}; recipes in \cref{app:policy-recipes}).

\para{Shared recipe.} The policy receives a task instruction, the current camera canvas and the measured state $q_0$, and predicts future video together with state offsets over horizon $H$. Training uses demonstrations of the target robot, with measured future states as supervision. At inference, the predicted targets are decoded relative to $q_0$ (\cref{eq:offsets}) into the robot's native command representation. The controller executes a prefix of the prediction, acquires a new observation and replans. The decoder and the execution interval are specific to each application.

\para{Appearance augmentation.} Clean demonstrations show one appearance of each scene, and benchmarks that randomize scenes test others. Two augmentations address this without touching the geometry. Background replacement: in half of the samples the static background (table, walls, floor), segmented by replaying the demonstration in the simulator, is replaced by a procedural texture modulated by the original shading, so that shadows are preserved; it needs the simulator for segmentation and is used on RoboTwin. Color augmentation perturbs the colors of each camera image independently; it is applied to 80\% of the samples and adds no crops or rotations. \Cref{sec:findings-augmentation} measures what background replacement brings.

The per-benchmark recipes, camera layouts, data volumes and execution settings are given in \cref{app:policy-recipes}.

\begin{table}[!htbp]
  \centering\small
  \caption{\textbf{RoboTwin 2.0, 50 tasks.} Average success rate, \%.}\label{tab:robotwin}
  \begin{tabular*}{\linewidth}{@{\extracolsep{\fill}}lccc@{}}
    \toprule
    Method & Clean & Randomized & Average \\
    \midrule
    \multicolumn{4}{l}{\textit{Leaderboard: 50 clean demonstrations per task}} \\
    ME-Dex-1.0~\cite{medex}$^{\dagger}$ & 89.6 & 68.1 & 78.8 \\
    OLA-Sem & 75.1 & 67.6 & 71.3 \\
    GigaBrain-0.7~\cite{gigabrain07} & 66.8 & 67.9 & 67.3 \\
    WorldScape Policy 2.0~\cite{worldscape} & 84.9 & 38.5 & 61.7 \\
    4D-WAM~\cite{fourdwam} & 81.5 & 41.8 & 61.6 \\
    $\pi_{0.5}$~\cite{pi05} & 70.7 & 46.0 & 58.4 \\
    OLA-Geo & 83.0 & 33.6 & 58.3 \\
    Spatial Forcing~\cite{spatialforcing} & 77.2 & 26.7 & 52.0 \\
    X-WAM$^{\dagger}$~\cite{xwam} & 70.0 & 25.8 & 47.9 \\
    X-VLA~\cite{xvla} & 68.0 & 20.9 & 44.5 \\
    ABot-M0~\cite{abotm0} & 57.4 & 30.4 & 43.9 \\
    EventVLA~\cite{eventvla} & 65.6 & 15.7 & 40.6 \\
    Xiaomi-Robotics-0~\cite{xiaomir0} & 62.9 & 18.2 & 40.5 \\
    Fast-WAM~\cite{fastwam} & 77.8 & 1.9 & 39.9 \\
    GalaxeaVLA~\cite{galaxea} & 62.7 & 12.7 & 37.7 \\
    AHA-WAM~\cite{ahawam} & 64.3 & 3.2 & 33.8 \\
    starVLA~\cite{starvla} & 46.5 & 3.2 & 24.8 \\
    \midrule
    \textbf{\GreenWAM}, leaderboard protocol & \textbf{86.98} & \textbf{48.10} & \textbf{67.54} \\
    \midrule
    \multicolumn{4}{l}{\textit{Outside the leaderboard: clean demonstrations only, own evaluation}} \\
    Qwen-RobotManip-Context~\cite{qwenmanip} & 84.7 & 69.4 & 77.05 \\
    \midrule
    \multicolumn{4}{l}{\textit{Published: clean and randomized demonstrations}} \\
    $\pi_0$~\cite{pi0} & 65.92 & 58.40 & 62.16 \\
    $\pi_{0.5}$~\cite{pi05} & 82.74 & 76.76 & 79.75 \\
    X-VLA~\cite{xvla} & 72.80 & 72.84 & 72.82 \\
    DeMaVLA~\cite{demavla} & 88.42 & 86.78 & 87.60 \\
    Motus~\cite{motus} & 88.66 & 87.02 & 87.84 \\
    Fast-WAM~\cite{fastwam} & 91.88 & 91.78 & 91.83 \\
    DSWAM~\cite{dswam} & 92.38 & 91.90 & 92.14 \\
    LingBot-VA~\cite{lingbotva} & 92.9 & 91.6 & 92.25 \\
    LingBot-VA 2.0~\cite{lingbotva2} & 93.8 & 93.4 & 93.6 \\
    Motubrain~\cite{motubrain} & 95.8 & 96.1 & 95.95 \\
    \bottomrule
  \end{tabular*}
\tablenote{Upper block: the official leaderboard, which trains on 50 clean demonstrations per task (co-train track, ordered by the leaderboard average of clean and randomized success; the single-task baselines $\pi_0$, DP3, RDT, ACT and DP, at most 31.4 on average, are omitted)~\cite{robotwinboard}; $^{\dagger}$: the checkpoint that post-training starts from was trained on randomized RoboTwin scenes. The reports of GigaBrain-0.7, WorldScape Policy~2.0 and OLA-Geo, and of Qwen-RobotManip outside the leaderboard, do not state whether their pretraining included randomized RoboTwin scenes; for the other entries of these two blocks it was not. Outside the leaderboard: a report that also trains on clean demonstrations only but uses its own evaluation: Qwen-RobotManip-Context with joint control~\cite{qwenmanip} (Table~5, easy and hard settings). Lower block: published results whose training adds 500 randomized demonstrations per task (stated for Motus, Fast-WAM, LingBot-VA, LingBot-VA~2.0 and Motubrain), as reported in DSWAM~\cite{dswam} (Table~1) for $\pi_0$, $\pi_{0.5}$, DeMaVLA, Motus, Fast-WAM and DSWAM; in Motus~\cite{motus} (Table~2) for X-VLA; in the LingBot-VA repository~\cite{lingbotvarepo} for LingBot-VA; in the reports of LingBot-VA 2.0~\cite{lingbotva2} (Table~1) and Motubrain~\cite{motubrain}. \GreenWAM follows the leaderboard protocol and is listed under its block.}
\end{table}

\begin{figure}[!htbp]
  \centering
  \includegraphics[width=\linewidth]{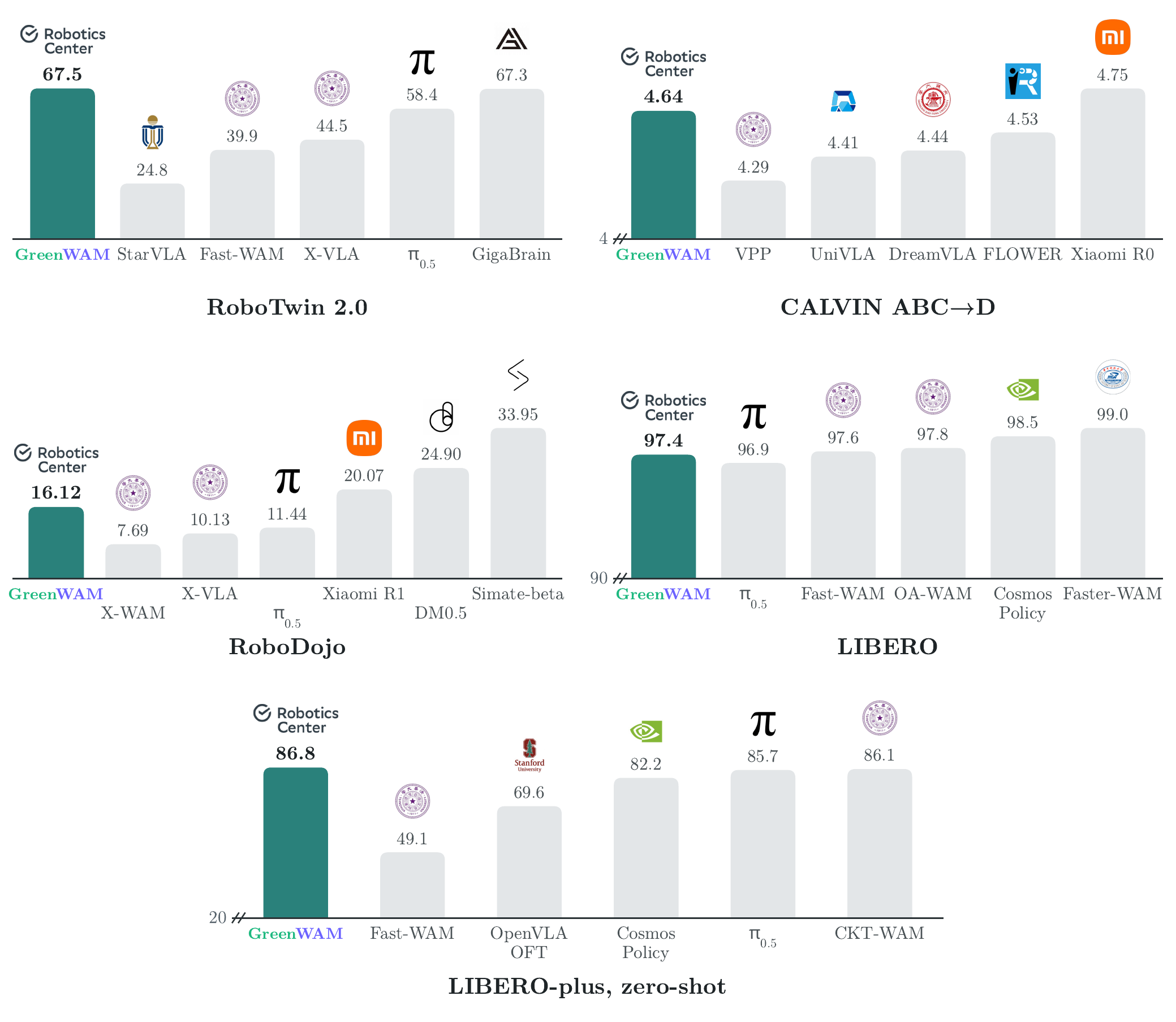}
  \caption{\textbf{Manipulation benchmarks.} \GreenWAM (first bar of each panel) and selected baselines from \cref{tab:robotwin,tab:calvin,tab:robodojo} and from published LIBERO~\cite{libero} and LIBERO-plus~\cite{liberoplus} results. RoboTwin reports mean clean/randomized success over 50 tasks with 100 trials per task; pretraining on randomized RoboTwin scenes is marked in \cref{tab:robotwin}. LIBERO reports the average success over its four task suites and LIBERO-plus the zero-shot success over all perturbed episodes; \GreenWAM values on both come from the same policy (\cref{tab:libero}). LIBERO and LIBERO-plus values and their sources are listed in \cref{tab:libero}. Cut marks identify axes that do not start at zero.}\label{fig:benchmarks-manipulation}
\end{figure}

\subsection{Simulation Benchmarks}\label{sec:exp-simulation}

We evaluate the adapted policies on RoboTwin~2.0, CALVIN, RoboDojo and LIBERO. The target benchmark protocols are specified in \cref{tab:protocols,app:protocols}. No RoboTwin, RoboDojo, CALVIN or LIBERO data enter the corpus of \GreenWAM. Policies use benchmark-specific denoising steps, guidance and replanning intervals (\cref{app:policy-recipes}). Results of other methods are taken from their reports or leaderboards, as noted in each table; \cref{fig:benchmarks-manipulation} summarizes the four benchmarks, and \cref{app:sim-tables} gives the full CALVIN, RoboDojo and LIBERO tables.

\subsubsection{RoboTwin 2.0}\label{sec:exp-robotwin}

\begin{figure}[t]
  \centering
  \includegraphics[width=\linewidth]{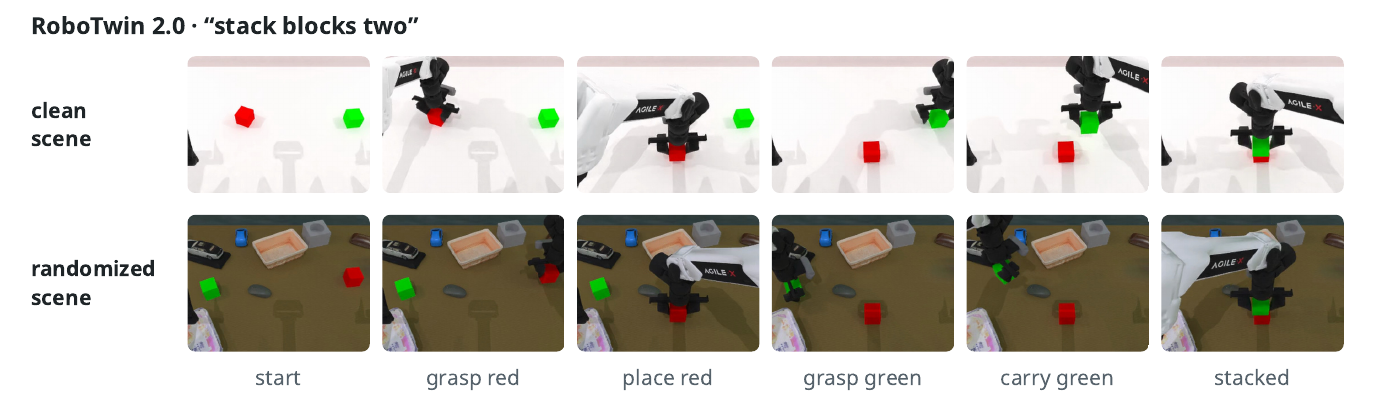}
  \caption{\textbf{RoboTwin~2.0 rollouts of the \GreenWAM policy.} The task stack\_blocks\_two in a clean (top) and a randomized scene (bottom). Head-camera frames at six phases: start, red block grasped, red block placed, green block grasped, green block carried, tower built. Both episodes succeed.}\label{fig:robotwin-rollouts}
\end{figure}

On RoboTwin~2.0, \GreenWAM scores among the top entries of the leaderboard (\cref{tab:robotwin}); \cref{fig:robotwin-rollouts} shows one task in the clean and the randomized setting. The leaderboard fixes the post-training data, 50 clean demonstrations per task, but not the pretraining.

Our RoboTwin policy reaches 86.98\% clean and 48.10\% randomized success, averaging 67.54\%, over 50 tasks with 100 trials per task in each setting. It is evaluated under the leaderboard protocol on the official seeds. Controlled comparisons at a fixed adaptation budget support both design choices: initializing from \GreenWAM and splitting off an action expert at the start of adaptation together raise clean success by 10.8\% over a single-tower policy adapted from Cosmos~3 Edge (\cref{sec:findings-specialization,sec:findings-expert}). The leaderboard trains on clean demonstrations only; \cref{sec:findings-random} examines how individual scene changes affect randomized success.

The \GreenWAM corpus contains no RoboTwin data, whereas pretraining can expose a policy to randomized RoboTwin demonstrations. ME-Dex-1.0 inherits Motus experts~\cite{medex}, and Motus and X-WAM list 27\,500 RoboTwin pretraining episodes, matching the set of 2\,500 clean and 25\,000 randomized demonstrations~\cite{motus,xwam}; WorldScape Policy~2.0 lists 35.7 thousand without specifying their split~\cite{worldscape}. \Cref{tab:robotwin} marks the two leaderboard entries whose pretraining includes randomized RoboTwin scenes.

\subsubsection{CALVIN}\label{sec:exp-calvin}

CALVIN tests whether a policy can chain language-conditioned skills in an unseen environment: each successful instruction advances a chain of five without resetting the scene, so earlier errors carry over to later instructions. We adapt on environments A, B and C, with no data from D, and evaluate on 1\,000 chains in D (\cref{tab:calvin}; protocol in \cref{app:protocols}). \GreenWAM completes 4.64 instructions per chain on average, with a standard error of about 0.03, above FLOWER (4.53).

\subsubsection{RoboDojo}\label{sec:exp-robodojo}

RoboDojo tests generalization, precision, long-horizon execution, memory and open-vocabulary instructions in simulation (\cref{tab:robodojo}). Its tasks use the same ARX5 arms as RoboTwin. The RoboDojo aggregate gives equal weight to its five capability dimensions rather than to every individual task. Trained on the official training set and compared with the official leaderboard, \GreenWAM reaches a score of 16.12 and a success rate of 10.49\% (\cref{tab:robodojo}), higher on both aggregates than $\pi_{0.5}$, X-VLA and X-WAM. The recipe is given in \cref{app:policy-recipes}.

\needspace{12\baselineskip}
\subsubsection{LIBERO and LIBERO-plus}\label{sec:exp-libero}

\sidefigure[0.37]{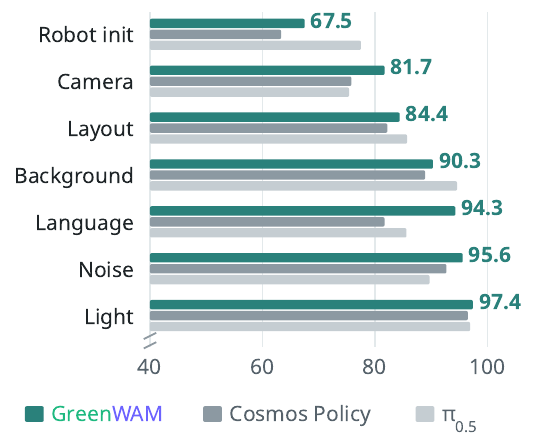}{\textbf{LIBERO-plus by axis.} Zero-shot success; baselines from~\cite{oawam}.}{fig:liberoplus-axes}
LIBERO evaluates single-arm manipulation with a Franka arm in simulation over four task suites that vary spatial layout, objects, goals and task length~\cite{libero}. Recent policies exceed 96\% average success on it. LIBERO-plus keeps the LIBERO tasks and perturbs them along seven axes, from camera viewpoint to object layout~\cite{liberoplus}. In the zero-shot setting a policy trained on standard LIBERO is evaluated on the perturbed episodes, which exposes robustness gaps that the standard suites hide: Fast-WAM~\cite{fastwam}, for example, reaches 97.6\% on LIBERO and 49.1\% on LIBERO-plus~\cite{fasterwam} (\cref{tab:libero}). The single-arm Franka embodiment is absent from the robotics corpus, whose families are all two-armed. With one policy, \GreenWAM is close to the dedicated policies on LIBERO and reaches the highest zero-shot LIBERO-plus success in \cref{tab:libero}; \cref{fig:failures} shows typical failures.

Success is highest under lighting, sensor-noise and language perturbations, where \GreenWAM leads both baselines (\cref{fig:liberoplus-axes}). Under changed viewpoints, success is 81.7\%, above both baselines. Changed backgrounds are handled without the background replacement used on RoboTwin, which needs simulator segmentation of the demonstrations.

\subsection{Distillation for Real-Time Control}\label{sec:train-distillation}

\begin{figure}[!htb]
  \centering
  \includegraphics[width=\linewidth]{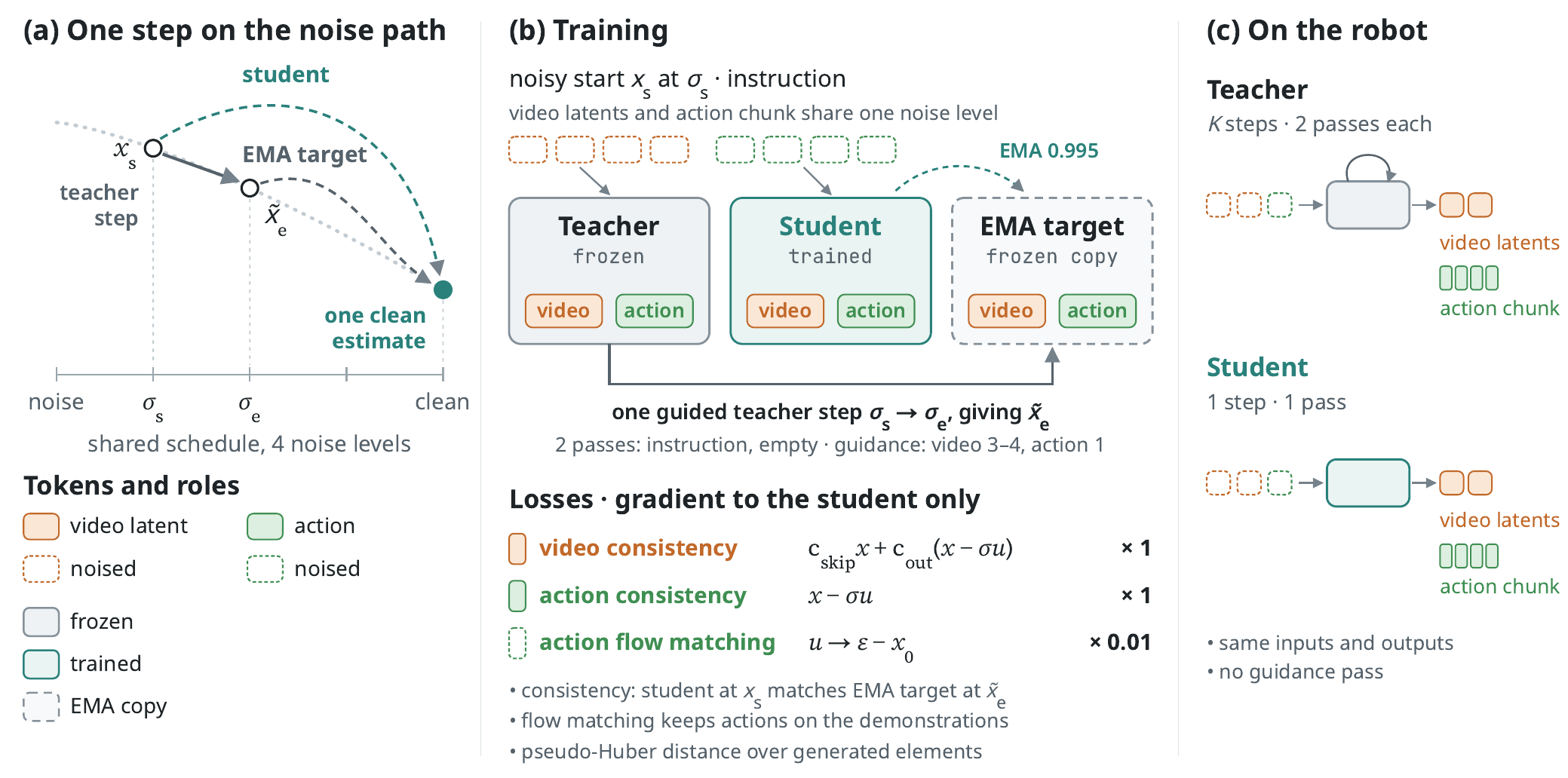}
  \caption{\textbf{Consistency distillation of the policy.} The losses are those of \cref{eq:distill}. (a)~One step on the shared noise path: the teacher steps from $x_s$ to $\tilde x_e$ between neighboring anchors of the schedule, and the student at $x_s$ and the exponential-moving-average (EMA) target at $\tilde x_e$ must give the same clean estimate. (b)~Training: a frozen teacher, the trained student and its EMA copy as the target; video and action consistency losses and an action flow-matching term, with the gradient to the student only. (c)~On the robot the student replaces the teacher's guided multi-step sampling with one step and one pass. The recipe follows Flash-WAM~\cite{flashwam}, with one noise level for both streams and an added action flow-matching term.}\label{fig:distillation}
\end{figure}

We distill the adapted policy into a one-step student with modality-specific consistency objectives, following Flash-WAM~\cite{flashwam} (\cref{fig:distillation}). The student keeps the teacher's observations, state representation and joint video--action outputs; only its sampling changes at deployment. We initialize a frozen teacher, a trainable student and an EMA target student from the adapted policy. Unlike Flash-WAM, which uses separate noise schedules for video and actions, video latents and the action chunk share one noise level. The teacher steps between neighboring levels of a shared schedule of 4 noise levels, from a noisier sample to a less noisy one. The teacher makes two passes, with the instruction and with an empty prompt, and applies guidance to video (scale 3--4) but not to actions; the student learns to produce a consistent clean prediction at both points, with the lower-noise target supplied by the EMA student and detached from the gradient.

Video and action targets use separate consistency maps. The video map uses noise-dependent skip and output coefficients, while the action map directly estimates the clean state target as $x_\sigma-\sigma\hat v_\theta$. An action flow-matching term, which Flash-WAM uses only in a regularized baseline, keeps the student anchored to the demonstration targets:
\begin{equation}\label{eq:distill}
\mathcal{L}_{\mathrm{distill}}=\lambda_v\mathcal{L}_{\mathrm{CD},v}+\lambda_a\mathcal{L}_{\mathrm{CD},a}+\lambda_{\mathrm{FM}}\mathcal{L}_{\mathrm{FM},a}.
\end{equation}
Loss distances, weights and the EMA decay are given in \cref{app:green-adaptation}.

\begin{figure}[!htb]
  \centering
  \includegraphics[width=\linewidth]{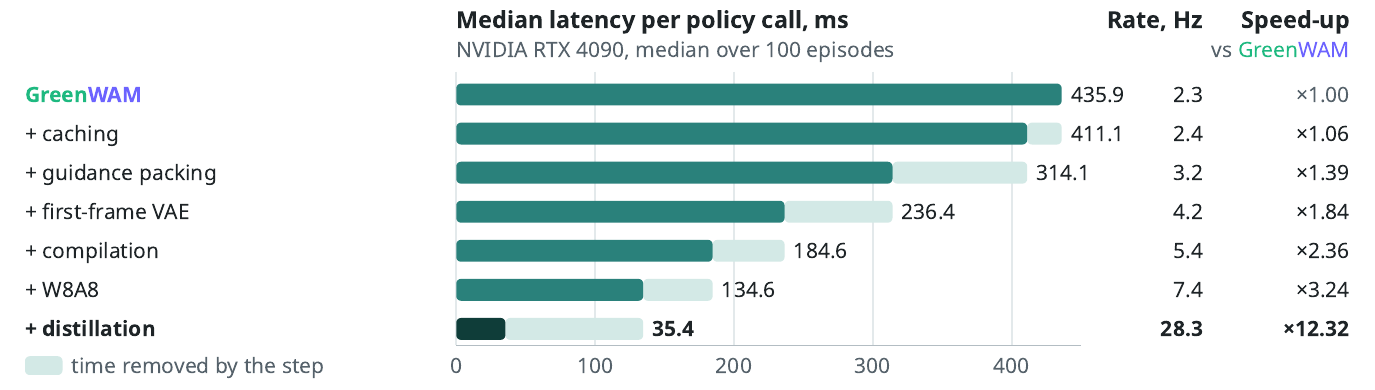}
  \caption{\textbf{Policy inference latency on one NVIDIA RTX 4090.} Median latency per policy call over 100 episodes. The optimizations are applied cumulatively to the \GreenWAM policy; the light segment of each row is the time removed by that step. Speed-ups are relative to \GreenWAM without optimizations; \cref{tab:latency} also gives the time to the first action.}\label{fig:latency-ladder}
\end{figure}

\para{Latency.} Caching the instruction and vision key--value states across denoising steps, evaluating both branches of classifier-free guidance in one pass, encoding only the conditioning frame, compiling the generator and running the feed-forward layers with 8-bit weights and activations bring the \GreenWAM policy to 134.6\,ms (\cref{fig:latency-ladder}). Distillation then replaces the four guided denoising steps with one consistency step: the distilled policy answers in 35.4\,ms, 3.8 times faster than the optimized teacher and 12.3 times faster than the unoptimized policy, at 28.3\,Hz. Distillation pays off most on the robot's onboard computer, whose compute budget is well below that of an RTX~4090; running the student there is our next step (\cref{sec:discussion}). \Cref{app:latency} describes the benchmark and each optimization; \cref{sec:exp-green} reports the task success of the distilled policy on the humanoid.

\subsection{Real-World Humanoid Evaluation}\label{sec:exp-green}

\begin{figure}[!htb]
  \centering
  \includegraphics[width=\linewidth]{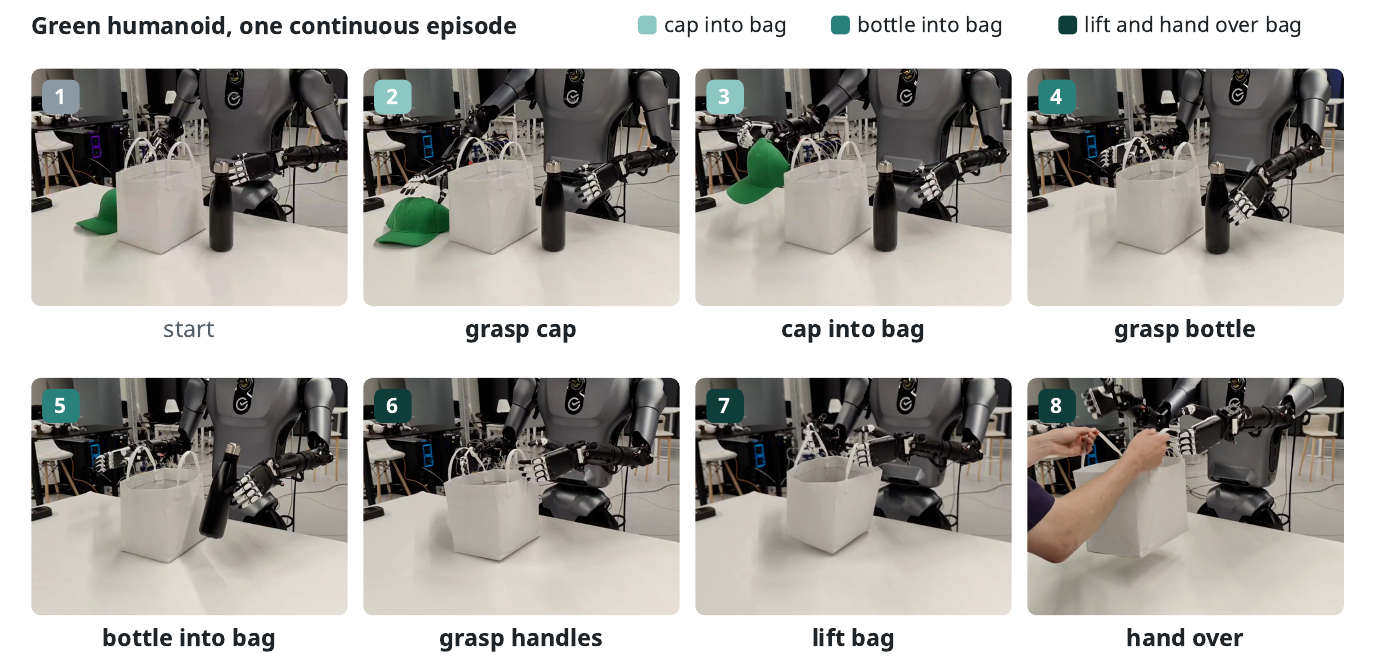}
  \caption{\textbf{\GreenWAM policy inference on the Green humanoid.} Eight frames of one continuous episode, cropped from a side camera: the robot puts a cap and a bottle into a bag, lifts the bag by its handles and hands it to a person. Colors mark the three subtasks.}\label{fig:real-humanoid}
\end{figure}

We evaluate the policies on a packing task in which the Green humanoid places a bottle and a cap into a bag, then lifts the loaded bag with both hands. For evaluation the task is given as five consecutive instructions, grasping and placing each object and then lifting the bag; each names the object and, for the single-arm actions, the hand to use, and the scene is not reset between them. Variants use different bottles and assign the pick-and-place steps to the left or right hand. We measure success at each instruction. \Cref{fig:real-humanoid} shows a packing rollout followed by a handover to a person.

We adapt $\pi_{0.5}$~\cite{pi05,openpi}, Cosmos~3 Edge~\cite{cosmos3} and \GreenWAM using the same demonstration split. We also evaluate the distilled \GreenWAM policy (\cref{sec:train-distillation}) against its teacher.

\para{Observation and control.} Every policy receives the task instruction, the current camera observations and the measured robot state. Cameras and controlled joints are the same for all policies; preprocessing and action representation follow each model's adaptation recipe (\cref{app:green-adaptation}). \GreenWAM, its distilled student and the Cosmos~3 Edge policy predict 16 actions per call, and $\pi_{0.5}$ predicts 50.

\begin{figure}[!htb]
  \centering
  \includegraphics[width=\linewidth]{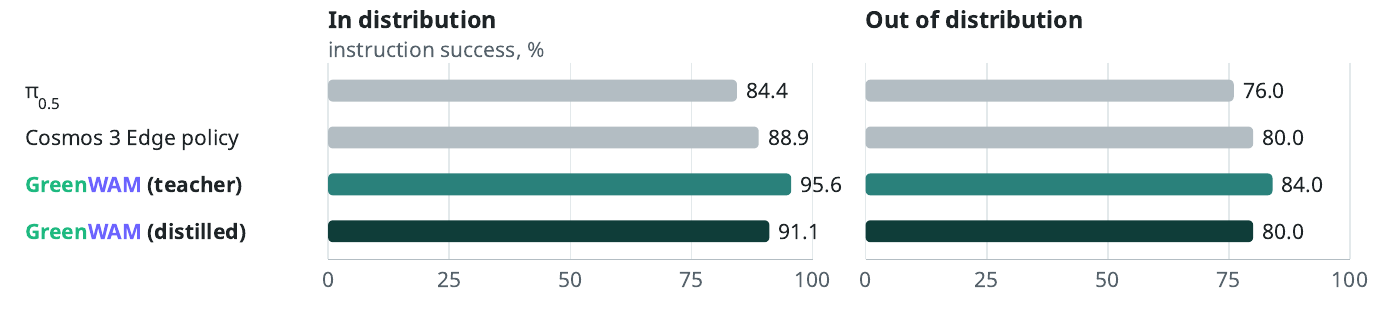}
  \caption{\textbf{Instruction success on the Green humanoid.} In-distribution and out-of-distribution success of the four policies, the fraction of successful instructions within each split (\cref{eq:success-rate}). The teacher and the distilled student run with the same execution settings.}\label{fig:humanoid-success}
\end{figure}

\para{In-distribution (ID) and out-of-distribution (OOD) evaluation.} ID trials use new resets of the objects and scenes represented in policy adaptation; OOD trials keep the task but change objects or scene conditions excluded from adaptation. Each policy is evaluated on 70 instructions in 14 scene configurations, 9 ID and 5 OOD, with one attempt per instruction; an instruction succeeds if it is completed before its timeout without human help, and success is the fraction of successful instructions in each split (\cref{fig:humanoid-success}; protocol in \cref{app:green-protocol}). \GreenWAM succeeds in 95.6\% of ID and 84.0\% of OOD trials, ahead of the Cosmos~3 Edge policy (88.9\% and 80.0\%) and $\pi_{0.5}$ (84.4\% and 76.0\%).

\para{Distilled policy.} The teacher and the distilled student run with the same observations, action horizon and replanning schedule, so their success rates are directly comparable (\cref{fig:humanoid-success}): the student reaches 91.1\% ID and 80.0\% OOD success.

Across the four simulation benchmarks and the humanoid, the same foundation and the same action interface yield policies for two-armed robots with grippers, a single-arm Franka and a humanoid with dexterous hands. \Cref{sec:findings} isolates the choices behind these results.

\section{Empirical Analysis}\label{sec:findings}

\begin{figure}[!t]
  \centering
  \includegraphics[width=\linewidth]{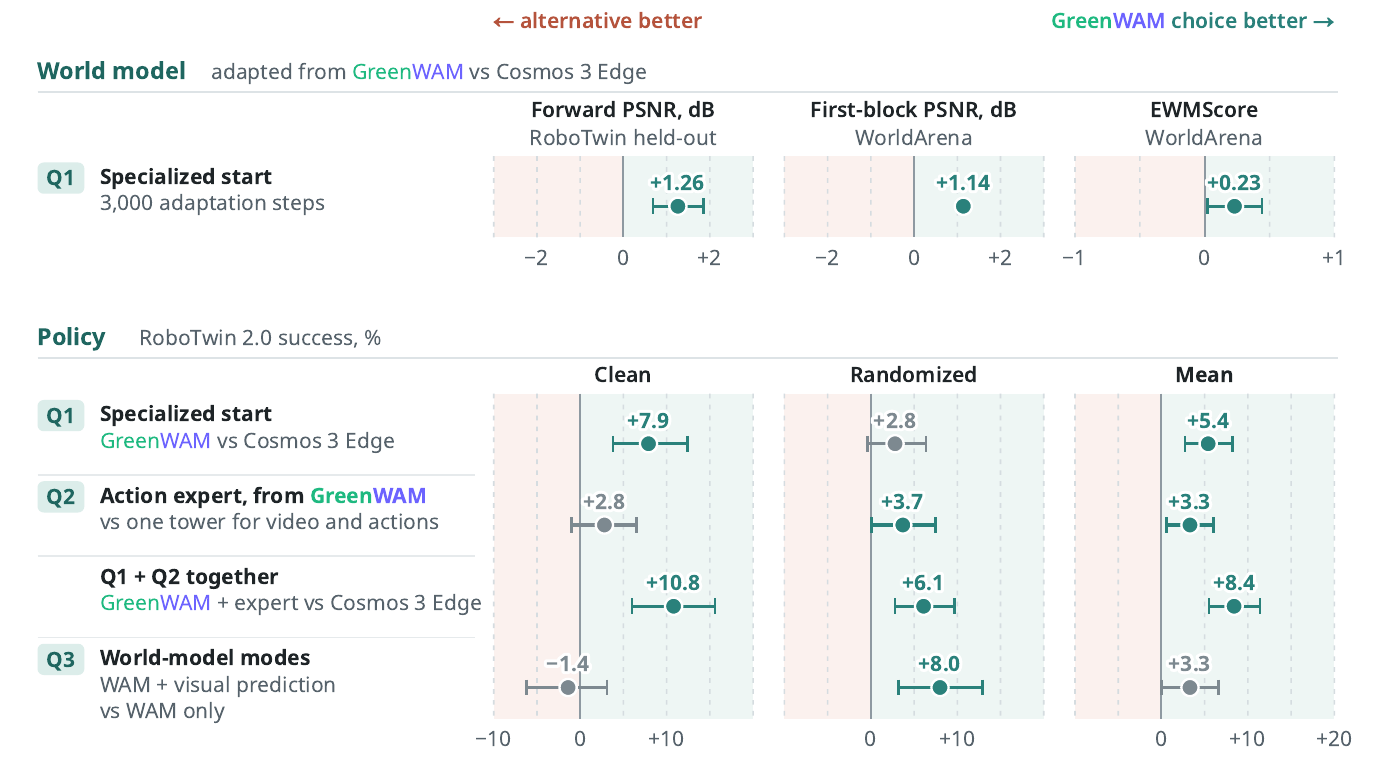}
  \caption{\textbf{Controlled comparisons at a fixed budget.} Paired differences (\cref{eq:paired-diff}) with 95\% intervals from a hierarchical bootstrap over tasks and episodes; the sign is the choice made in \GreenWAM minus the alternative, and teal marks intervals that exclude zero. World model: forward PSNR on 50 held-out RoboTwin windows, and first-block PSNR and capped EWMScore on the 1\,000 WorldArena test episodes, bootstrapped over episodes; each column has its own scale. Policy: RoboTwin~2.0 success over 50 tasks with 20 trials per task and setting. World-model modes: WAM with visual prediction, that is forward dynamics and image-to-video with weights 0.05 each next to WAM at 0.9, compared with WAM only. Q1: robot specialization, \GreenWAM against Cosmos~3 Edge; Q2: an action expert against a single tower, both from \GreenWAM; Q3: world-model modes kept during policy adaptation. The row Q1\,+\,Q2 compares \GreenWAM with an action expert split off at the start against the single-tower Cosmos~3 Edge run; each pair is estimated on the common seeds of its two runs (\cref{eq:paired-diff}), so this row need not equal the sum of Q1 and Q2.}\label{fig:ablations}
\end{figure}

We test three choices in adapting the shared foundation: robot-specialized initialization, a separate action expert and the retention of world-model modes during policy adaptation. Each comparison holds the downstream data and the adaptation budget fixed (\cref{fig:ablations}). We then examine how individual scene changes affect the policy and what appearance augmentation recovers; evaluation details are given in \cref{sec:findings-eval}.

\para{Comparisons.} The pairs are specified below; a further comparison combines the first two choices (row Q1\,+\,Q2 in \cref{fig:ablations}).
\begin{itemize}
  \item \textbf{Robot specialization.} \GreenWAM and its parent Cosmos~3 Edge are adapted with the same recipe for 3\,000 steps: to WorldArena~1.0, evaluated on its 1\,000 test episodes, and to RoboTwin~2.0 as single-tower policies, evaluated with 20 trials per task and setting; the two RoboTwin policies are also compared by forward prediction on 50 held-out clips.
  \item \textbf{Action expert.} From \GreenWAM, the single-tower RoboTwin policy is compared with one trained by the same recipe but with the action expert split off at the start; 3\,000 steps each.
  \item \textbf{World-model modes.} Both runs are warm-started from the same RoboTwin policy, trained with the full recipe: WAM at 0.9, forward dynamics and image-to-video at 0.05 each. One run then continues with this mixture, the other with WAM only; 2\,500 steps each.
\end{itemize}
Both runs of a pair are evaluated with the same evaluation code, sampler and seeds.

\para{Statistics.} Differences average over tasks, as the benchmarks do, and their 95\% intervals come from a hierarchical bootstrap over tasks and episodes (\cref{app:paired}).

\subsection{What Robot Specialization Brings to Each Role}\label{sec:findings-specialization}

Starting from \GreenWAM improves both forward prediction and policy success at a fixed adaptation budget. We adapt \GreenWAM and its parent Cosmos~3 Edge with the same recipe and budget, once as a world model and once as a policy. The parent learns its joint-space action projections from scratch, as it must for joint-space actions, while the prompt format, the WAM mode and the multi-camera canvas are native to both.

\figpair{\includegraphics[width=.78\linewidth]{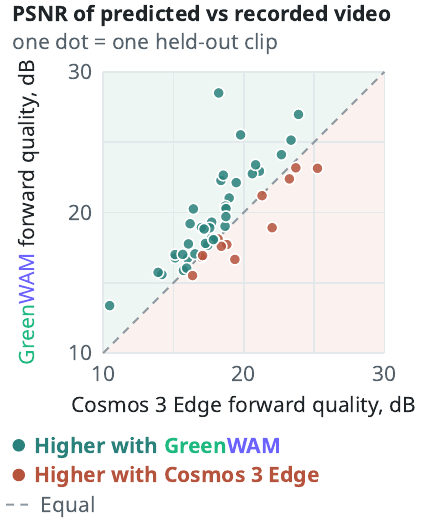}}{\textbf{Forward PSNR per held-out clip} of the policies from \GreenWAM and from Cosmos~3 Edge: 38 of 50 RoboTwin clips lie above the diagonal.}{fig:fwd-windows}{\includegraphics[width=\linewidth]{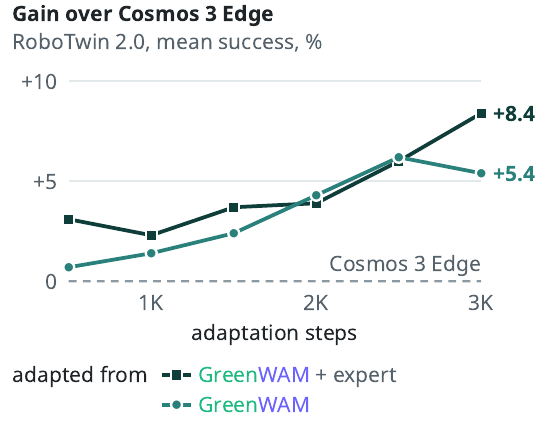}}{\textbf{Gain over Cosmos~3 Edge during adaptation.} Mean RoboTwin~2.0 success (\cref{fig:ablations}).}{fig:adapt-curve}

\para{World model.} We adapt both checkpoints to WorldArena with the recipe of \cref{sec:exp-worldarena} for 3\,000 steps and compare them on all 1\,000 test episodes. The model adapted from the robot-specialized checkpoint is closer to the recording in 87.1\% of the episodes. PSNR rises by 1.1\,dB on the first block, generated from the recorded frame alone, and by 1.2\,dB on the later blocks, which continue from the model's own frames, so the error falls by about a quarter and the gain persists through autoregressive continuation. Specialization improves the metrics on which \GreenWAM is strongest: trajectory accuracy rises by 3.0, from 44.4 to 47.4, and depth accuracy by 0.5, from 96.65 to 97.12. The aggregate score rises by 0.23, mostly from physics adherence (+1.3).

The RoboTwin policies compared below show the same effect on held-out frames. On one held-out window per task, the policy adapted from \GreenWAM predicts the next 20 frames 1.3\,dB more accurately than the one adapted from the parent and is closer to the recording in 38 of the 50 windows (\cref{fig:fwd-windows}).

\para{Policy.} Adapted to RoboTwin~2.0 with the single-tower recipe for 3\,000 steps, the policy from \GreenWAM succeeds more often than the policy from the parent: by 7.9\% in clean scenes and 5.4\% on average (\cref{fig:ablations}).

The advantage emerges early in training and persists: under 1\% after 500 steps, 6.2\% at 2\,500 and 5.4\% at 3\,000 (\cref{fig:adapt-curve}). The validation losses behave differently: they start far apart, 8.15 for \GreenWAM against 19.61 for the parent, and nearly meet by step 3\,000 (1.51 and 1.56), while the gap in closed-loop success remains. The advantage is similar against a parent trained with the WAM mode only (5.2\%). Adding the action expert to the specialized start (\cref{sec:findings-expert}) widens it to 8.4\% on average and 10.8\% in clean scenes. On the real humanoid, the comparison of policies adapted from $\pi_{0.5}$, from the parent and from \GreenWAM poses the same question (\cref{sec:exp-green}).

\subsection{Coupling the World Model to Action}\label{sec:findings-expert}

Splitting the action expert at the start of adaptation raises policy success while preserving the initial predictions. During policy adaptation, \GreenWAM moves action tokens into an expert (\cref{eq:expert}) that starts as an exact copy of the generator tower, so the split changes the trainable parameterization without changing the initial predictions (\cref{sec:action-expert}). Both runs therefore start from the same function.

Trained for 3\,000 steps from \GreenWAM with the recipe of \cref{sec:findings-specialization}, the policy with the expert succeeds 3.3\% more often on average than the one that keeps the shared generator, 2.8\% in clean and 3.7\% in randomized scenes. It raises success on 28 tasks and lowers it on 11 in clean scenes, and on 22 and 7 in randomized scenes. The gain comes with almost equal validation losses (1.503 with the expert and 1.510 without).

\needspace{18\baselineskip}
\subsection{Retaining World-Model Modes during Policy Adaptation}\label{sec:findings-modes}

\sidefigure[0.32]{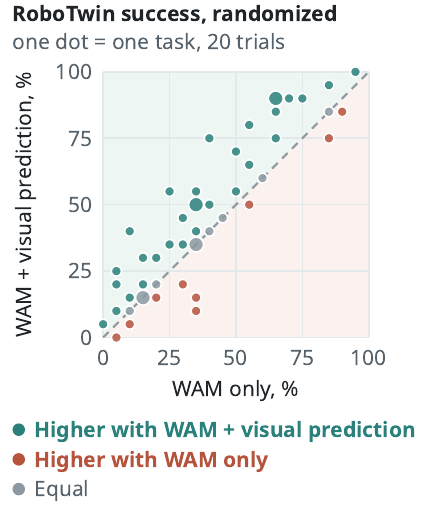}{\textbf{Per-task effect of visual prediction.} Randomized RoboTwin success per task, 20 trials, with and without forward dynamics and image-to-video.}{fig:aux-objectives}
Keeping world-model modes during policy adaptation improves robustness to scene changes. On RoboTwin, policy adaptation trains the WAM mode and keeps small shares of forward prediction and image-to-video generation (0.9, 0.05 and 0.05). Against WAM alone, trained for 2\,500 steps from the same checkpoint and evaluated identically, the additional modes raise randomized success by 8.0\%, with gains on 31 of the 50 tasks and losses on 9, and leave clean success without a significant change (\cref{fig:aux-objectives}). Both runs predict video inside the WAM mode, so the gain comes from the additional modes rather than from predicting video at all. Modes that receive a tenth of the training examples thus preserve robustness to scene changes that the WAM objective alone loses.

Together, the three comparisons show what makes the shared foundation work as a policy: robot specialization, an action expert copied from the generator and retained world-model modes each raise success at a fixed adaptation budget, the last mainly in randomized scenes, and specialization improves the world model at the same time.

\subsection{What the Policy Imagines}\label{sec:findings-imagination}

\begin{figure}[!t]
  \centering
  \includegraphics[width=\linewidth]{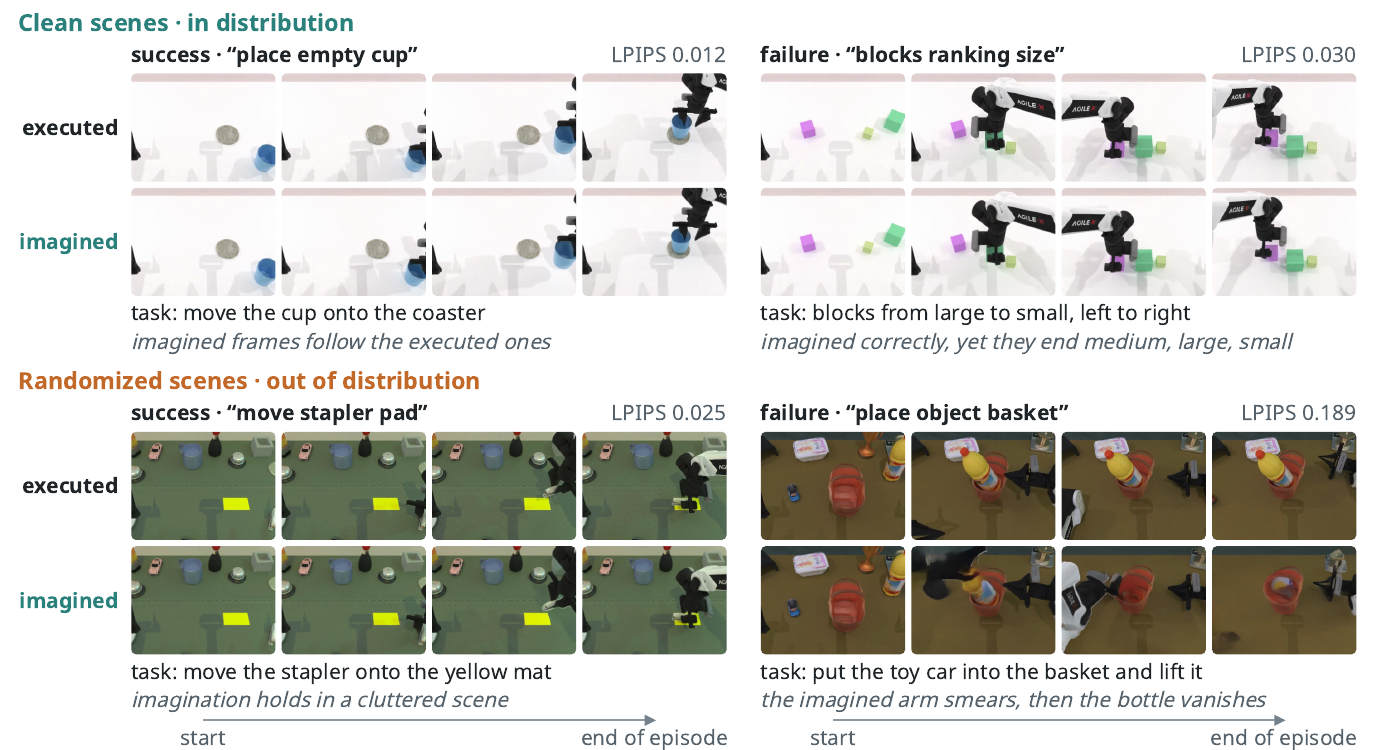}
  \caption{\textbf{What the RoboTwin policy imagines while it acts.} A success and a failure in clean scenes (in distribution) and in randomized scenes (out of distribution). For each episode, the head-camera frame rendered after an action chunk is executed (top strip) and the frame the policy predicted for the same moment in the WAM mode (bottom strip), at four moments from the start to the end of the episode; LPIPS is the mean over the episode. In the clean failure the policy imagines the outcome of its actions correctly, including the wrong order in which the blocks end; in the randomized failure the imagined scene departs from the executed one.}\label{fig:imagination}
\end{figure}

A world--action model predicts the video of each action chunk together with the actions, so, unlike a policy that predicts actions alone, it shows what it expects to happen: decoding its predictions at every replanning step gives a view of the policy's imagination (\cref{fig:imagination}). We evaluate the final RoboTwin policy on all 50 tasks in two settings: clean scenes, which match the 50 clean demonstrations per task used for adaptation (in distribution), and randomized scenes with new backgrounds, clutter, lighting and table heights, which appear neither in these demonstrations nor in the \GreenWAM corpus (out of distribution). At every replanning step we compare the predicted head-camera frames with the frames observed after the chunk is executed, by LPIPS and by PSNR against a baseline that repeats the last observed frame. In distribution the imagination is accurate: in successful clean episodes the predictions exceed this baseline by a median of 9.6\,dB of PSNR, and in the clean success of \cref{fig:imagination} the imagined frames follow the executed ones to the end of the episode.

\begin{figure}[!t]
  \centering
  \includegraphics[width=\linewidth]{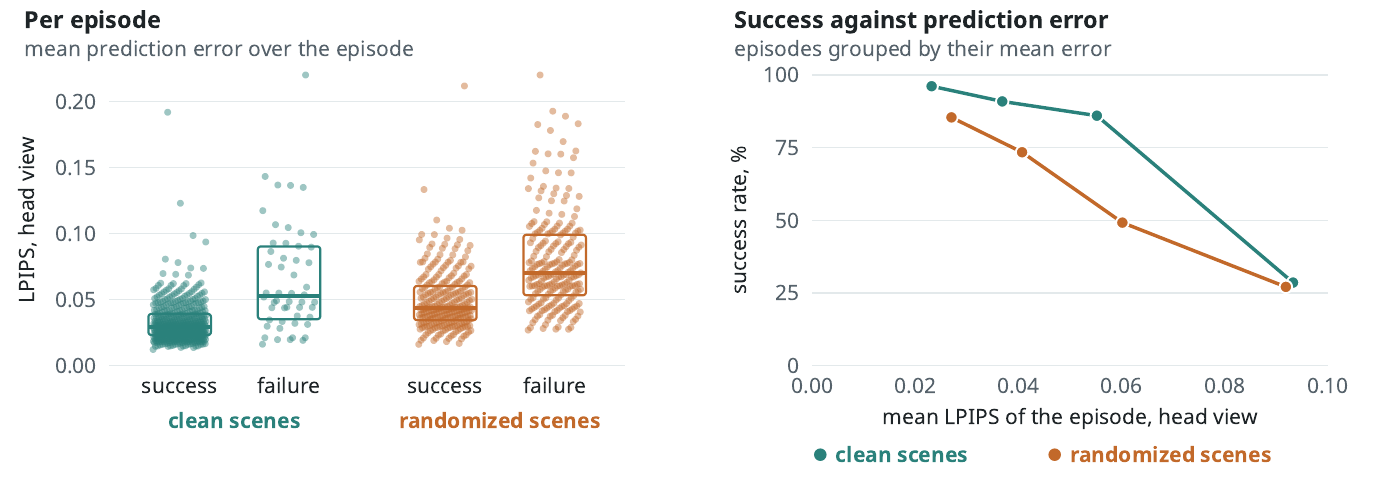}
  \caption{\textbf{Prediction error and success on RoboTwin~2.0.} LPIPS between the predicted and the executed head-camera frames, averaged over each episode. Left: successful and failed episodes in clean and randomized scenes (box: quartiles; line: median). Right: success rate in this run of the episodes grouped by their mean error (below 0.03, 0.03--0.05, 0.05--0.07 and above 0.07), drawn at the median error of each group.}\label{fig:imagination-error}
\end{figure}

Success follows imagination in both settings, but depends on it far more out of distribution (\cref{fig:imagination-error}). In clean scenes, success stays between 86\% and 96\% for all but the 6\% of episodes with the largest error. In randomized scenes, it falls from 85\% for the most accurate imagination to 27\% for the least accurate, and a third of these episodes are in that worst group. The link is not only a consequence of failing: judged by the first 100 steps alone, randomized episodes with an accurate imagination succeed twice as often as those with a poor one.

A review of 20 randomly drawn failures in each setting shows how they arise. In clean scenes, 19 of them imagine the executed scene correctly, outcome included: a block that is never released, blocks stacked in the wrong order, an object dropped beside its target (\cref{fig:imagination}). The policy foresees what its actions do and still chooses the wrong ones. In randomized scenes, 13 of the 20 failures come from how the policy sees the changed scene: its imagination departs from what happens, or it grasps a distractor instead of the target and imagines doing so. DreamZero attributes most of its failures to video generation rather than action prediction~\cite{dreamzero}; our measurements show that this holds out of distribution, while in distribution the policy imagines correctly and fails in its actions.

This accounts for the result of \cref{sec:findings-modes}. Retaining forward dynamics and image-to-video during adaptation raised randomized success by 8.0\% and left clean success unchanged: the training signal of video prediction pays off where failures come from imagination, not where the policy already imagines correctly. Robustness to new scenes thus goes with imagining them well, and \GreenWAM keeps the video prediction of its world-model role while it learns to act.

\subsection{Robustness to Scene Randomization}\label{sec:findings-random}

\begin{figure}[!t]
  \centering
  \includegraphics[width=\linewidth]{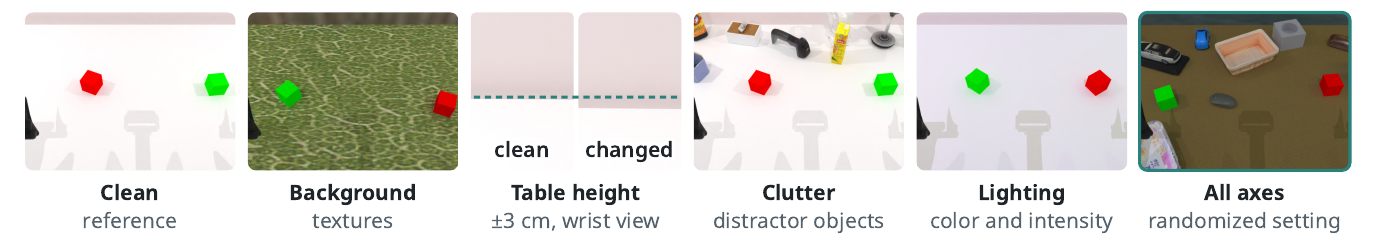}
  \caption{\textbf{Randomization axes of RoboTwin~2.0.} One scene of stack\_blocks\_two right after reset, without a policy, under each axis: background textures, table height (lowered by up to 3\,cm in the official setting, raised or lowered by 3\,cm in our test), distractor objects on the table, lighting color and intensity, and all axes together as in the official randomized setting. All tiles use the head camera except table height, which is barely visible there and is shown from the left wrist camera; the dashed line marks the table edge of the clean scene. The scene is generated anew for each setting, so object positions differ between tiles.}\label{fig:axes}
\end{figure}

Randomization changes appearance and geometry at once, so a single success rate does not show which change matters most. We therefore vary one axis at a time (\cref{fig:axes}) and then test appearance augmentation; \cref{tab:robotwin} lists the clean and randomized success of other methods with their training data.

\para{Individual scene changes.} The single-axis tests use seeds outside the official set. Background textures and table height cause the largest drops, 12\% and 8\% below the clean reference; distractors and lighting change success by less than 1\%, and unseen instructions do not lower it. A new background changes only the image, which augmentation can target. A table 3\,cm higher or lower changes the trajectory the arm must follow, and the augmentations used here change observations while keeping the recorded states and actions, so they add no demonstrations of such motions.

\para{Trajectory accuracy and replanning.} Execution is sensitive to small errors: replaying expert trajectories with joint noise of 0.05\,rad succeeds in 80\% of trials, and with 0.1\,rad in 30\%. Executing 8 instead of 16 predicted steps before replanning lowers clean success by 14.8\%.

\para{Appearance augmentation.}\label{sec:findings-augmentation} Background replacement targets the costliest appearance change: in a paired comparison on tasks chosen before training, it raises randomized success by 13.1\% and leaves clean success unchanged.

\section{Discussion and Limitations}\label{sec:discussion}

\para{One foundation for both roles.} A single robotics checkpoint yields both a world model and policies. The controlled comparisons connect the two roles: robot specialization improves forward prediction and policy success at the same budget, and the policy that predicts held-out frames more accurately also succeeds more often.

\para{World modeling as part of control.} Keeping forward dynamics and image-to-video next to WAM during adaptation raises randomized RoboTwin success by 8.0\%, although both variants predict video. Decoding the policy's own predictions shows the same link inside one model: out of distribution, success follows the accuracy of the imagined scene, which is where retaining world-model modes pays off (\cref{sec:findings-imagination}).

\para{One action space for all bodies.} One projector serves all 20 families, so a new robot needs a codec rather than new action parameters (\cref{app:anatomical-mappings}). The single-arm Franka of CALVIN and LIBERO, absent from the robotics corpus, enters through the same interface and reaches 4.64 on CALVIN and 97.4\% on LIBERO.

\para{Limitations.} The policies remain sensitive to changes in task geometry, such as table height on RoboTwin or the robot's initial configuration on LIBERO-plus: the augmentations used here change observations but add no new motions. Long rollouts continue from the model's own frames, so prediction error accumulates from block to block. Each benchmark is reached through its own post-training, and the real-robot evaluation covers one humanoid and one packing task.

\para{Outlook.} Natural next steps are running the distilled single-step policy on the robot's onboard computer, transfer to bodies that neither the parent nor the robotics stage has seen, motion-level augmentation for robustness to task geometry, and the use of the world-model role to evaluate policies in imagined rollouts.

\section{Conclusion}\label{sec:conclusion}

We asked whether one robot world model can serve both as a world model and as a policy across different bodies, and what makes this work. Adapted from Cosmos~3 Edge to 20 robot families through a shared anatomical action space and joint video--action training, one foundation yields a world model with top-tier trajectory, depth and 3D accuracy on WorldArena~1.0, ahead of all compared world--action models and video generators, and policies that score among the top entries of the RoboTwin~2.0 leaderboard, are competitive on CALVIN and LIBERO and, on a real humanoid, succeed more often than policies adapted from $\pi_{0.5}$ and from Cosmos~3 Edge. Controlled comparisons identify three choices that make this work: robot specialization, which improves both roles together, an action expert split off the generator, and world-model modes kept during policy adaptation. A distilled one-step student answers at 28.3\,Hz and keeps its lead over $\pi_{0.5}$ on the robot. We release the weights of the RoboTwin~2.0, LIBERO and CALVIN policies and of the WorldArena world model, together with the code that evaluates them. Because a new robot needs a codec rather than new action parameters, the next steps are bodies unseen in training and more reliable long-horizon prediction and control under changes in task geometry.

\FloatBarrier
{\footnotesize\setlength{\bibsep}{0pt}\setlength{\itemsep}{0pt}\enlargethispage{\baselineskip}
\bibliographystyle{abbrvnat}
\bibliography{greenwm}}

\FloatBarrier
\appendix
\section{Data Inventory and Anatomical Mappings}\label{app:data}

\subsection{Family and Source Inventory}

\begin{figure}[!htb]
  \centering
  \includegraphics[width=\linewidth]{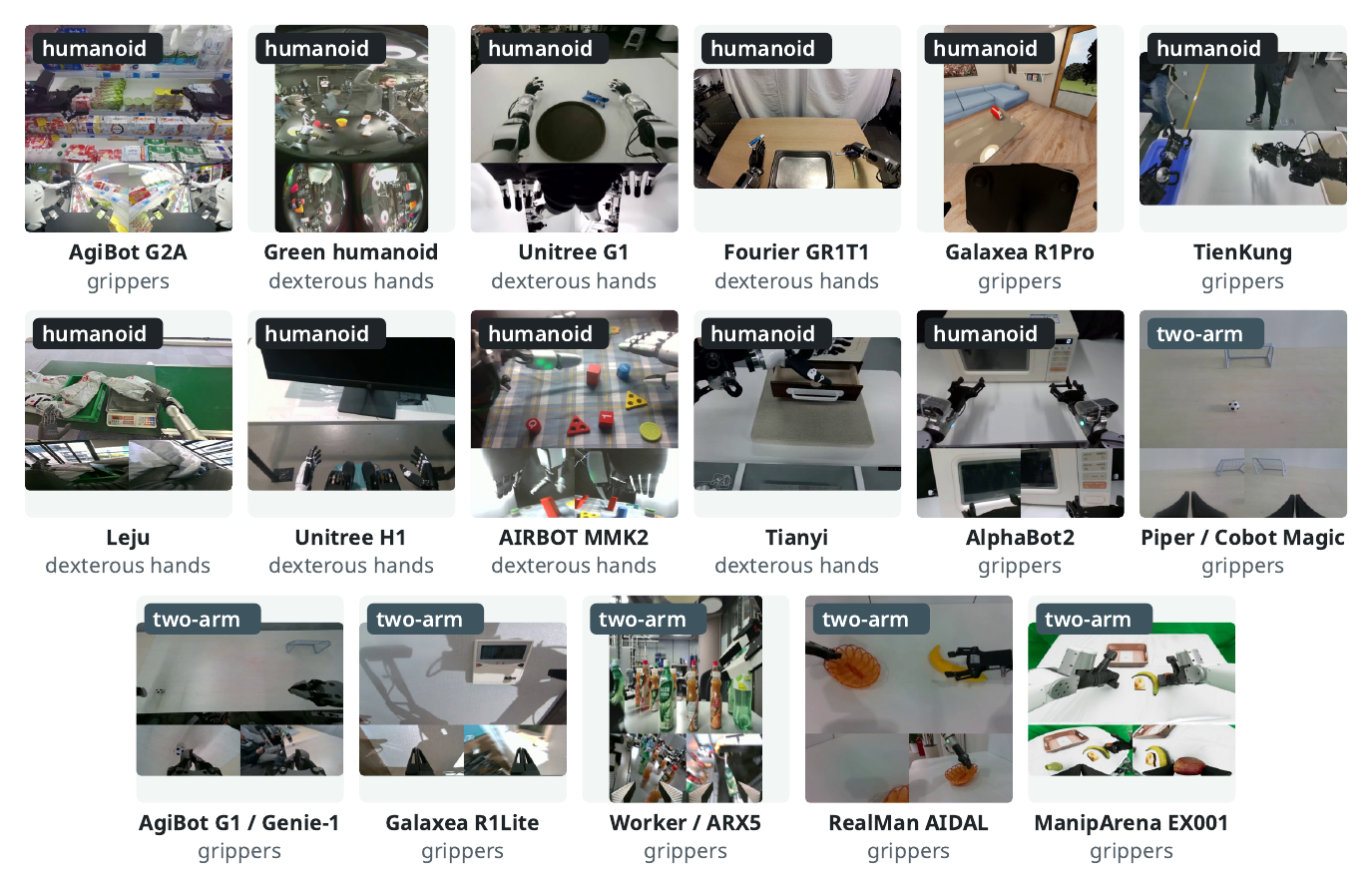}
  \caption{\textbf{Robot families of the corpus.} Training canvases of the robot families, as the model receives them: the head view on top and the other cameras in a strip below. Each tile shows one dataset of its family, so other datasets of the same family may differ in hands or cameras. Humanoids come first, then two-arm platforms; the chip on each tile gives the body type, and the label below gives the effector.}\label{fig:families}
\end{figure}

\begin{figure}[!htb]
  \centering
  \includegraphics[width=\linewidth]{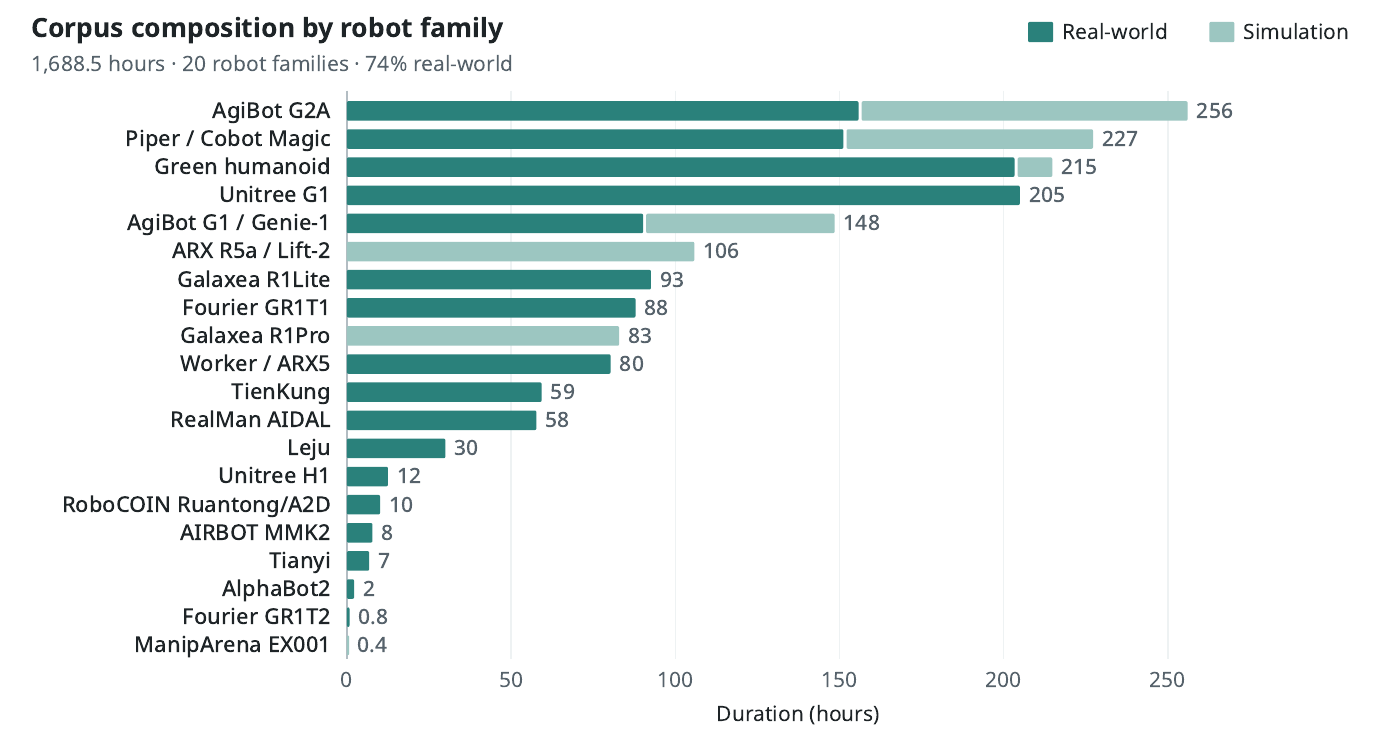}
  \caption{\textbf{Hours per robot family, real-world and simulated.} Combined names such as Piper / Cobot Magic denote datasets of one robot body grouped into one family (\cref{app:data}). End labels are rounded. Hour shares differ from the sampling probabilities used in training (\cref{app:mixture}).}
  \label{fig:corpus-families}
\end{figure}

The corpus groups datasets by the robot body they describe (\cref{fig:families}). All twelve Unitree G1 layouts, which differ in hands and cameras, share one family; the two Worker datasets share another, as do the Green humanoid and its simulator. The largest family contributions to the admitted corpus are AgiBot G2A (256.0\,h), Piper / Cobot Magic (227.2\,h), the Green humanoid (214.8\,h), Unitree G1 (204.9\,h) and AgiBot G1 / Genie-1 (148.5\,h). \Cref{fig:corpus-families} shows the 20 main families and their real/simulated composition. The inventory covers 1\,217 datasets.

Source-level filtering retains 1.43 million atoms totaling 1\,674.7\,h. InternData-A1~\cite{interndataa1}, the Green humanoid, RoboCOIN~\cite{robocoin}, GM100~\cite{gm100} and AgiBotWorld-2026~\cite{agibotworld2026} supply the largest retained volumes; further public sources include GR00T-Teleop-GR1~\cite{dreamdojo}, BEHAVIOR-1K~\cite{behavior1k}, Genie Sim~3.0~\cite{geniesim3}, RoboMIND~\cite{robomind}, Humanoid Everyday~\cite{humanoideveryday}, HIW-500~\cite{hiw500}, the G1 IKEA assembly recordings~\cite{g1ikea} and Unitree datasets~\cite{unitreedata}. Galaxea Open-World~\cite{galaxea} has the largest quality-filter reduction (6.84\,h, 9\% of its admitted hours).

\subsection{Corpus Admission and Quality Filtering}\label{app:admission}

\para{Sources.} Each source is converted to one LeRobot layout and registered with its camera roles, state layout and three static facts that the captioner later receives: embodiment, end-effector type and real or simulated domain. Camera roles and the meaning of state and action columns are established from metadata, sampled tables and video rather than from column names. A video stream needs more than 10\,FPS and at least 448 pixels on the short side; depth and infrared streams are not used. Failed and recovery demonstrations are kept, since for a world model they are valid physical rollouts.

\para{Semantic atoms.} An atom is the smallest annotated temporal unit of a recording, recovered from subtask tables, task-index runs, skill annotations or instruction intervals. An entire episode becomes one atom only when no finer annotation exists; missing or overlapping boundaries are marked unresolved rather than replaced by the whole episode. The source label is kept separately from the training instruction.

\para{Length.} Admission uses the camera in the head or ego role. The maximum admitted length is
\begin{equation}\label{eq:length}
e-s \leq \min\!\left(B(h),\lfloor 10\,f\rfloor\right),\qquad
 B(h)=\begin{cases}300,&h\leq480,\\
 200,&480<h\leq1000,\end{cases}
\end{equation}
where $h$ is the admission-view height and $f$ its frame rate. An overlong atom is rejected as a whole; the gate does not shorten it or invent new boundaries.

\para{Quality filtering.} The quality pass, computed on the admission camera, removes idle, repeated-frame, abruptly cut and unreadable clips (\cref{tab:dataqa}). An atom is idle when more than 70\% of its motion values fall below 0.2 of the episode median; a whole episode is compared with the median motion of episodes from the same dataset instead. Repeated frames require both a repeated share above 0.30 and a run of at least two repeated transitions, which separates sustained freezes from isolated duplicates left by frame-rate conversion. A cut requires a histogram-distance peak together with large color and grayscale changes.

\para{Splits and text.} Training and holdout are separated at the episode level, giving 1.40 million training and 28.6 thousand held-out atoms before text exclusions. Atoms without an accepted caption or derived command stay in the quality-admitted pool but do not enter the text-dependent modes; native instructions are not used as a fallback.

\begin{table}[!htbp]
  \centering\small
  \caption{\textbf{Data-quality pass over the admitted corpus.} Atoms in thousands, hours rounded to one decimal. An atom can be removed for several reasons (40 atoms are), so the reason rows do not add up to the total.}\label{tab:dataqa}
  \begin{tabular}{@{}lrr@{}}
    \toprule
    & Atoms, thousands & Hours \\
    \midrule
    Admitted subtask atoms & 1\,437 & 1\,688.5 \\
    Not scored, excluded & 0.1 & 0.2 \\
    \midrule
    \multicolumn{3}{@{}l}{\emph{Removed}} \\
    \quad idle clip (idle score above 0.7) & 7.3 & 8.5 \\
    \quad repeated frames & 2.1 & 2.9 \\
    \quad abrupt cut within the clip & 1.8 & 2.3 \\
    \quad unreadable video & $<0.1$ & $<0.1$ \\
    \quad total & 11.1 & 13.6 \\
    \midrule
    Kept for training & 1\,426 & 1\,674.7 \\
    \bottomrule
  \end{tabular}
\end{table}

\subsection{Anatomical Mappings}\label{app:anatomical-mappings}

\begin{figure}[!htb]
  \centering
  \includegraphics[width=\linewidth]{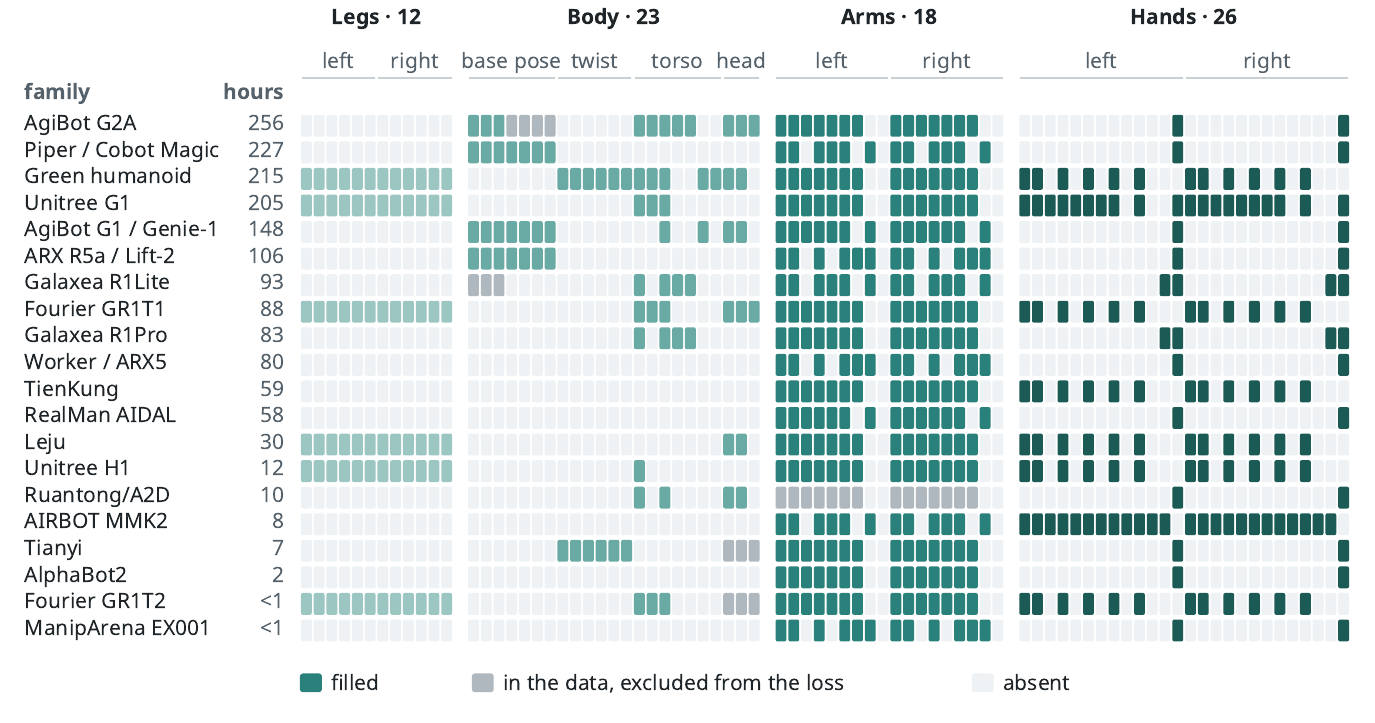}
  \caption{\textbf{Coverage of the 79 anatomical slots by family.} Rows: the 20 families ordered by admitted hours. Columns: slots grouped into legs (12), body (23: base pose, twist, torso and head), arms (18) and hands (26), with left and right subgroups. A colored cell is filled by a named channel of at least one layout of the family. A gray cell holds only unresolved channels of that body part: they stay in the data, are excluded from the loss and claim no correspondence to the slot. A light cell marks an absent slot. The hands of Ruantong/A2D are unresolved.}\label{fig:slot-coverage}
\end{figure}

\Cref{fig:slot-coverage} shows which slots each family fills. Each of the 43 registered source layouts has a codec that selects its channels, permutes them into slot order and fixes sign and zero for every slot role; scale and offset are absorbed by per-layout statistics. A coordinate enters a shared slot only when its correspondence follows from the URDF at zero pose, the manufacturer documentation or named channels, and for the main families the joint assignment is validated by forward kinematics against measured end-effector poses. Channels without an established correspondence stay outside the shared slots. A validity mask removes absent and unresolved coordinates from the loss, so equal tensor width never implies equal meaning. The Worker platform, for example, stores two 6-DoF arms and two grippers under generic channel names, and its codec assigns each channel an anatomical role (\cref{fig:interface}). Robots that appear only in post-training enter through the same interface as layout extensions: a new codec maps their channels onto existing slots, and the shared projectors of \GreenWAM stay unchanged. RoboTwin uses ARX5 arms, so its layout reuses the Worker family, whose forward kinematics reproduces the recorded RoboTwin gripper paths to within 5.2 and 7.1\,mm; the single Franka arm of CALVIN enters through the Cartesian end-effector slots. State rows are aligned to video frames per source. Normalization statistics are computed on the training split only, separately for absolute and offset encodings and per layout, and pooled per slot with a lower bound on the scale in physical units; WAM statistics use successful demonstrations only.

\section{Training and Adaptation Recipes}\label{app:recipes}

\subsection{Robotics Training Recipe}

\begin{figure}[!htb]
  \centering
  \includegraphics[width=\linewidth]{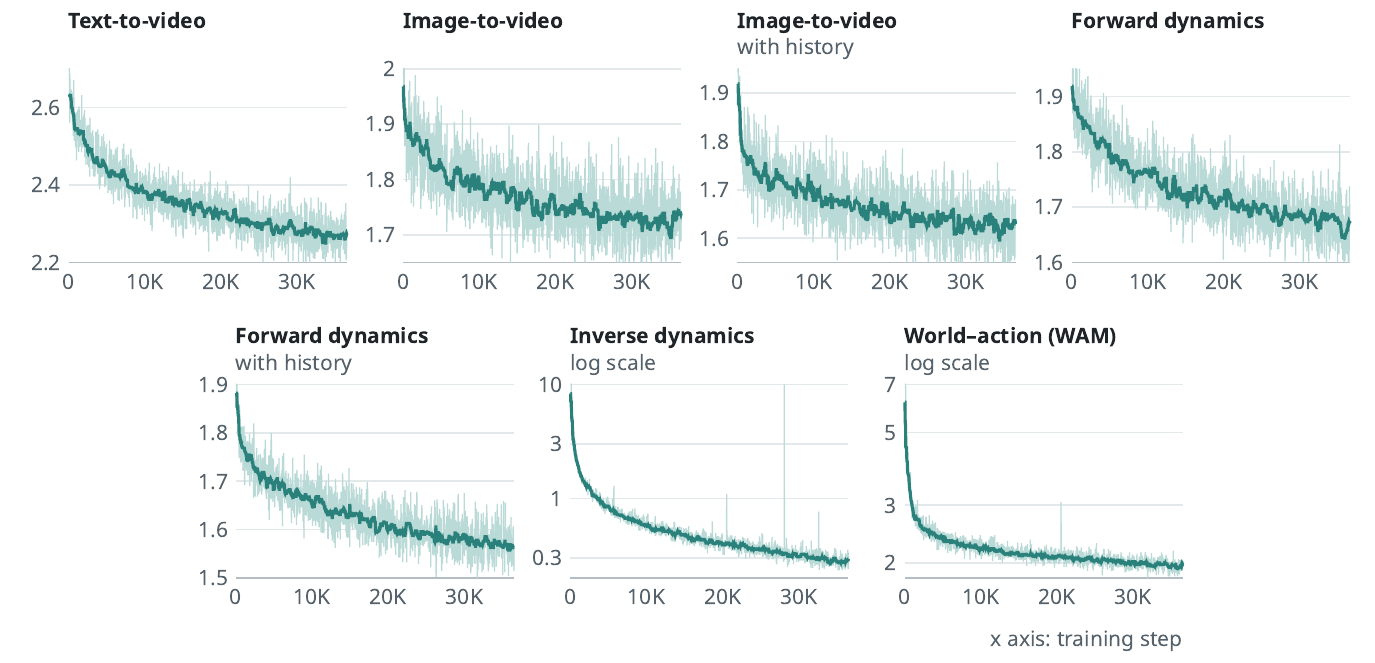}
  \caption{\textbf{Robotics-stage training.} Training loss of each mode as reported (light) and its running median over about 500 steps (dark); inverse dynamics and WAM are on a log scale because they start an order of magnitude higher, and their medians fall from 8.3 to 0.29 and from 6.2 to 1.98.}\label{fig:training-curves}
\end{figure}

\para{Recipe.} \Cref{fig:training-curves} shows the training losses. The noise level shared by all generated streams is drawn as in the video training of Cosmos~3 Edge: $\tilde\sigma=1-u-s\,(\cos^2\tfrac{\pi u}{2}-1+u)$ with $u\sim\mathcal{U}(0,1)$ and $s=1.29$, shifted to $\sigma=k\tilde\sigma/\big(1+(k-1)\tilde\sigma\big)$ with $k=5$ for 480p canvases. Each micro-batch packs 4 examples of one mode, so that examples with history anchors are not mixed with examples without them. The learning rate follows a square-root rule from the base rate of $2\times10^{-5}$ at batch 128, i.e.\ $\eta = 2\times10^{-5}\sqrt{B/128}$, and the newly initialized action projections and modality embeddings use a rate five times larger.

\begin{table}[!htbp]
  \centering\small
  \caption{\textbf{Recipe of the robotics stage.}}\label{tab:gw-recipe}\label{tab:app-gw}
  \begin{tabular}{@{}lp{.70\linewidth}@{}}
    \toprule
    Setting & Value \\
    \midrule
    Initialization & Cosmos 3 Edge 4B \\
    Trainable & Generator, action projections, action-modality embeddings \\
    Frozen & Reasoner, video VAE \\
    Modes, \% of examples & text-to-video 20, image-to-video 10, image-to-video with history 10, forward dynamics 15, forward dynamics with history 20, inverse dynamics 10, WAM 15 \\
    Frame rate & native, capped at 30 FPS \\
    Training unit & whole subtask; each autoregressive block is one subtask \\
    Rollout & up to 30\,s of future video in whole-subtask blocks; 4 RGB anchors of history \\
    WAM horizon & about 1.07\,s; successful demonstrations only \\
    Joint targets & inverse dynamics: absolute, including $q_0$; other modes: offsets from the block start \\
    Canvas sizes & 640$\times$640, 736$\times$544, 544$\times$736, 832$\times$480, 480$\times$832 \\
    Global batch & 256 examples (64 GPUs $\times$ 4 packed examples) \\
    Optimizer & AdamW~\cite{adamw}, $\beta=(0.9, 0.99)$, weight decay 0.05 \\
    Learning rate & $2.83\times10^{-5}$; $\times5$ on action projections and embeddings \\
    Schedule & 500 warm-up steps, then linear decay \\
    Gradient clipping & 1.0 \\
    Noise level & shared by all generated streams; Cosmos 3 Edge video schedule, shift 5 at 480p (\cref{sec:train-gw}) \\
    Loss & video and state terms, each scaled by 10; state: mean within valid body-part groups, then across groups \\
    Prompt & Cosmos 3 JSON, up to 768 tokens, text dropout 0.1 \\
    Validation & every 1\,200 steps \\
    Steps trained & 36\,500 (9.34 million examples) \\
    \bottomrule
  \end{tabular}
\end{table}

\subsection{Temporal Sampling and Prediction Targets}\label{app:temporal-sampling}

\para{Physical time.} We retain each source's physical timing and cap the effective video rate at 30 FPS. A source recorded at $f_s$ is read with stride $\max(1,\lceil f_s/30\rceil)$: 15, 20 and 30 FPS are retained, while 50 FPS becomes 25 FPS. Video frames and measured states are selected on the same time grid, and the effective rate is passed to the model. This preserves motion speed in physical time; joint positions and displacements are not multiplied by the sampling-rate ratio. Three-view sources use the 480p canvas buckets 640$\times$640, 736$\times$544, 544$\times$736, 832$\times$480 and 480$\times$832 according to the head-camera aspect ratio.

\para{Whole subtasks.} A fixed-length clip can cut an action in the middle and pair an incomplete grasp with the description of the whole grasp. Text-to-video, image-to-video, forward dynamics and inverse dynamics therefore use a complete prepared subtask rather than a fixed-length clip. In the autoregressive modes one block is one whole subtask; a following block is added only if it fits completely, and the last block is never truncated. The autoregressive budget is expressed in physical time, up to 30\,s of future video, so the number of predicted frames follows the frame rate of the source.

\para{History anchors.} Let $a$ be the start of the contiguous kept history and $b$ the last observed boundary, both in native-frame indices, and let $s$ be the stride on the effective video grid. With $N=\lfloor(b-a)/s\rfloor$, the four RGB anchors are
\begin{equation}\label{eq:history-anchors}
h_i=a+s\left\lfloor\frac{iN}{3}\right\rfloor,\qquad i=0,1,2,3.
\end{equation}
Each frame is encoded separately. At the start of an episode or an admitted rollout, $a=b$ and all four anchors are the initial frame; short histories can likewise repeat an anchor. Target frames begin at $b+s$, so every anchor precedes the predicted block. History extends across adjacent kept subtasks of the same episode and camera and resets at rejected gaps. A target block must fit within its admitted subtask; decoder padding does not supply missing history or target frames.

\para{WAM horizon.} WAM predicts a short future inside a subtask with a physical horizon of about 1.07\,s. Video and state targets cover this interval; VAE padding creates no additional action targets. The command is bound to the enclosing admitted atom and is retained when sampling the shorter WAM window. It specifies the ongoing task rather than promising its completion within $H$. WAM admits only positive success labels; failed, unknown and absent outcomes remain outside its training subset.

\begin{figure}[!htbp]
  \centering
  \includegraphics[width=\linewidth]{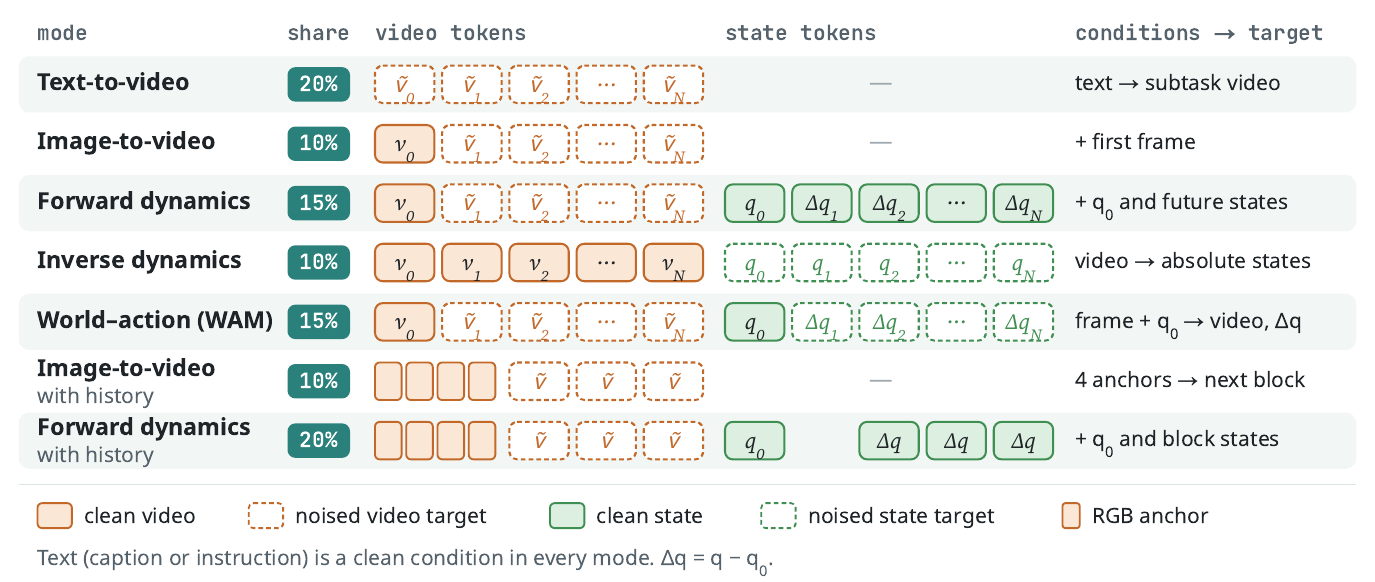}
  \caption{\textbf{Clean conditions and noised targets in the seven modes, with each mode's share of training examples.} Text (caption or instruction) is a clean condition in every mode. Forward dynamics, with and without history, conditions on $q_0$ and the clean future targets $\delta_k$ of \cref{eq:offsets}; inverse dynamics predicts absolute states including $q_0$. WAM predicts future frames and state offsets over about 1.07\,s at the effective frame rate of the source. Autoregressive modes predict the next whole-subtask block from four RGB anchors of the history.}
  \label{fig:modes}
\end{figure}

\subsection{Structured Captions and Commands}\label{app:caption-details}

A visual description records what is visible, including the scene and the observed outcome; an action instruction specifies what the robot should do during the same fragment. We keep both for every atom instead of treating a retrospective caption as a policy command.

\para{Captioner.} A vision-language model (Qwen3.8-27B~\cite{qwen38} with FP8 weights) captions the admission-camera clip of one atom. It also receives the clip duration, the source subtask label and three registry facts: embodiment (humanoid or bimanual platform), end effector (dexterous hand or two-finger gripper) and domain (real or simulation). The label supplies the goal and the target object, while only the frames establish how far the attempt got. The registry facts are taken as given, so a cropped or occluded effector does not change the described embodiment.

\para{Field-by-field generation.} The captioner writes one field per turn in a single dialogue, and each answer is appended to the context before the next request (\cref{tab:caption-fields}). Later fields must therefore agree with earlier ones, and one object keeps one name and one color throughout the record. Action phases start at 0:00, stay ordered and contiguous and end at the clip duration. A shared instruction block fixes the vocabulary: actor-relative left and right, the registry's effector term, no camera language, no hedging and no contact detail that the view cannot resolve. An attempt record, written before the action phases, states the end state, whether the target moved and where the effector ends; a clip that ends mid-motion then receives phases and a narrative that also end mid-motion.

\begin{table}[!htbp]
\centering\small
\caption{\textbf{Fields of the caption record in generation order.} The two checks are generated in the same dialogue but stored outside the caption and never enter a training prompt.}\label{tab:caption-fields}
\begin{tabular}{@{}r p{.22\linewidth} p{.47\linewidth} p{.19\linewidth}@{}}
\toprule
& Field & Content & Role \\
\midrule
1 & Subjects & Robot, and a person if present: embodiment, effector, location, orientation, pose, action, state changes, arm and leg counts; person attributes stay empty for the robot & Caption \\
2 & Background & Setting, furniture and objects with colors and materials; real or simulated domain & Caption \\
3 & Lighting & Conditions, direction, shadows, effect on the scene & Caption \\
4 & Aesthetics & Composition, color scheme, mood, patterns & Caption \\
5 & Context & One sentence: activity, setting, active effector and its side & Caption; rewriter \\
6 & Attempt record & End state in the final frame, whether the target moved, effector position at the end, outcome & Check \\
7 & Action phases & One to five timed phases: effector, object, where it goes & Caption; rewriter \\
8 & Temporal caption & One chronological paragraph of 120--180 words that ends on the final state & Caption; rewriter \\
9 & Consistency & Agreement with the label and the registry facts; mismatch clause & Check \\
\bottomrule
\end{tabular}
\end{table}

\para{Content gates.} Each answer is checked for forbidden phrases, effector terms that contradict the registry, contact claims and completion words that the attempt record rules out. A violation returns the phrase to the model for a rewrite of the same field, at most twice; a remaining violation restarts the atom, up to 4 attempts. Atoms without an accepted caption stay in the quality-admitted pool but not in the text-dependent modes.

\para{Coverage.} In total, 1.24 million atoms have an accepted visual caption and 187 thousand do not; 1.19 million of them also have an accepted command.

\para{From caption to command.} A second language-model pass turns the ordered action phases into one imperative command per phase, keeping the phase order, the effector, significant motions, releases and spatial relations. It does not see native operator commands, and partial or failed motions remain attempts rather than completed actions. A command is accepted only if every phase is present, in order and within the token budget; it is never truncated. \Cref{tab:caption-command-example} follows one atom from the source label to the derived command.

\begin{table}[!htbp]
\centering\small
\caption{\textbf{One atom from the source label to the derived command: the clip of \cref{fig:corpus-data}c (RoboCOIN, Unitree G1 with dexterous hands, real).} Captioner and rewriter outputs are quoted verbatim.}\label{tab:caption-command-example}
\begin{tabular}{@{}p{.22\linewidth} p{.74\linewidth}@{}}
\toprule
Source label & Grasp the bread from the table and place it into the plate. \\
Registry facts & humanoid; dexterous hand; real \\
\midrule
Action phases & 0:00--0:04\enspace The left dexterous hand reaches toward the orange and white bread roll on the left side of the table and grasps it. \\
& 0:04--0:07\enspace The left dexterous hand lifts the bread roll, carries it over the purple plate, and lowers it onto the plate. \\
& 0:07--0:10\enspace The left dexterous hand releases the bread roll on the plate and retracts away from the workspace. \\
Attempt record & target moved: yes; effector at end: away from the workspace; outcome: reached \\
Consistency & agrees \\
\midrule
Derived command & Extend the left dexterous hand toward the orange and white bread roll on the left side of the table and grasp it. Lift the bread roll, carry it over the purple plate, and lower it onto the plate. Release the bread roll on the plate and retract the left hand away from the workspace. \\
\bottomrule
\end{tabular}
\end{table}

\subsection{Training Mixture}\label{app:mixture}

Sampling in proportion to hours would let four families supply more than half of all examples (\cref{tab:clusters}); sampling uniformly over datasets would do the opposite and repeat the smallest datasets many times. We use a middle path. Sampling first chooses a mode, then a robot cluster, then a dataset and a window. Mode shares are those of \cref{tab:modes}. Cluster shares are fixed for our two main targets: the Worker/ARX5 cluster receives 12\% of examples and the Green humanoid cluster (real and simulation) 18\%. In this configured mixture, the remaining 19 clusters share 70\% in proportion to $\sqrt{N}$, where $N$ is the number of unique admissible training subtasks of the cluster:

\begin{equation}\label{eq:family-weight}
p_c = 0.70\cdot\frac{\sqrt{N_c}}{\sum_{c'} \sqrt{N_{c'}}} .
\end{equation}

The sampler first enumerates eligible unique subtasks for each mode, cluster and dataset after text admission, modality checks and temporal-window admission. Let $p_m$ and $p_c$ be the configured mode and cluster shares, and let $\mathcal{E}$ contain the mode--cluster pairs with eligible data. We obtain their joint probabilities by maximum entropy:
\begin{equation}\label{eq:joint-mixture}
\begin{gathered}
P^*=\arg\max_{P\geq0}\ -\sum_{m,c}P_{mc}\log P_{mc},\\
\sum_c P_{mc}=p_m,\qquad \sum_m P_{mc}=p_c,\qquad
P_{mc}=0\quad\text{if }(m,c)\notin\mathcal{E}.
\end{gathered}
\end{equation}
The implementation checks support feasibility before solving for the joint table, rather than renormalizing one marginal at the expense of the other. Within a supported pair, a dataset $d$ has conditional probability proportional to $\sqrt{N_{mcd}}$, where $N_{mcd}$ counts its eligible unique subtasks in that mode. Sampling then chooses a subtask uniformly within the dataset before selecting its temporal window and, where available, its joint or EEF view. Thus additional windows or a second action view do not increase a subtask's sampling mass. \Cref{tab:clusters} lists the configured cluster targets next to the admitted-hour shares; all sampling shares refer to examples.

\begin{table}[!htbp]
  \centering\small
  \caption{\textbf{Configured sampling targets by registered family, computed from the training split, next to admitted hours.} Targets are set before text and mode admission.}\label{tab:clusters}
  \begin{tabular}{@{}lrrr@{}}
    \toprule
    Family & Admitted, h & Hours, \% & Sampling, \% \\
    \midrule
    Green humanoid & 214.8 & 12.7 & 18.0 \\
    Worker / ARX5 & 80.3 & 4.8 & 12.0 \\
    AgiBot G2A & 256.0 & 15.2 & 9.1 \\
    Piper / Cobot Magic & 227.2 & 13.5 & 7.8 \\
    Unitree G1 & 204.9 & 12.1 & 7.2 \\
    Fourier GR1T1 & 87.9 & 5.2 & 6.9 \\
    AgiBot G1 / Genie-1 & 148.5 & 8.8 & 5.2 \\
    Galaxea R1Pro & 82.9 & 4.9 & 4.9 \\
    Galaxea R1Lite & 92.6 & 5.5 & 4.3 \\
    ARX R5a / Lift-2 & 105.8 & 6.3 & 4.2 \\
    TienKung & 59.3 & 3.5 & 4.1 \\
    RealMan AIDAL & 57.7 & 3.4 & 4.0 \\
    Leju & 30.0 & 1.8 & 3.2 \\
    AIRBOT MMK2 & 7.8 & 0.5 & 2.4 \\
    Unitree H1 & 12.5 & 0.7 & 2.2 \\
    RoboCOIN Ruantong/A2D & 10.1 & 0.6 & 1.7 \\
    Tianyi & 6.8 & 0.4 & 1.4 \\
    AlphaBot2 & 2.2 & 0.1 & 0.7 \\
    Fourier GR1T2 & 0.8 & $<0.1$ & 0.4 \\
    ManipArena EX001 & 0.4 & $<0.1$ & 0.2 \\
    OpenArm & $<0.01$ & $<0.01$ & 0.04 \\
    \bottomrule
  \end{tabular}
\end{table}

The table makes the trade-off explicit. Our two deployment targets are sampled well above their share of hours, the largest public families are sampled below it, and families with a few hours of data are seen several times more often than their volume alone would warrant, at the cost of repeating their subtasks.

\subsection{Policy Adaptation Recipes}\label{app:policy-recipes}

\para{RoboTwin~2.0.} The policy is trained under the leaderboard protocol on all 2\,500 clean demonstrations (50 per task); a split into 40 training and 10 held-out episodes per task is used only for recipe selection. The policy commands both arms and grippers in joint space. RoboTwin records a frame every 15 planner steps of 1/250\,s, so the physical frame rate is 16.67\,Hz rather than the 30\,Hz in its metadata, and the WAM horizon keeps the 1.07\,s of \GreenWAM. Episodes are cut into equal contiguous atoms of at most 10\,s, and WAM windows start at every frame where a full window fits, which gives about 470 thousand windows instead of 15 thousand tiled ones. The mode mixture is WAM-heavy (0.9 WAM, 0.05 forward, 0.05 image-to-video), each of the 14 action coordinates carries the same loss weight, and each sample draws a random paraphrase from the seen instructions of its episode. The reported policy is trained in stages from \GreenWAM with background replacement, color augmentation and the action expert. At evaluation the policy predicts 16 steps, samples them with 8 denoising steps without guidance and executes the whole prediction before replanning.

\para{WorldArena~1.0.} The world model starts from \GreenWAM and is adapted for 10\,000 steps with a global batch of 64, a peak learning rate of $1.41\times10^{-5}$ after 500 warm-up steps and a cosine decay to 0.1 of the peak. It is trained on the clean RoboTwin~2.0 demonstrations, whose ARX5 arms are those of the benchmark, rendered as a single head-camera canvas of 736$\times$544 and cut into blocks of at most 15\,s. Only forward dynamics with history is trained, and the text prompt is always empty, as in the action variant of the benchmark. At evaluation every block, including the first, is generated in this mode from four anchors of the history, which for the first block repeat the recorded first frame; sampling uses 35 steps without guidance.

\para{LIBERO.} The LIBERO policy is adapted from \GreenWAM with the action expert in stages, with appearance augmentation, a random perspective distortion of the third-person camera image and paraphrased instructions. No LIBERO-plus episode is used in training, and the same policy is evaluated on both benchmarks.

\para{RoboDojo.} The policy follows the official protocol: it is trained on the official training set and compared with the entries of the official leaderboard (\cref{app:public-benchmarks}). It uses the generator--expert architecture of \GreenWAM with attention restricted by stream: the history attends to the instruction and to itself, the current frame to the instruction, the history and itself, the state to the current frame and itself, and the actions to the generated video, the state and themselves. The history is pinned after the instruction, and the current frame, the generated video and the actions are placed at their time offsets from it. Training runs for 60\,000 steps with a global batch of 352, a learning rate warmed up over 3\,000 steps to $8\times10^{-5}$ and decayed by a cosine to $10^{-6}$, weight decay 0.01, an exponential moving average of the weights and quantile-normalized actions; a fifth of the training examples are image-to-video.

\para{CALVIN.} The CALVIN policy is trained on the ABC training split only (\cref{app:protocols}). Its single Franka arm enters \GreenWAM through the Cartesian end-effector slots with per-frame SE(3) offsets and the absolute initial pose as the current state. At evaluation the policy predicts 20 steps, samples them with 4 UniPC steps, a timestep shift of 3 and guidance 1.5, and executes the first 10 before replanning.

\para{Green humanoid.}\label{app:green-adaptation} The comparison uses three initializations: the public $\pi_{0.5}$ base checkpoint~\cite{openpi}, Cosmos~3 Edge, and \GreenWAM. They are adapted on the same Green demonstration split. The distilled policy starts from the reported \GreenWAM teacher and uses only the training split for the consistency and demonstration losses. Its consistency losses use a masked pseudo-Huber distance~\cite{ict}, with unavailable action coordinates and clean conditions excluded; the loss weights are $(1,1,0.01)$ and the EMA decay is 0.995. Only the online student is optimized; the teacher and the EMA target are used only during training. Validation episodes select checkpoints and inference settings; neither ID test episodes nor OOD trials are used for selection.

Each policy retains its own observation preprocessing, action representation and training recipe, with a common set of camera streams and controlled joints at the robot interface. Normalization statistics are fitted on training data.

\section{Evaluation Protocols}\label{app:protocols}

\begin{figure}[!htb]
  \centering
  \includegraphics[width=\linewidth]{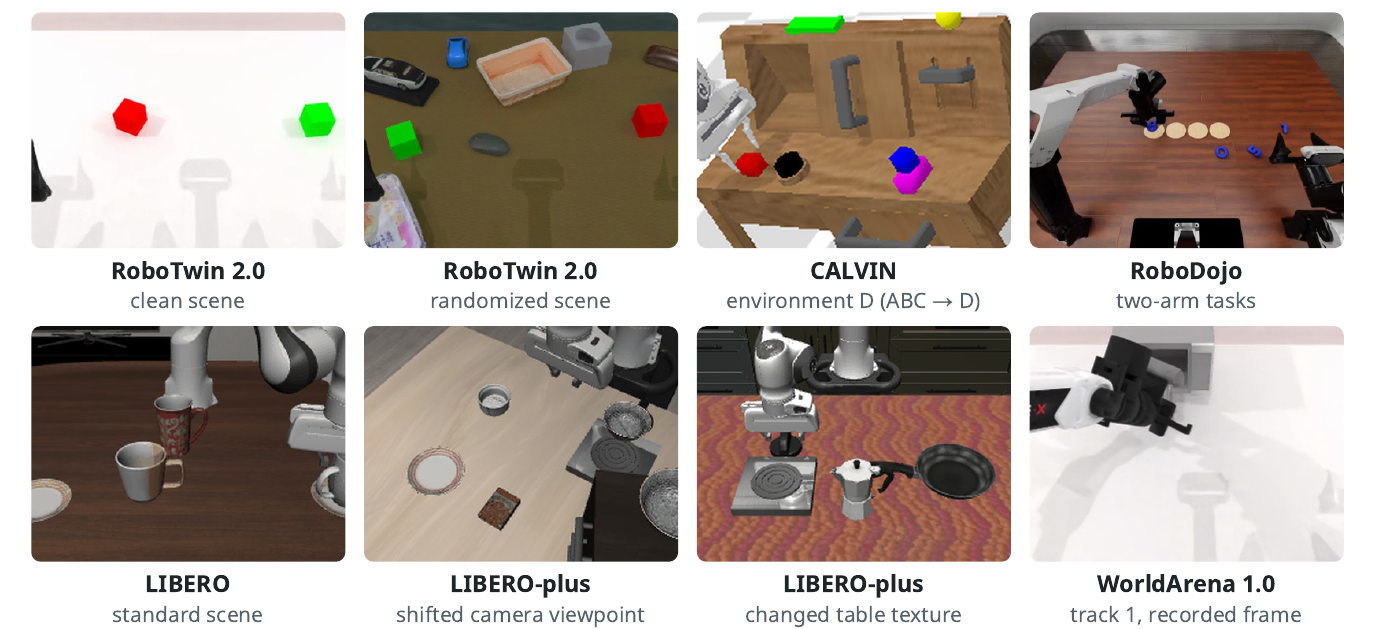}
  \caption{\textbf{Benchmark scenes.} One frame per benchmark and setting. The RoboTwin~2.0, LIBERO and LIBERO-plus frames come from evaluation episodes of our policies: the first head-camera frames of the clean and randomized episodes of \cref{fig:robotwin-rollouts}, a standard LIBERO scene, a LIBERO-plus scene with the camera turned by $-32^\circ$ and one with a changed table texture. The other frames are recorded benchmark data: the static camera of an environment-D episode of CALVIN, the head camera of a RoboDojo demonstration and a recorded frame of a WorldArena~1.0 test episode.}\label{fig:benchmark-scenes}
\end{figure}

\begin{table}[!htbp]
  \centering\small
  \caption{\textbf{Evaluation protocols.} \Cref{fig:benchmark-scenes} shows one frame of each benchmark.}\label{tab:protocols}
  \begin{tabular}{@{}l>{\raggedright\arraybackslash}p{.31\linewidth}>{\raggedright\arraybackslash}p{.34\linewidth}@{}}
    \toprule
    Benchmark & Protocol & Metric \\
    \midrule
    WorldArena 1.0, track 1~\cite{worldarena} & Action-conditioned prediction on 1\,000 test episodes & Perceptual and embodiment metrics; motion-capped aggregate in \cref{fig:benchmarks-worldarena1} \\
    RoboTwin 2.0~\cite{robotwin2} & 50 bimanual tasks, 100 trials per task, clean and randomized scenes & Success rate, clean and randomized \\
    LIBERO~\cite{libero} & Four task suites of a single Franka arm in simulation & Average success over the suites \\
    LIBERO-plus~\cite{liberoplus} & LIBERO tasks under seven perturbation axes, zero-shot after standard LIBERO training & Success over all perturbed episodes \\
    RoboDojo~\cite{robodojo} & 42 simulation tasks in five capability dimensions, 150 trials per task over 3 seeds & Score and success rate averaged over dimensions \\
    CALVIN~\cite{calvin} & ABC$\rightarrow$D, chains of five instructions & Average completed length \\
    Green humanoid & Shared ID/OOD trial set for $\pi_{0.5}$, Cosmos~3 Edge, \GreenWAM{} and its distilled student & Instruction success by split \\
    \bottomrule
  \end{tabular}
\end{table}

\subsection{Public Benchmarks}\label{app:public-benchmarks}

\para{RoboTwin 2.0.} The benchmark comprises 50 bimanual tasks on the Aloha-AgileX embodiment, whose arms share the ARX5 kinematics (\cref{app:anatomical-mappings}). The leaderboard protocol trains on 50 clean demonstrations per task (2\,500 in total) and evaluates 100 trials per task in clean scenes (clean2clean, easy) and in randomized scenes with clutter, random lighting, background textures and table height (clean2random, hard); the leaderboard ranks by the average of the two. A leaderboard entry requires public code, public weights and a technical report~\cite{robotwinboard}.

\para{WorldArena.} Evaluation uses track~1 of WorldArena~1.0. The metric breakdown includes image quality, aesthetic quality, JEPA similarity, dynamic degree, flow score, motion smoothness, subject consistency, background consistency, photometric consistency, interaction quality, trajectory accuracy, depth accuracy, perspectivity, instruction following and semantic alignment~\cite{worldarena}. \GreenWAM is evaluated on all 1\,000 test episodes. For \GreenWAM, dynamic degree, flow score and motion smoothness are capped separately against each episode's ground truth before averaging. For competitors, only model-level metric means are available, so the caps are the corresponding ground-truth means 23.05, 12.50 and 69.79 over all 1\,000 reference episodes. This follows the motion-capping rule of WorldArena~2.0~\cite{worldarena2,worldarena2motion}; the videos and test episodes remain those of WorldArena~1.0. The recomputed aggregate therefore differs from the official WorldArena~1.0 leaderboard score. The figure shows a selected comparison: BLM, WorldScape~v0.2, GigaWorld-1, GE-Sim~2.0 and SisyphusWorld are omitted. Visual quality combines two scores computed without reference to the recording, MUSIQ image quality~\cite{musiq} and LAION aesthetic quality~\cite{laion5b}, with JEPA similarity; \GreenWAM reaches 57.2 against 59.5 for the recordings, while 103 of the 127 board entries score above the recordings in image quality. Instruction following and interaction quality are assigned by a vision--language judge, Qwen3-VL in the WorldArena evaluator~\cite{worldarena}. Across the judged \GreenWAM videos, both correlate weakly with trajectory accuracy ($r=0.05$) and more strongly with aesthetic quality ($r=0.29$ and 0.32), and 41 and 67 board entries score above the recordings on them.

\para{RoboDojo.} The benchmark comprises 42 simulation tasks grouped into five capability dimensions (generalization, memory, precision, long-horizon, open-vocabulary instructions) and 18 real-world tasks. Each policy is evaluated with 150 trials per task over three seeds; score and success rate are averaged within dimensions and then across dimensions~\cite{robodojo}. The simulated robot is a bimanual ARX5 at 25\,Hz with a head and two wrist cameras; training uses 3\,500 demonstrations, 100 for each of 34 tasks and 100 auxiliary domain-randomized ones, and the open-vocabulary tasks are not in training. A verified leaderboard entry reports three seeds on the public layouts, passes a hidden-layout consistency check and releases the evaluated checkpoint and code~\cite{robodojoboard}.

\para{LIBERO and LIBERO-plus.} LIBERO provides suites of simulated tasks for a single Franka arm~\cite{libero}; following common practice~\cite{openvlaoft}, evaluation uses LIBERO-Spatial, LIBERO-Object, LIBERO-Goal and LIBERO-Long, the long-horizon subset of LIBERO-100, and reports the success rate averaged over the four suites. LIBERO-plus applies perturbations of camera viewpoint, robot initial state, language, lighting, background, sensor noise and object layout to the LIBERO tasks, with one trial per perturbed episode; in the zero-shot setting the policy is trained on standard LIBERO only~\cite{liberoplus}.

\para{CALVIN.} Policies train on environments A, B and C and are evaluated in the unseen environment D on chains of five language instructions; the metric is the average number of completed instructions~\cite{calvin}. Training uses the ABC training split only: no episode of environment D, including its validation split, is used for training, and results are reported on 1\,000 chains.

\subsection{Real-Humanoid Comparison Protocol}\label{app:green-protocol}

\para{Tasks and splits.} The packing task is given as five consecutive instructions without resetting the scene (\cref{sec:exp-green}). ID trials use new resets of the objects and scenes represented in policy adaptation; OOD trials change objects or scene conditions excluded from adaptation, validation and distillation data. Every policy is evaluated on the same 14 scene configurations, 9 ID and 5 OOD, each with the five instructions of the task, which gives 45 ID and 25 OOD instructions per policy, each attempted once. All policies receive the same instructions, cameras and controlled joints.

\para{Success.} An instruction succeeds when it is completed before its timeout without human assistance; failures, timeouts and stops caused by the policy count as failed instructions. For split $d$ and policy $p$, let $s_{pd}$ be the number of successful instructions and $n_{pd}$ the number of evaluated instructions. The split score is
\begin{equation}\label{eq:success-rate}
\mathrm{SR}_{p,d}=100\,\frac{s_{pd}}{n_{pd}},
\end{equation}
reported separately for ID and OOD.

\subsection{Latency Benchmark}\label{app:latency}

We benchmark inference latency on a single NVIDIA RTX 4090 with FlashAttention-2 (\cref{tab:latency}). The video branch works on three latent frames: one conditioning frame and two generated frames, which decode into an eight-frame prediction at 10\,Hz. Each request also conditions on the proprioceptive state and a 128-token instruction. To isolate the cost of the model, the observation and the state stay fixed within an episode, and the instruction changes once, at its start. Median latency is the steady-state latency after the instruction-dependent quantities have been cached; time to first action (TTA) is the latency of the first request after a new instruction and includes building that cache. Each value is computed over 100 episodes, and the rate is 1000 divided by the latency in milliseconds.

\begin{table}[!htbp]
  \centering\small
  \caption{\textbf{Policy inference latency on one NVIDIA RTX 4090.} Optimizations are applied cumulatively to the \GreenWAM policy, and speedups are relative to it. Median latency over 100 episodes; TTA is the time to the first action after a new instruction.}\label{tab:latency}
  \begin{tabular*}{\linewidth}{@{\extracolsep{\fill}}lrrrr@{}}
    \toprule
    Configuration & Median, ms & Rate, Hz & TTA, ms & Speedup \\
    \midrule
    \GreenWAM & 435.9 & 2.3 & 435.9 & 1.00$\times$ \\
    \quad + caching & 411.1 & 2.4 & 430.8 & 1.06$\times$ \\
    \quad + guidance packing & 314.1 & 3.2 & 334.0 & 1.39$\times$ \\
    \quad + first-frame VAE & 236.4 & 4.2 & 256.5 & 1.84$\times$ \\
    \quad + compilation & 184.6 & 5.4 & 195.5 & 2.36$\times$ \\
    \quad + W8A8 & 134.6 & 7.4 & 148.6 & 3.24$\times$ \\
    \quad + distillation & \textbf{35.4} & \textbf{28.3} & \textbf{48.8} & \textbf{12.32$\times$} \\
    \bottomrule
  \end{tabular*}
\end{table}

The optimizations are applied in turn. Caching reuses the instruction and vision key--value states across denoising steps, so later steps run only the generative backbone. Guidance packing evaluates the conditional and empty-prompt branches of classifier-free guidance in one forward pass and also caches the empty-prompt key--value states. First-frame VAE encoding passes only the conditioning frame through the VAE encoder, since the other two latent frames are generated by the policy. Compilation traces the generator blocks together with this single-frame encoder, and W8A8 runs the feed-forward layers with 8-bit weights and activations. Distillation replaces the four guided denoising steps with one consistency step and keeps the instruction cache. Once caching is enabled, a new instruction adds 11--20\,ms to the first request, and 13.4\,ms for the distilled policy.

\subsection{Paired Comparisons}\label{app:paired}

Let $y^{A}_{te},y^{B}_{te}\in\{0,1\}$ be the outcomes of the two runs on episode $e$ of task $t$, over the common seeds $E_t$ of the two runs. The reported difference averages over tasks, as the benchmark does:

\begin{equation}\label{eq:paired-diff}
\Delta=\frac{1}{|\mathcal{T}|}\sum_{t\in\mathcal{T}}\frac{1}{|E_t|}\sum_{e\in E_t}\big(y^{A}_{te}-y^{B}_{te}\big).
\end{equation}

Its 95\% interval comes from a hierarchical bootstrap with 5\,000 replicates: tasks are resampled with replacement, then episode pairs within each resampled task, and the interval spans the 2.5th to 97.5th percentiles. The interval therefore describes the uncertainty of evaluating the two checkpoints, not the variation between training runs. Forward prediction uses the same estimator with PSNR per clip in place of $y$.

\subsection{Evaluation Details}\label{sec:findings-eval}

With an empty text prompt, as in the action variant of WorldArena, classifier-free guidance has no effect, and that variant is sampled without it. CALVIN test instructions are excluded from the language-augmentation pools. The reported RoboTwin result uses 100 trials per task and setting, and the controlled comparisons pair episodes on common seeds (\cref{eq:paired-diff}).

\section{Additional Results and Qualitative Examples}\label{app:results}

\subsection{Benchmark Tables}\label{app:sim-tables}

\Cref{tab:worldarena-categories} gives the WorldArena categories behind \cref{fig:benchmarks-worldarena1}, and \cref{tab:calvin,tab:robodojo,tab:libero} list the CALVIN, RoboDojo, LIBERO and LIBERO-plus results summarized in \cref{fig:benchmarks-manipulation}, with the source of every value.

\begin{table}[!htbp]
  \centering\footnotesize
  \caption{\textbf{WorldArena 1.0, track 1: overall score, categories and fidelity of the models in \cref{fig:benchmarks-worldarena1}.} EWMScore (\cref{eq:ewmscore}) is the mean of the 15 board metrics with dynamic degree, flow score and motion smoothness capped at the ground truth: per video for \GreenWAM, at the leaderboard means for the other models (an upper bound for them). The last row scores the recorded test videos themselves with the same evaluator.}\label{tab:worldarena-categories}
  \setlength{\tabcolsep}{3.5pt}
  \begin{tabular}{@{}lrrrrrrrrr@{}}
    \toprule
    & & \multicolumn{5}{c}{Categories} & \multicolumn{3}{c}{Fidelity to ground truth} \\
    \cmidrule(lr){3-7}\cmidrule(l){8-10}
    Model & EWMScore & Visual & Content & Physics & 3D & Control & JEPA & Traj. & Depth \\
    \midrule
    \textbf{\GreenWAM} & 63.31 & 57.2 & 65.6 & 62.2 & 93.4 & 84.1 & 98.4 & 56.0 & 98.2 \\
    \midrule
    BWM-Fast~\cite{bwm} & 63.80 & 63.1 & 58.2 & 62.4 & 91.8 & 89.7 & 97.9 & 44.9 & 86.4 \\
    UNIS~\cite{xiaomiu0} & 63.65 & 61.8 & 55.9 & 64.6 & 92.0 & 91.6 & 90.6 & 41.9 & 85.2 \\
    ABot-PhysWorld (text)~\cite{abotphysworld} & 63.65 & 64.4 & 63.4 & 56.7 & 85.5 & 90.8 & 90.4 & 31.5 & 72.0 \\
    Motubrain~\cite{motubrain} & 63.16 & 64.0 & 59.5 & 58.5 & 91.5 & 85.8 & 96.9 & 43.1 & 88.7 \\
    FlowWAM-FiveAges~\cite{flowwam} & 62.99 & 63.5 & 57.3 & 60.5 & 90.9 & 87.0 & 95.3 & 44.4 & 85.6 \\
    CogVideoX~\cite{cogvideox} & 61.30 & 55.8 & 67.7 & 50.9 & 88.3 & 82.6 & 94.8 & 34.8 & 91.1 \\
    Wan2.6~\cite{wan} & 60.96 & 61.4 & 64.0 & 42.7 & 84.7 & 89.0 & 72.6 & 12.2 & 75.4 \\
    Ctrl-World~\cite{ctrlworld} & 60.56 & 57.4 & 62.2 & 55.4 & 88.5 & 78.2 & 92.8 & 48.2 & 93.2 \\
    IRASim~\cite{irasim} & 59.63 & 56.1 & 65.7 & 49.3 & 88.8 & 77.4 & 93.7 & 35.9 & 93.5 \\
    Cosmos-Predict 2.5~\cite{cosmospredict25} & 57.83 & 58.1 & 61.7 & 44.0 & 87.5 & 75.2 & 90.7 & 27.5 & 90.3 \\
    \midrule
    \textit{Recorded test videos} & 67.23 & 59.5 & 66.7 & 79.6 & 95.4 & 87.2 & 99.9 & 86.9 & 100.0 \\
    \bottomrule
  \end{tabular}
\end{table}

\begin{table}[!htb]
\begin{minipage}[t]{.40\linewidth}\vspace{0pt}
  \centering\small
  \caption{\textbf{CALVIN ABC$\rightarrow$D, 1\,000 chains.} Average number of completed instructions in a chain of five. Values are taken from Xiaomi-Robotics-0~\cite{xiaomir0} (Table~2), except DreamVLA~\cite{dreamvla} (Table~1), MMaDA-VLA~\cite{mmadavla} (Table~1) and $\pi_{0.5}$, the supervised fine-tuned checkpoint reported in $\pi_{\mathrm{RL}}$~\cite{pirl} (Table~5).}\label{tab:calvin}
  \begin{tabular*}{\linewidth}{@{\extracolsep{\fill}}lc@{}}
    \toprule
    Method & Avg.\ length \\
    \midrule
    \multicolumn{2}{l}{\textit{Published methods}} \\
    $\pi_{0.5}$ (SFT)~\cite{pi05} & 3.84 \\
    RoboVLMs~\cite{robovlms} & 4.25 \\
    Seer-Large~\cite{seer} & 4.28 \\
    VPP~\cite{vpp} & 4.29 \\
    UniVLA~\cite{univlawang} & 4.41 \\
    DreamVLA~\cite{dreamvla} & 4.44 \\
    FLOWER~\cite{flower} & 4.53 \\
    Xiaomi-Robotics-0~\cite{xiaomir0} & 4.75 \\
    MMaDA-VLA~\cite{mmadavla} & 4.78 \\
    \midrule
    \textbf{\GreenWAM} & 4.64 \\
    \bottomrule
  \end{tabular*}
\end{minipage}\hfill
\begin{minipage}[t]{.56\linewidth}\vspace{0pt}
  \centering\small
  \caption{\textbf{RoboDojo simulation benchmark.} Score and success rate averaged over the five capability dimensions, from the official leaderboard~\cite{robodojoboard}: the top five entries and the methods shown in \cref{fig:benchmarks-manipulation}. Human teleoperation reaches a score of 80.42 and a success rate of 76.03\%~\cite{robodojo}.}\label{tab:robodojo}
  \begin{tabular*}{\linewidth}{@{\extracolsep{\fill}}lccc@{}}
    \toprule
    Method & Rank & Score & \makecell{Success\\rate, \%} \\
    \midrule
    \multicolumn{4}{l}{\textit{Leaderboard entries}} \\
    PhysicalRSI & 1 & 36.27 & 31.38 \\
    Awomo-0.5 & 2 & 35.34 & 29.64 \\
    Simate-beta & 3 & 33.95 & 27.96 \\
    Rex-M1 Preview & 4 & 33.79 & 27.44 \\
    VPP2-Preview & 5 & 31.40 & 25.62 \\
    DM0.5 & 10 & 24.90 & 19.34 \\
    Xiaomi-Robotics-1 & 13 & 20.07 & 13.93 \\
    $\pi_{0.5}$~\cite{pi05} & 20 & 11.44 & 6.93 \\
    X-VLA~\cite{xvla} & 24 & 10.13 & 6.52 \\
    X-WAM~\cite{xwam} & 28 & 7.69 & 3.83 \\
    \midrule
    \multicolumn{4}{l}{\textit{Our adapted policy}} \\
    \textbf{\GreenWAM} & & 16.12 & 10.49 \\
    \bottomrule
  \end{tabular*}
\end{minipage}
\end{table}

\begin{table}[!htb]
  \centering\small
  \caption{\textbf{LIBERO and LIBERO-plus.} Average success over the four LIBERO task suites and zero-shot LIBERO-plus success over all perturbed episodes, \%.}\label{tab:libero}
  \begin{tabular*}{\linewidth}{@{\extracolsep{\fill}}lcc@{}}
    \toprule
    Method & LIBERO & LIBERO-plus, zero-shot \\
    \midrule
    Faster-WAM~\cite{fasterwam} & 99.0 & 73.6 \\
    CKT-WAM~\cite{cktwam} & 98.8 & 86.1 \\
    Cosmos Policy~\cite{cosmospolicy} & 98.5 & 82.2 \\
    OA-WAM~\cite{oawam} & 97.8 & 83.9 \\
    Fast-WAM~\cite{fastwam} & 97.6 & 49.1 \\
    OpenVLA-OFT~\cite{openvlaoft} & 97.1 & 69.6 \\
    $\pi_{0.5}$~\cite{pi05} & 96.9 & 85.7 \\
    \midrule
    \textbf{\GreenWAM} & 97.4 & 86.8 \\
    \bottomrule
  \end{tabular*}
\tablenote{\GreenWAM: the same policy for both benchmarks (recipe in \cref{app:policy-recipes}); LIBERO-plus over all 10\,030 perturbed episodes, one trial each. LIBERO: Cosmos Policy~\cite{cosmospolicy}, Table~1 ($\pi_{0.5}$, OpenVLA-OFT, Cosmos Policy), OA-WAM~\cite{oawam}, Table~1, the LIBERO table of Faster-WAM~\cite{fasterwam} (Fast-WAM, Faster-WAM) and CKT-WAM, Table~9. LIBERO-plus: OA-WAM, Table~2, zero-shot after standard LIBERO training ($\pi_{0.5}$, Cosmos Policy, OpenVLA-OFT, OA-WAM), the LIBERO-plus table of Faster-WAM (Fast-WAM, Faster-WAM) and CKT-WAM~\cite{cktwam}, Table~1. All LIBERO-plus values are success rates over all perturbed episodes, as in the benchmark.}
\end{table}

\subsection{Failure Cases}\label{app:failures}

\begin{figure}[!tb]
  \centering
  \includegraphics[width=.9\linewidth]{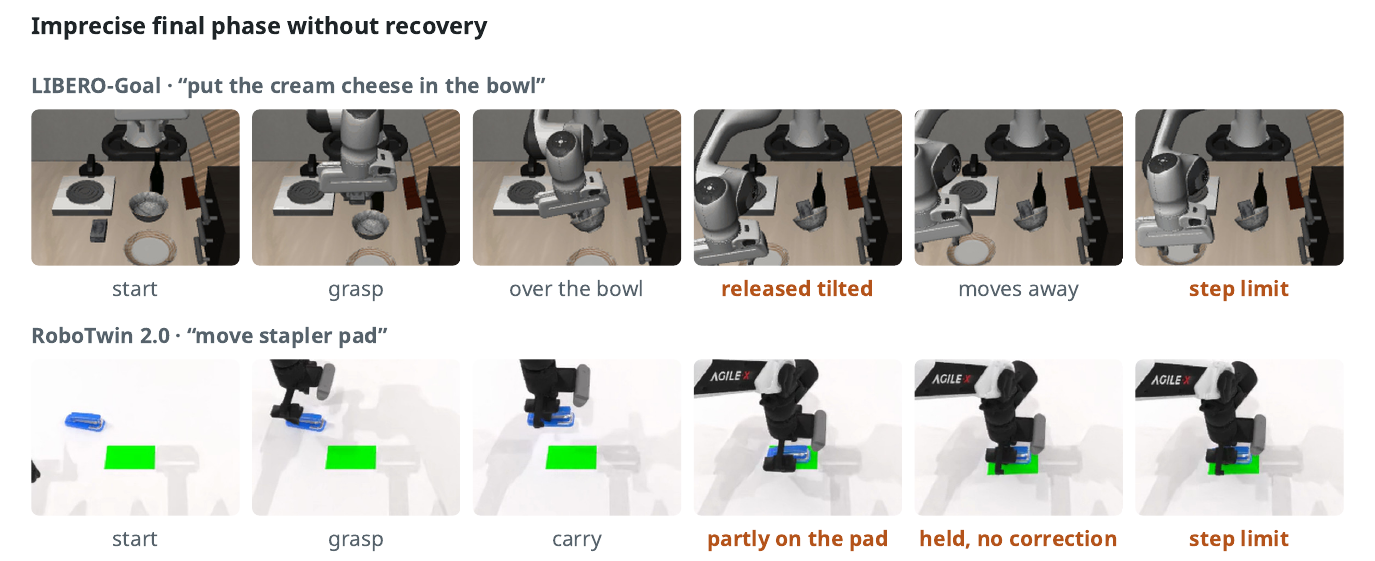}
  \caption{\textbf{Failure cases of the LIBERO and RoboTwin policies} (\cref{tab:libero,tab:robotwin}). Imprecise final phase without recovery. LIBERO-Goal, ``put the cream cheese in the bowl'': the cheese reaches the bowl but rests tilted on its rim, which the success check does not accept; the policy then moves toward the plate and does not return before the limit of 310 steps. RoboTwin~2.0, move\_stapler\_pad in a clean scene: the stapler is carried to the green pad but released only partly onto it, and the gripper holds it without correction until the limit of 400 control steps.}\label{fig:failures}
\end{figure}

\Cref{fig:failures} shows typical failures of the LIBERO and RoboTwin policies. In both episodes the task is carried out up to its last phase: the object is grasped, carried and brought to its target. The error lies in the final placement, a cream cheese tilted on the rim of the bowl and a stapler released only partly onto the pad, and the policy does not correct it before the step limit. The errors lie in precise placement and recovery.

\end{document}